\def\draft{1}  

\documentclass[11pt,letterpaper]{article}
\usepackage[margin=1in]{geometry}

\ifnum\draft=1
\fi 

\usepackage[utf8]{inputenc}
\usepackage{amsthm}
\usepackage{complexity}
\usepackage{mathtools,amsmath,amssymb}
\usepackage{graphicx}
\usepackage[table,xcdraw]{xcolor}
\usepackage{multirow}
\usepackage{tabularx}
\usepackage{makecell}
\usepackage[colorinlistoftodos]{todonotes}
\usepackage[colorlinks=true, allcolors=blue]{hyperref}
\usepackage{enumitem}
\usepackage{thm-restate}
\usepackage{natbib}
\usepackage{pifont}
\usepackage{subcaption}
\usepackage{wrapfig}
\usepackage{booktabs}

\usepackage[colorlinks=true, allcolors=blue]{hyperref}
\usepackage[nameinlink,capitalise]{cleveref}
\hypersetup{
    citecolor={violet}
}

\ifnum\draft=1
\newcommand{\Cnotes}[1]{{\color{red} [Chi-Ning: #1}]}

\else  \ifnum\draft=0
\newcommand{\Snotes}[1]{}
\newcommand{\Cnotes}[1]{}
\fi \fi 

\usepackage{thmtools}
\usepackage{thm-restate}
\RequirePackage[colorlinks=true]{hyperref}
\hypersetup{
  linkcolor=[rgb]{0,0,0.5},
  citecolor=[rgb]{0, 0.5, 0},
  urlcolor=[rgb]{0.5, 0, 0}
}

\usepackage[breakable]{tcolorbox}
\newtcolorbox{reduction}[2][]
{
  colframe = gray!50,
  colback  = gray!10,
  coltitle = gray!10!black,
  before skip = 10pt,
  after skip = 10pt,
  title    = \textbf{#2},
  #1,
  breakable,
}

\newtcolorbox{game}[2][]
{
  colframe = blue!50,
  colback  = blue!10,
  coltitle = blue!10!black,
  before skip = 10pt,
  after skip = 10pt,
  title    = \textbf{#2},
  #1,
  breakable,
}

\newtcolorbox{examplebox}[2][]
{
  breakable,
  colframe = gray!50,
  colback  = gray!10,
  coltitle = gray!10!black,
  before skip = 10pt,
  after skip = 10pt,
  title    = \textbf{#2},
  #1,
}

\usepackage{algorithmicx}
\usepackage{algorithm}
\usepackage[noend]{algpseudocode}

\algnewcommand\algorithmicinput{\textbf{Input:}}
\algnewcommand\Input{\item[\algorithmicinput]}

\algnewcommand\algorithmicoutput{\textbf{Output:}}
\algnewcommand\Output{\item[\algorithmicoutput]}

\usepackage{amsthm}
\usepackage{thmtools,thm-restate}

\newcommand{\Exp}{\mathop{\mathbb{E}}}

\newcommand{\cM}{\mathcal{M}}
\newcommand{\cN}{\mathcal{N}}
\newcommand{\cX}{\mathcal{X}}
\newcommand{\cY}{\mathcal{Y}}

\newcommand{\bS}{\mathbf{S}}

\newcommand{\bh}{\mathbf{h}}
\newcommand{\bs}{\mathbf{s}}
\newcommand{\bt}{\mathbf{t}}

\newcommand{\bv}{\mathbf{v}}

\newcommand{\bx}{\mathbf{x}}
\newcommand{\by}{\mathbf{y}}
\newcommand{\bz}{\mathbf{z}}

\newcommand{\Real}{\mathbb{R}}

\newcommand{\proj}{\textsf{proj}}
\newcommand{\conv}{\textsf{Hull}}

\newcommand{\crit}{\textsf{crit}}

\usepackage{etoolbox}
\makeatletter
\patchcmd{\hyper@makecurrent}{%
    \ifx\Hy@param\Hy@chapterstring
        \let\Hy@param\Hy@chapapp
    \fi
}{%
    \iftoggle{inappendix}{
        \@checkappendixparam{chapter}%
        \@checkappendixparam{section}%
        \@checkappendixparam{subsection}%
        \@checkappendixparam{subsubsection}%
        \@checkappendixparam{paragraph}%
        \@checkappendixparam{subparagraph}%
    }{}%
}{}{\errmessage{failed to patch}}

\newcommand*{\@checkappendixparam}[1]{%
    \def\@checkappendixparamtmp{#1}%
    \ifx\Hy@param\@checkappendixparamtmp
        \let\Hy@param\Hy@appendixstring
    \fi
}
\makeatletter

\newtoggle{inappendix}
\togglefalse{inappendix}

\apptocmd{\appendix}{\toggletrue{inappendix}}{}{\errmessage{failed to patch}}

\usepackage{tikz}

\title{Two Speeds of Learning: A Representation-Readout Decomposition of Grokking and Double Descent}

\usepackage{authblk}
\author[1,2,3]{Chi-Ning Chou\thanks{These authors contributed equally as first authors.}}
\author[4]{Oscar Uzdelewicz$^*$}
\author[2]{Neng-Chun Chiu}
\author[1]{\protect\\Yao-Yuan Yang}
\author[1,2,3,4]{SueYeon Chung \thanks{Correspondence: \texttt{cchou@flatironinstitute.org}, \texttt{sueyeonchung@g.harvard.edu}}}
\affil[1]{Center for Computational Neuroscience, Flatiron Institute, New York, NY, USA.}
\affil[2]{Department of Physics, Harvard University, Cambridge, MA, USA.}
\affil[3]{Kempner Institute, Harvard University, Allston, MA, USA.}
\affil[4]{New York University, New York, NY, USA.}
\begin{document}
\date{}
\maketitle

\vspace{-6mm}
\begin{abstract}
Training loss and accuracy are the standard signals used to monitor generalization during deep neural network training. Two well-documented phenomena complicate this picture: in grokking, train loss falls rapidly while test performance improves abruptly only after a long delay; in epoch-wise double descent, train loss decreases monotonically while test loss or error rises and falls. Existing accounts are often task-specific, and a task-agnostic analysis framework for diagnosing and explaining these phenomena across realistic tasks and architectures is missing. We address this challenge by analyzing two competing processes that underlie learning dynamics: representation learning in the encoder and readout calibration in the final classifier. Using tools from representational geometry, neural tangent kernels, and linear probing, we show that both processes are active throughout training, with the fluctuations of their relative speed giving rise to seemingly anomalous generalization dynamics. Applying the representation-readout decomposition to grokking across a wide range of tasks and architectures, we find that the readout is train-biased before grokking onset, and representation learning is gradual but not absent—contrary to the lazy-to-rich account. The framework further provides diagnostic signatures distinguishing spurious from genuine generalization: in a previously reported MNIST grokking example and an epoch-wise double descent example, apparent delayed or non-monotone generalization is shown to arise from representation degradation and readout misalignment induced by non-standard training recipes. Together, these results establish the representation-readout decomposition as a top-down framework for understanding learning dynamics and revealing underlying algorithms for interpretability research.
\end{abstract}

\renewcommand{\thefootnote}{\fnsymbol{footnote}}
\renewcommand{\thefootnote}{\arabic{footnote}}

\def\arxiv{1}

\section{Introduction}
\begin{figure}[h!]
    \centering
    \includegraphics[width=\linewidth]{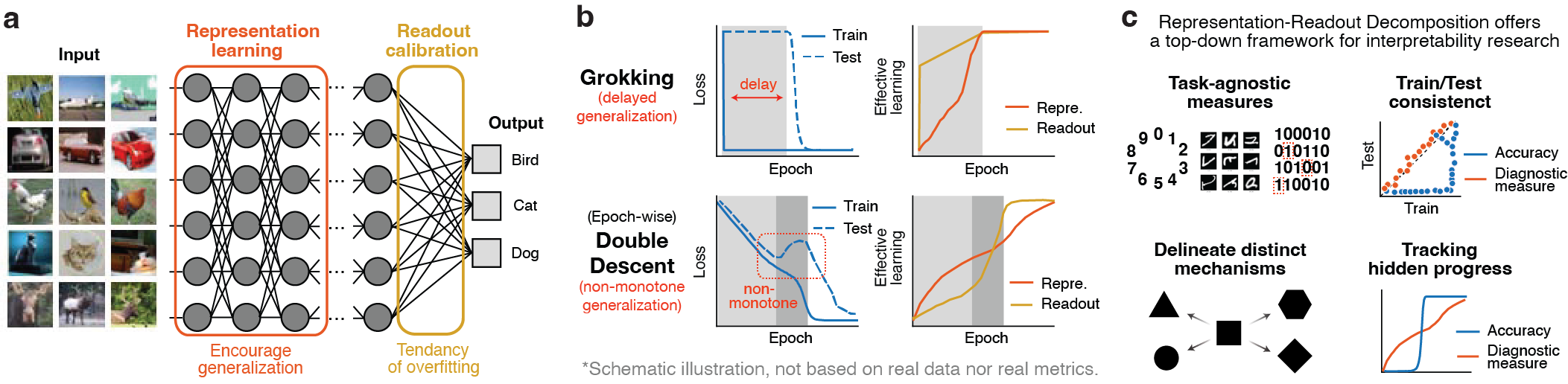}
    \caption{\small\textbf{Overview of our framework.}
    \textbf{a}. A neural network can be conceptually decomposed into two parts: a \textit{feature learning} part (orange) that encourages generalization, and a \textit{readout calibration} part (yellow) that is prone to overfitting (see~\autoref{sec:methods} for details).
    \textbf{b}. We propose that grokking (top) and epoch-wise double descent (bottom) can be explained by the decoupling between the dynamics of representation learning and readout calibration. 
    \textbf{c}. Our proposed representation-readout decomposition offers an analysis framework for future interpretability research.}
    \label{fig:fig1}
    \ifnum\arxiv=0
    \vspace{-5mm}
    \fi
\end{figure}

Training loss and accuracy are the standard signals used to monitor progress during deep neural network training~\citep{goodfellow2016deep}. In practice, we monitor these signals to assess whether a model is properly learning, decide when to stop training, and forecast eventual test performance. Two well-documented phenomena complicate this picture. In \emph{grokking}~\citep{power2022grokking,liu2022omnigrok,humayun2024deep,wang2024grokkedtransformersimplicitreasoners}, a model first achieves near-perfect performance on the training objective while test accuracy remains near chance level, and only much later does generalization appear to emerge abruptly (\autoref{fig:fig1}b, top). In \emph{epoch-wise double descent}~\citep{belkin2019reconciling,nakkiran2021deep}, train loss decreases monotonically while test loss rises and falls non-monotonically over the course of training (\autoref{fig:fig1}b, bottom).

These phenomena raise two linked concerns. Practically, they undermine the standard signals used to monitor training and forecast generalization. Scientifically, they have been interpreted as evidence of phase transitions or other qualitative shifts in learning dynamics. Much existing work addresses these phenomena case by case, with task-specific progress measures for particular grokking settings~\citep{barak2022hidden,nanda2023progress} and tailored mechanistic accounts for individual phenomena~\citep{liu2022omnigrok,kumar2023grokking}. Addressing both concerns calls for a unified, task-agnostic framework with reliable observables for monitoring generalization across diverse tasks and architectures.

An exciting recent task-agnostic account for grokking is the kernel-based \textit{lazy-to-rich} framework~\citep{kumar2023grokking}, which attributes grokking to a transition from the lazy regime—where features are essentially fixed, and readout does the heavy-lifting—to the rich regime—where features actively reorganize for better generalization. 
In parallel, another recent line of work~\citep{chou2025featurelearninglazyrichdichotomy,kirsanov2025,zheng2024delays} uses representation geometry to reveal properties of learning dynamics invisible to kernel-based measures.
Building on these two perspectives, we propose to track the learning dynamics through an explicit decomposition of the network into an \textit{encoder} $\phi_\theta$ that maps inputs to representations and a \textit{readout matrix} $W$ that maps representations to logits (\autoref{fig:fig1}a).
The representation-readout decomposition enables new insights into two speeds\footnote{\label{footnote:speed}Speed here refers to relative rate of progress over training, inferred from the diagnostic measures in~\autoref{sec:methods measures}.} of learning in grokking and double descent:(\autoref{fig:fig1}b)—regarding their underlying principles (\autoref{sec:grokking}) and spurious modes (\autoref{sec:spurious learning})—and establishes a top-down\footnote{\label{footnote:top-down}Here, top-down refers to characterizing the global structure of representations, e.g., through manifold geometry, separability, and alignment, as opposed to bottom-up approaches such as reverse-engineering circuits or features.} framework for studying learning dynamics that can enrich future mechanistic interpretability research (\autoref{fig:fig1}c).

\ifnum\arxiv=0
\vspace{-1.5mm}
\fi
\subsection{Our contributions}
\ifnum\arxiv=0
\vspace{-1.5mm}
\fi

\textbf{A representation-readout decomposition for learning dynamics.} We propose to study the learning dynamics of deep neural networks through an explicit decomposition into an encoder and a readout, tracking the two separately throughout training (\autoref{sec:methods}). The diagnostic measures we use—critical dimension, task-relevant geometric measures~\citep{chou2025glue,chou2025featurelearninglazyrichdichotomy,chung2018classification}, linear probing~\citep{alain2016understanding}, and Neural Tangent Kernel~\citep{jacot2018neural}—are existing tools, but their joint use within the decomposition yields a new vocabulary for describing learning dynamics, going beyond the binary lazy/rich characterization of~\citep{kumar2023grokking}.

\ifnum\arxiv=1
\paragraph{Two speeds of learning in grokking and double descent.} 
\else
\textbf{Two speeds of learning in grokking and double descent.} 
\fi
Applied to canonical grokking examples (e.g., modular addition~\citep{power2022grokking,nanda2023progress, gromov2023grokkingmodulararithmetic}, permutation composition~\citep{chughtai2023toymodeluniversalityreverse,stander2024grokkinggroupmultiplicationcosets}, and sparse parity~\citep{barak2022hidden,merrill2023tale}), our analysis refines the previous lazy-to-rich account: representation learning is not silent in early epochs, it is just slower than the speed$^{\ref{footnote:speed}}$ of readout calibration (\autoref{sec:grokking}). Moreover, the underlying improvement in representation is gradual and persistent rather than sudden.  Moreover, the diagnostic measures discriminate genuine grokking phenomena from spurious cases: a previously reported grokking example on MNIST~\citep{liu2022omnigrok} is revealed as readout artifacts, in which the apparent delayed generalization arises because of representation quality degrading in early epochs under ill-posed training setups (\autoref{sec:results spurious grokking}). The same analysis sheds light on epoch-wise double descent (\autoref{sec:double descent}).

\ifnum\arxiv=1
\paragraph{A top-down analysis framework for interpretability research.} 
\else
\textbf{A top-down analysis framework for interpretability research.} 
\fi
The decomposition and its diagnostic measures offer a top-down$^{\ref{footnote:top-down}}$ analysis framework that advances existing bottom-up approaches to mechanistic interpretability research. Unlike circuit-level analyses, our framework is task-agnostic: the diagnostic measures are consistent across train and test splits, track generalization reliably throughout training, and the geometric vocabulary they provide can describe and discriminate qualitatively different mechanisms at a structural level (\autoref{fig:fig1}c and~\autoref{fig:result geometry}). We anticipate that this framework will be useful for identifying generalization failures, diagnosing unsafe or spurious learning dynamics, and guiding mechanistic investigations in more complex and safety-critical settings.

\subsection{Related work}\label{sec:related work}
\ifnum\arxiv=1
\paragraph{Mechanistic explanations of grokking and double descent.}
\else
\vspace{-1mm}
\textbf{Mechanistic explanations of grokking and double descent.}
\fi
The phenomenon of grokking was first reported by Power et al.~\citep{power2022grokking}, who observed delayed generalization in modular arithmetic tasks despite rapid convergence of training loss. Subsequent work has demonstrated grokking-like dynamics across a range of synthetic and real-world settings by varying weight decay value and/or scaling up the loss value~\citep{liu2022omnigrok}. A substantial body of research has sought mechanistic explanations for this behavior. Analytical studies of simplified models, including sparse parity and Fourier-based analyses~\citep{barak2022hidden,nanda2023progress}, characterize regimes in which delayed generalization arises from the interaction between optimization and representational structure. Empirical investigations have connected grokking to feature learning dynamics~\citep{kumar2023grokking}, circuit efficiency and competition~\citep{varma2023explaining,merrill2023tale}, representational geometry~\citep{zheng2024delays}, and evolving complexity or robustness properties~\citep{humayun2024deep,demoss2024complexity}. Double descent was introduced by Belkin et al.,~\citep{belkin2019reconciling} in the model-wise setting and characterized in the epoch-wise setting by Nakkiran et al.,~\citep{nakkiran2021deep} through introducing label noises, and subsequent analyses have attributed it to slower learning dynamics~\citep{stephenson2021and}. 
Together, these works illuminate when and how grokking and double descent can occur by analyzing changes in the underlying learning dynamics and representations. See Appendix \ref{app:lazy-rich disc},\ref{app:task-specific disc},\ref{app:weight-based disc} for more discussion.

\ifnum\arxiv=1
\paragraph{Progress measures for grokking.}
\else
\textbf{Progress measures for grokking.}
\fi
Several studies have proposed specific progress measures to track generalization in grokking regimes, including Fourier-based~\citep{barak2022hidden,nanda2023progress}, weight-based~\citep{liu2022omnigrok}, kernel-based~\citep{kumar2023grokking}, and representation-based~\citep{golechha2024progress,zheng2024delays} progress measures. 
In contrast, the measures used in our analysis framework go beyond tracking progress toward generalization: they diagnose the state of the learning dynamics and reveal distinct underlying forces such as representation degradation and readout misalignment. This is why we refer to them as \textit{diagnostic} rather than \textit{progress} measures.
See~\autoref{app:other measure disc} for more discussion.

\ifnum\arxiv=1
\paragraph{Lazy-to-rich account for grokking.}
\else
\textbf{Lazy-to-rich account for grokking.}
\fi
The lazy and rich training regimes are grounded in the Neural Tangent Kernel (NTK) theory~\citep{jacot2020neuraltangentkernelconvergence,chizat2019lazy}: in the lazy regime, the network behaves as a kernel method with fixed initial features; in the rich regime, features actively reorganize beyond the linearization.
\ifnum\arxiv=1
\cite{kumar2023grokking}
\else
Kumar et al.~\citep{kumar2023grokking} 
\fi
propose that grokking arises from a transition between these two regimes, identifying two key determinants: a laziness parameter that controls the rate of feature learning by scaling the network output, and the alignment between the initial NTK and the task labels. They develop this theory in the setting of polynomial regression with a two-layer MLP and vanilla gradient descent.
The scale factor view also mathematically coincides with the other approach to induce grokking via large initial weight norm~\citep{liu2022omnigrok}.
Our work builds directly on this account but refines it: rather than an abrupt binary transition, we show that representation learning is slow but continuously active throughout training (\autoref{sec:results beyond lazy-to-rich}), and that the readout remains systematically train-biased until representations are rich enough to support generalization (\autoref{sec:result readout direction}). See~\autoref{app:lazy-rich disc} for more discussion.


\section{Methods}\label{sec:methods}
In this section, we formally set up the representation-readout decomposition (\autoref{sec:methods decomposition}), the task-agnostic measures used in the analysis (\autoref{sec:methods measures}), and the experiments (\autoref{sec:methods tasks and models}).

\subsection{Representation-Readout Decomposition}
\label{sec:methods decomposition}
We consider a neural network as a function $f: \cX \to \Real^C$, where $\cX$ is the input domain (e.g., the pixel space of natural images, or the group $\mathbb{Z}/n\mathbb{Z}$ for modular arithmetic) and $C$ is the dimension of the output space (e.g., the number of classes/labels for classification). Outputs are interpreted as logits, with prediction $\arg\max_i f_i(\bx)$. See~\autoref{sec:methods tasks and models} for the specific tasks and architectures studied in this paper, and see Appendix \ref{app:tasks and architectures} for full details.

\begin{figure}[ht]
    \centering
    \includegraphics[width=\linewidth]{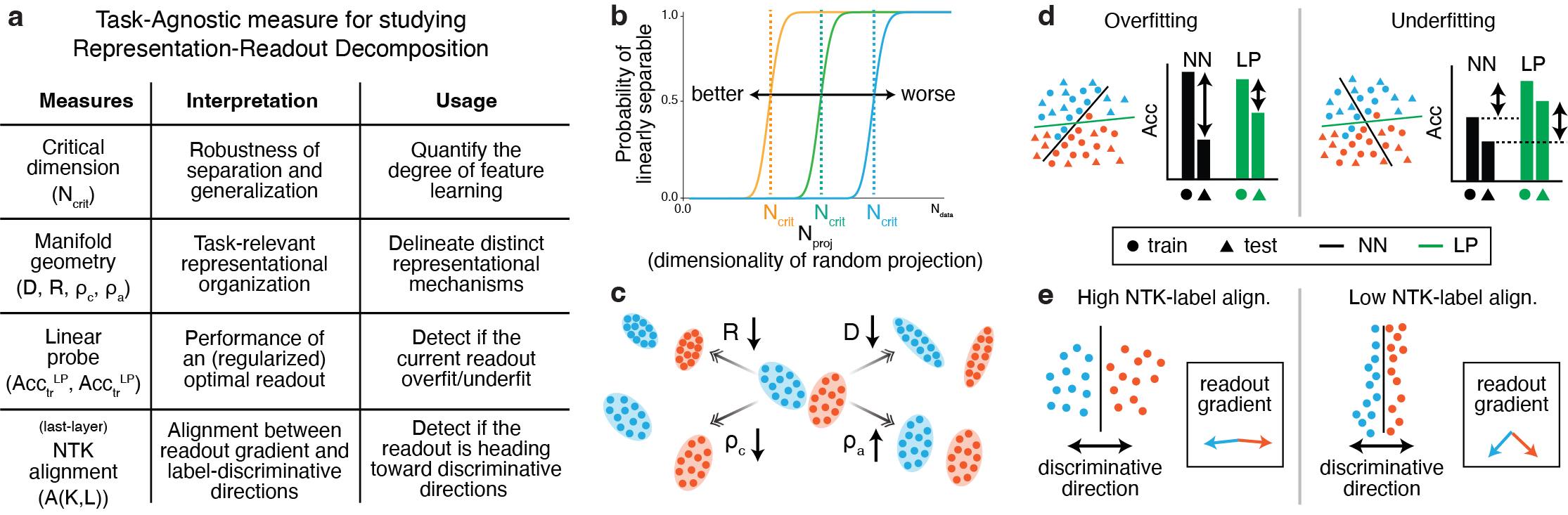}
    \caption{\small\textbf{Task-agnostic diagnostic measures for the representation-readout decomposition.}
    \textbf{a}, Summary of the four measures, their interpretations, and their diagnostic usage.
    \textbf{b}, The critical dimension $N_\crit$ is defined as the median threshold of the random-projection separability probability $p(N_\proj)$; a smaller $N_\crit$ indicates more robustly untangled manifolds.
    \textbf{c}, Four task-relevant geometric measures characterize the underlying geometric mechanisms for separating/untangling label-conditioned representation manifolds $\{\cM_c\}$.
    \textbf{d}, The linear-probe gap compares the best regularized linear readout to the actual readout.
    \textbf{e}, Last-layer NTK alignment measures whether the readout's gradient updates point along label-discriminative directions.}
    \label{fig:measures}
    \vspace{-4mm}
\end{figure}

We decompose $f$ as the composition of a nonlinear \emph{encoder} $\phi_\theta$ and a linear \emph{readout} $W$:
\begin{equation}
f(\bx)=W\,\phi_\theta(\bx), 
\qquad \phi_\theta: \cX \to \Real^N, 
\quad W \in \Real^{C \times N}.
\label{eq:decomposition}
\end{equation}
Here $\phi_\theta$ folds all upstream parameters $\theta$ into a single map to an $N$-dimensional representation, and $W$ is the final linear layer mapping representations to logits.\footnote{Bias terms are absorbed into $W$ via an appended constant feature, so we write $W\phi_\theta(\bx)$ rather than $W\phi_\theta(\bx) + b$. Also, post-readout non-linearity (e.g., softmax) is omitted for simplicity and would not affect the analysis.}

Our choice to decompose the network at the final linear layer is deliberate, motivated by the elegance and generality it affords the analysis. We adopt it because the linear readout is a canonical architectural element shared across MLPs, CNNs, and transformers, and because it allows us to define task-agnostic measures to probe learning dynamics (\autoref{sec:methods measures}). While one could in principle decompose at intermediate layers, we defer a systematic exploration of such choices to future work. 
See also~\autoref{fig:fig1}a for a schematic illustration.
Below are examples of representation-readout decompositions for MLP and Transformer architectures.

\ifnum\arxiv=1
\paragraph{MLP example.}
\else
\textbf{MLP example.}
\fi
For an $L$-layer MLP with the penultimate layer width $N$ and element-wise nonlinearity $\sigma$, set $\bh_0 = \bx$ and $\bh_\ell=\sigma(W_\ell \bh_{\ell - 1})$, where $\ell = 1, \ldots, L-1$.
The decomposition is then $\phi_\theta(\bx)=\bh_{L-1}$ and $W=W_L$, with $\theta = \{W_\ell\}_{\ell=1}^{L-1}$. The encoder is the network up to the penultimate layer's activations; the readout is the final linear layer.

\ifnum\arxiv=1
\paragraph{Transformer example.}
\else
\textbf{Transformer example.}
\fi
For a transformer with $L$ attention-plus-MLP blocks, the decomposition takes $\phi_\theta(\bx) = \mathrm{Pool}(\bh_L)$, where $\bh_L$ is the final hidden state and $\mathrm{Pool}$ extracts a single vector (e.g., the last-position token), and $W = U$ is the unembedding matrix.

\subsection{Task-agnostic diagnostic measures}\label{sec:methods measures}
The representation-readout decomposition (\autoref{sec:methods decomposition},~\autoref{fig:fig1}a) lets us track two sources of learning dynamics separately: \textbf{representation learning}, where the encoder $\phi_\theta$ changes to encourage learning generalizable features, and \textbf{readout calibration}, where the readout $W$ changes to exploit the current representation for label prediction, independent of whether that representation generalizes.
To analyze the effect of these two competing processes, we introduce diagnostic measures applied to the embedding vectors $\phi_\theta(\bx)$ and the readout matrix $W$ during training.


For a task with $C$ output labels, group inputs by their label $y(\bx) \in \{0, \ldots, C-1\}$ as $\cX_c := \{\bx \in \cX : y(\bx) = c\}$, and group their embeddings as $\cM_c := \{\phi_\theta(\bx) : \bx \in \cX_c\} \subset \Real^N$. Although $\cM_c$ lives in $\Real^N$, an extensive literature in machine learning~\citep{bengio2013representation,fefferman2016testing,ansuini2019intrinsic,cohen2020separability,valeriani2023geometry} and computational neuroscience~\citep{cunningham2014dimensionality,chung2021neural} reports that these sets are highly structured and locally low-dimensional, and $\cM_c$ is often referred to as the \textit{representation manifold} conditioned on label $c$.\footnote{Also known as object manifolds, neural manifolds, or category manifolds.}

We emphasize that the \textbf{diagnostic measures} introduced below are simultaneously task-relevant and task-agnostic. They are task-relevant in that they operate on the label-conditioned manifolds $\{\cM_c\}_{c=1}^C$. They are task-agnostic in that they exploit no further structure—such as the algebraic properties in modular arithmetic~\citep{nanda2023progress} (see~\autoref{app:task-specific} for preliminary task-specific analyses)—of the task.
We view this as a starting point, and defer a systematic extension to future work.
See~\autoref{fig:measures} and below for an overview of these measures and~\autoref{app:measures} for detailed definitions and interpretations.




\ifnum\arxiv=1
\paragraph{Measure 1} (Critical dimension $N_\crit$). 
\else
\textbf{Measure 1} (Critical dimension $N_\crit$). 
\fi
The critical dimension $N_\crit$ is defined as the expected smallest projection dimension preserving separability~\citep{cohen2020separability,chou2025glue} and is related to manifold capacity theory~\citep{chung2018classification,wakhloo2023linear,chou2025glue}. 
Roughly speaking, $N_\crit$ is the smallest dimensionality $N_\proj$ such that the label-conditioned manifolds $\{\cM_c\}$ are linearly separable with probability at least $0.5$ after being randomly projected to an $N_\proj$-dimensional subspace (\autoref{fig:measures}b, definitions in~\autoref{app:implementation critical dimension}).
A small $N_\crit$ indicates that the representation manifolds are well \textit{untangled}~\citep{dicarlo2007untangling,chou2025glue}—their linear separability survives aggressive random projection (\autoref{fig:crit dim toy2}). 
Crucially, $N_\crit$ quantifies the robustness of linear separability in that a smaller value indicates a more generalizable representation~\citep{chou2025featurelearninglazyrichdichotomy}.
See~\autoref{app:critical dimension and GLUE} for details

\ifnum\arxiv=1
\paragraph{Measure 2} (Task-relevant manifold geometric measures $D,R,\rho_c, \rho_a$). 
\else
\textbf{Measure 2} (Task-relevant manifold geometric measures $D,R,\rho_c, \rho_a$). 
\fi
The GLUE theory~\citep{chou2025glue} also derived a small set of geometric quantities that are analytically linked to the value of $N_\crit$: the \textit{effective manifold dimension} $D$, \textit{effective manifold radius} $R$, and the center and axis \textit{manifold alignment} $\rho_c,\rho_a$ between pairs of representational manifolds (\autoref{fig:measures}c, definitions in~\autoref{app:implementation GLUE geometry}). 
The advantage of these GLUE geometric measures is that they are analytically linked to $N_\crit$ and can better track the task-relevant geometric changes underlying mechanisms of representational robustness~\citep{chou2025glue,chou2025featurelearninglazyrichdichotomy,choudiagnosing}. 
As the definitions of these geometric measures involve an elaborate introduction of the concept of anchor point distribution~\citep{chou2025glue}, we defer the details to~\autoref{app:glue geometric measures}. Previous work using GLUE geometric measures to better understand models' behaviors is discussed in \autoref{app:GLUE disc}. 



\ifnum\arxiv=1
\paragraph{Measure 3} (Linear probe accuracy). 
\else
\textbf{Measure 3} (Linear probe accuracy). 
\fi
We fit a regularized linear probe~\citep{alain2016understanding} on $\phi_\theta(\bx)$ using the same train/test split as the model and compare its accuracy to the model's. The probe identifies the best regularized linear readout on top of the current representation, providing a reference point that isolates encoder quality from the actual readout: a mismatch between probe and model accuracy signals that the readout is mis-calibrated relative to what $\phi_\theta$ affords (\autoref{fig:measures}d). 
See~\autoref{app:linear probe} for details.


\ifnum\arxiv=1
\paragraph{Measure 4} (Last-layer NTK-label alignment).
\else\textbf{Measure 4} (Last-layer NTK-label alignment).
\fi
We probe how well the current feature space supports linear readout learning via the alignment between the last-layer neural tangent kernel (NTK)~\citep{jacot2020neuraltangentkernelconvergence} and the label kernel (i.e., similarity matrix of the labels, see definitions in~\autoref{app:implementation NTK}). 
This measure, abbreviated NTK alignment, quantifies how closely the feature geometry mirrors the label structure from the readout's perspective: higher alignment indicates that gradient flow on $W$ makes faster progress toward label-aligned solutions (\autoref{fig:measures}e,~\autoref{app:NTK readout learning}).
We evaluate alignment on both train and test splits at each checkpoint, allowing us to compare how the feature geometry develops on seen versus unseen data.
See more details in~\autoref{app:ntk}.

In summary, $N_\text{crit}$ serves as a scalar summary of representational robustness, and the GLUE geometric measures provide a finer-grained probe of the geometric mechanisms supporting separability. The gap between linear probe and model accuracy diagnoses readout overfitting or underfitting, while the last-layer NTK-label alignment measures how effectively the current feature geometry directs readout gradients toward label-discriminative solutions (\autoref{fig:measures}a).


\vspace{-2mm}
\subsection{Tasks and models}\label{sec:methods tasks and models}
For grokking, there are three algorithmic tasks with known mechanistic solutions: modular addition~\citep{power2022grokking,nanda2023progress, gromov2023grokkingmodulararithmetic}, permutation composition~\citep{chughtai2023toymodeluniversalityreverse,stander2024grokkinggroupmultiplicationcosets}, and sparse parity~\citep{barak2022hidden,merrill2023tale}, which allow us to test our analysis framework for a task-agnostic explanation for grokking.
For each task, we train both an MLP and a transformer where applicable. Full data, optimizer, and hyperparameter details are deferred to~\autoref{app:tasks and architectures}. Below, we briefly specify the specifications of each task.

\ifnum\arxiv=1
\vspace{2mm}
\noindent\textbf{Modular addition.}
\else
\textbf{Modular addition.}
\fi
Predict $(a + b) \bmod p$ for $p = 113$, given $a$ and $b$ as inputs ($C=113$ classes). 

\ifnum\arxiv=1
\vspace{2mm}
\noindent\textbf{Permutation composition.}
\else
\textbf{Permutation composition.}
\fi
Predict $\sigma \circ \tau$ in the symmetric group $S_n$ for $n=5$, given the indices of 
$\sigma$ and $\tau$ as inputs ($C = 120$ classes).

\ifnum\arxiv=1
\vspace{2mm}
\noindent\textbf{Sparse parity.} 
\else
\textbf{Sparse parity.} 
\fi
Predict the parity of a fixed unknown subset of $k = 3$ coordinates of an input $x \in \{-1, +1\}^{n}$ with $n = 40$ ($C = 2$ classes). 

\ifnum\arxiv=1
\vspace{2mm}
\fi
We also study a fourth grokking example on the small-sample MNIST setup of Liu et al.~\citep{liu2022omnigrok}, whose delayed-generalization phenomenon the framework will diagnose as a readout artifact (\autoref{sec:results spurious grokking}). 

\ifnum\arxiv=0
\vspace{-2mm}
\fi
\section{Hidden Representation Learning and Biased Readout Under Grokking}\label{sec:grokking}
\ifnum\arxiv=0
\vspace{-2mm}
\fi
We argue that grokking reflects two speeds of learning operating simultaneously: a slow, continuous improvement in representation quality (\autoref{sec:results beyond lazy-to-rich}), and a faster readout calibration that outpaces the encoder and locks onto training data before generalizable representations are ready (\autoref{sec:result readout direction}).

\begin{figure}[ht]
    \centering
    \includegraphics[width=0.9\linewidth]{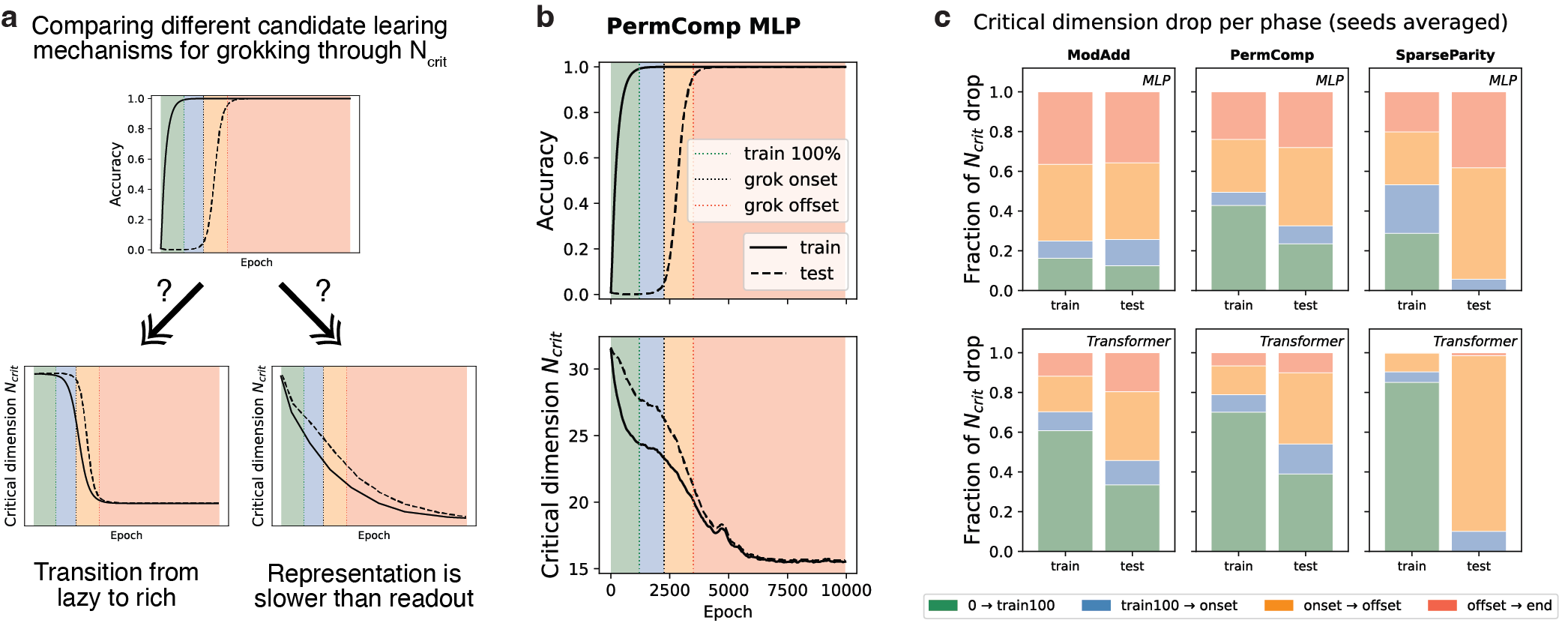}
    \caption{\small\textbf{Representation learning is slow but not absent before grokking.}
    \textbf{a}, Using critical dimension $N_\crit$ to compare the candidate mechanism of lazy-to-rich transition (bottom left) and slower-representation-learning (bottom right).
    \textbf{b}, A grokking example. Top: model's accuracy curves. Bottom: $N_\crit$ decreases throughout the training. 
    \textbf{c}, Results across tasks and architectures (averaged over 3 seeds, all results in~\autoref{app:all grok}).
    See~\autoref{app:event detection} for the definition of the shaded areas.
    }
    \label{fig:result grokking beyond lazy rich}
    \vspace{-3mm}
\end{figure}

\subsection{Representation learning is slow but not absent before grokking}\label{sec:results beyond lazy-to-rich}
The lazy-to-rich account~\citep{kumar2023grokking} (\autoref{app:lazy-rich disc}) attributes grokking to a transition from a \textit{lazy} regime—where features are approximately fixed and the readout does the heavy lifting—to a \textit{rich} regime—where features actively reorganize for better generalization. Under a strict reading of this account, representation learning should not begin until grokking onset.
Consequently, critical dimension $N_\crit$—which quantifies the degree of representation learning~\citep{chou2025featurelearninglazyrichdichotomy} (\autoref{app:critical dimension and GLUE})—should stay near its initial value during early epochs and drop sharply at grokking onset (\autoref{fig:result grokking beyond lazy rich}a, bottom left). 
In contrast, if representation learning were slower than readout calibration but still continuously active in early training, then critical dimension $N_\crit$ would already decrease before grokking onset (\autoref{fig:result grokking beyond lazy rich}a, bottom right). 

We compared these two candidate mechanistic explanations by empirically measuring $N_\crit$ throughout training. Take permutation composition on MLP as an example (\autoref{fig:result grokking beyond lazy rich}b), we first color the epochs into four stages according to the model’s accuracy curves: (i) before training accuracy reaches 100\%, (ii) before grokking onset, (iii) before grokking offset, and (iv) after grokking offset. We found that $N_\crit$ decreases gradually and continuously across all four stages (\autoref{fig:result grokking beyond lazy rich}b, bottom).

To quantify this pattern across tasks and architectures, we report in~\autoref{fig:result grokking beyond lazy rich}c  the fraction of the total $N_\crit$ drop accounted for by each phase, averaged over three seeds. In every (task, model) pair, a substantial portion of the drop occurs before grokking onset (i.e., in the green and blue phases combined), ranging from roughly $20\%$-$80\%$ (train) and $5\%$-$50\%$ (test). These findings suggest that representation learning does not begin abruptly at grokking onset—as the strict lazy-to-rich account would predict—rather, it proceeds continuously throughout training. For further discussion on our theory's relation to the lazy-to-rich account, see \autoref{app:lazy-rich disc}.

\vspace{-2mm}
\subsection{Readout learning direction is train-biased before grokking}\label{sec:result readout direction}
Having established that representation learning proceeds continuously throughout training, we now turn to the readout. While the encoder gradually refines the geometry of its representations, the readout has to map those evolving representations to correct outputs for both train and test inputs. A natural question is whether the readout's learning direction is balanced across train and test (generalization), or whether it becomes systematically biased toward the train side (memorization). 

\begin{figure}[ht]
    \centering
    \includegraphics[width=\linewidth]{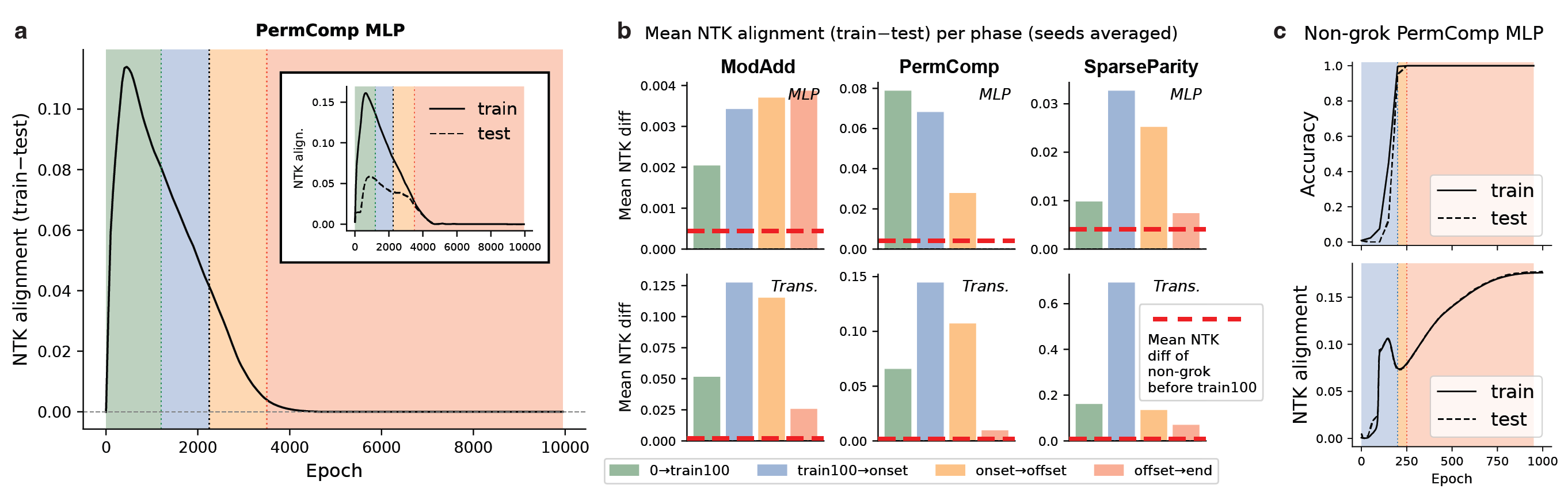}
    \caption{\small\textbf{Readout learning direction is train-biased before grokking.}
    \textbf{a}, NTK alignment difference (train minus test) throughout training for permutation composition on MLP (seed=0). The gap peaks near training saturation and decays to zero as grokking completes. The inset shows raw train (solid) and test (dashed) NTK alignment curves. See~\autoref{fig:result grokking beyond lazy rich}a for the model's accuracy curves.
    \textbf{b}, Mean NTK alignment difference per phase. The dashed lines indicate the max difference in the non-grokking counterparts throughout training, as baseline controls.
    \textbf{c}, Accuracy and NTK-label alignment curves for a non-grok PermComp MLP example.
    See~\autoref{app:event detection} for the definition of the shaded areas.
    }
    \label{fig:result NTK}
    \vspace{-4mm}
\end{figure}

To probe this, we track the NTK alignment difference (train minus test) throughout training. A positive value indicates that the readout is being updated in a direction that preferentially fits training examples over test examples—a signature of overfitting at the readout level (see~\autoref{fig:measures}e for a schematic illustration to interpret the value of NTK alignment, and see~\autoref{app:ntk} for details). As shown for permutation composition on MLP (\autoref{fig:result NTK}a), the train-test NTK alignment gap is large and positive throughout the pre-grokking phases (green and blue), peaks around the time training accuracy saturates (i.e., the end of green area), and then decays toward zero as grokking completes. 

This pattern holds broadly across tasks and architectures (\autoref{fig:result NTK}b). In every (task, model) pair, the mean NTK alignment difference is typically largest in the train100-to-onset phase (blue bars), substantially exceeding the pre-train100 phase (green) and decaying through the onset-to-offset (orange) and post-offset (red) phases. 

\ifnum\arxiv=1
\paragraph{The train-test NTK alignment gap disappears when grokking is removed.}
\else
\textbf{The train-test NTK alignment gap disappears when grokking is removed.}
\fi
As a control, we measure the mean NTK alignment difference in non-grokking counterparts for each setting and find that the gap is substantially smaller (\autoref{fig:result NTK}b, red dashed lines;~\autoref{fig:result NTK}c), confirming that the train-test NTK alignment gap is specific to the grokking regime. Together with the critical dimension results of \autoref{sec:results beyond lazy-to-rich}, this suggests that grokking reflects not a sudden transition in representation learning, but a gradual equalization of readout direction as representation quality catches up.



\ifnum\arxiv=0
\vspace{-1mm}
\fi
\subsection{Probing underlying algorithms via geometric mechanisms of representation learning}\label{sec:geometry}
\ifnum\arxiv=0
\vspace{-1mm}
\fi

\begin{wrapfigure}{r}{0.5\linewidth}
    \centering
    \vspace{-1em}
    \includegraphics[width=\linewidth]{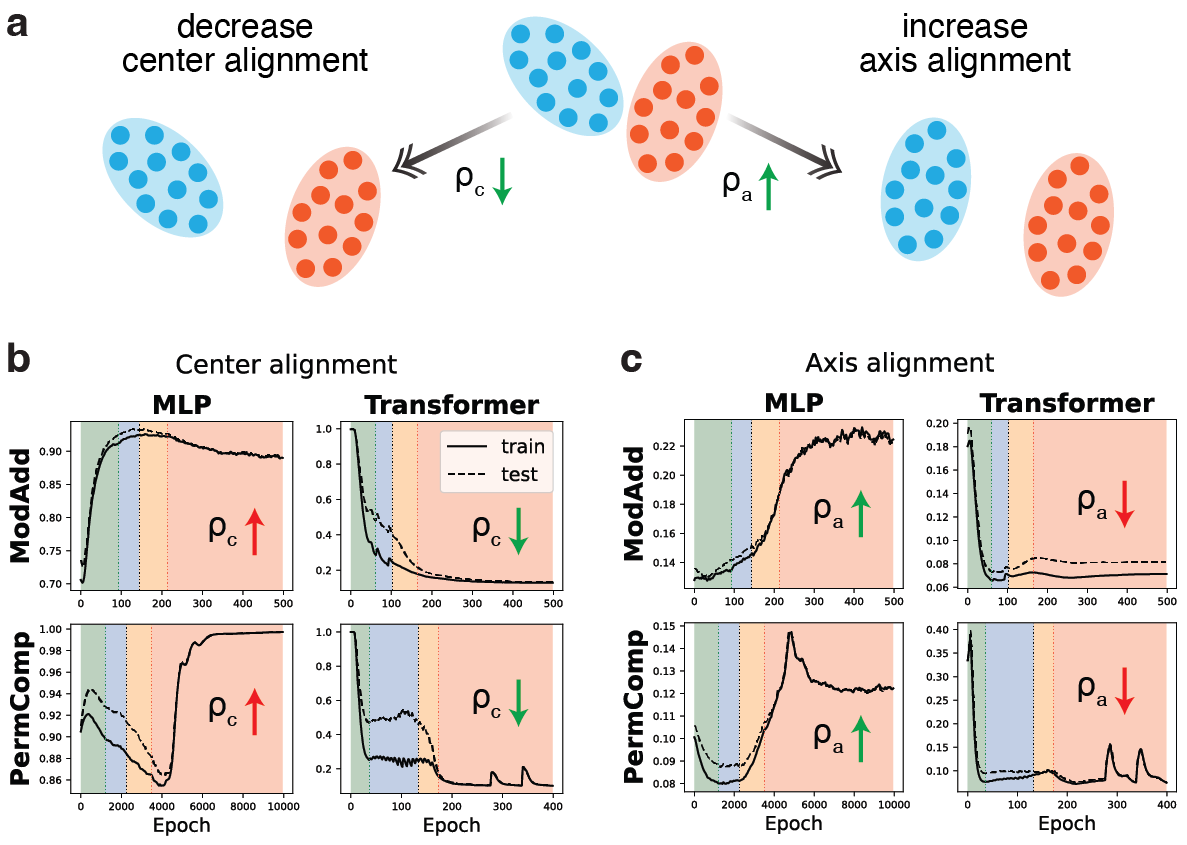}
    \caption{\small Different geometric mechanisms of representation learning in MLPs and Transformers.} 
    \label{fig:result geometry}
    \vspace{-3mm}
\end{wrapfigure}

A common mechanistic interpretability approach is to determine whether a network solves a task via a specific algorithm by identifying underlying algorithmic components~\citep{nanda2023progress}. 
Here we show that studying changes in label-conditioned manifolds offers a complementary, top-down$^{\ref{footnote:top-down}}$ route: the geometric measures (\autoref{fig:measures}c,~\autoref{app:glue geometric measures}) can probe the underlying mechanisms of representation learning and discriminate distinct algorithms implemented by different architectures on the same task.

We identify architecture-dependent differences in the geometric mechanisms underlying representation learning (\autoref{fig:result geometry}). According to GLUE theory~\citep{chou2025glue}, decreasing center alignment $\rho_c$ and increasing axis alignment $\rho_a$ both improve manifold separability (\autoref{fig:result geometry}a). Across all grokking tasks (ModAdd and PermComp in~\autoref{fig:result geometry} and SparseParity in~\autoref{app:all grok}), MLPs increase $\rho_c$ (hurting $N_\text{crit}$) and decrease $\rho_a$ (improving $N_\text{crit}$), while Transformers do the opposite (\autoref{fig:result geometry}b and c)—suggesting that the two architectures achieve representation learning through qualitatively different geometric pathways. Meanwhile, we find that the manifold radius $R$ and dimension $D$ consistently decrease across tasks in both MLPs and Transformers (\autoref{app:all grok}). Together, these results illustrate how the representation-readout decomposition, combined with geometric measures, can discriminate qualitatively different mechanisms that produce the same surface-level grokking behavior, offering a top-down tool for probing the algorithms implemented by neural networks.


\section{Diagnosing Spurious Learning}\label{sec:spurious learning}
Both grokking and epoch-wise double descent have served as playgrounds for understanding how neural networks generalize. The clean and extreme forms of these phenomena typically do not arise in real-world training; researchers deliberately induce them via non-standard recipes, such as decreasing weight decay, reducing training data  \citep{liu2022omnigrok}, scaling down the relative learning speed of hidden layers \citep{kumar2023grokking}, or introducing label noise \citep{nakkiran2021deep}. While studying how to induce these phenomena can be a fruitful approach to building understanding, it can also be misleading if the training recipe strays too far from standard practice. If the ingredients that trigger grokking inadvertently hinder neural network learning, then the scientific conclusions drawn from such settings should be taken with a grain of salt.

\begin{figure}[ht]
    \centering
    \includegraphics[width=\linewidth]{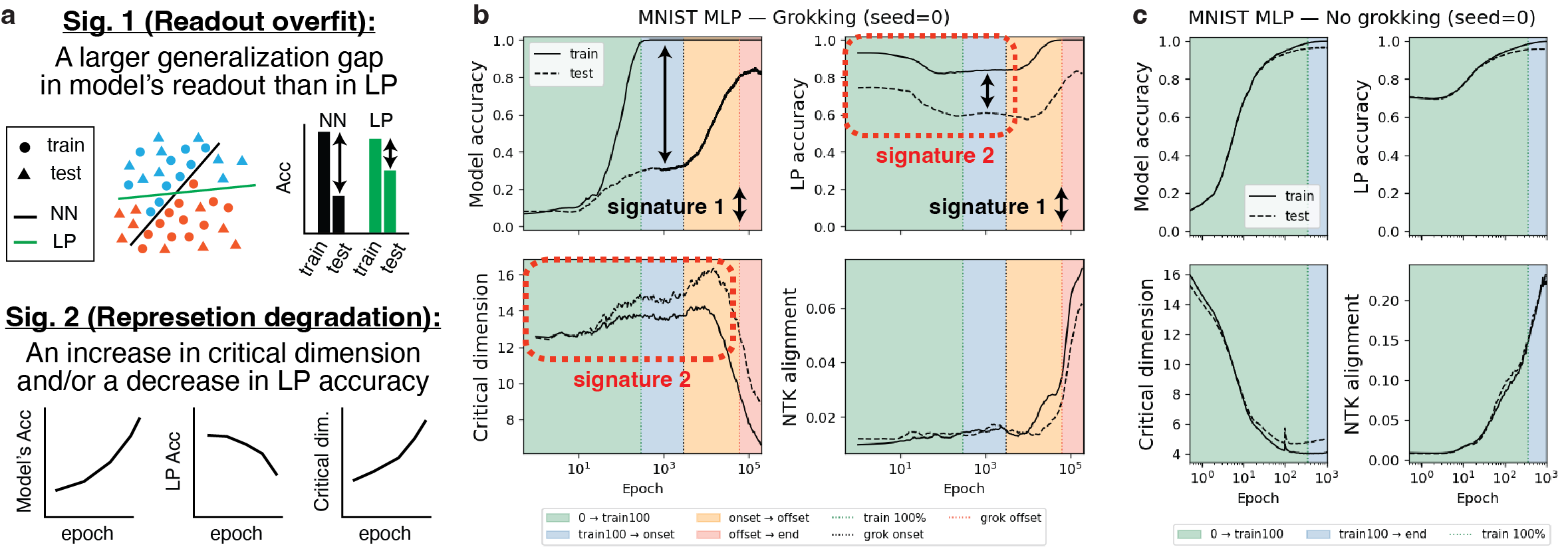}
    \caption{
    Diagnosing spurious learning in an MNIST grokking example.
    See~\autoref{sec:results spurious grokking} for details.
    }
    \label{fig:result spurious grokking}
    \ifnum\arxiv=0
    \vspace{-3mm}
    \fi
\end{figure}

\begin{figure}[ht]
    \centering
    \includegraphics[width=\linewidth]{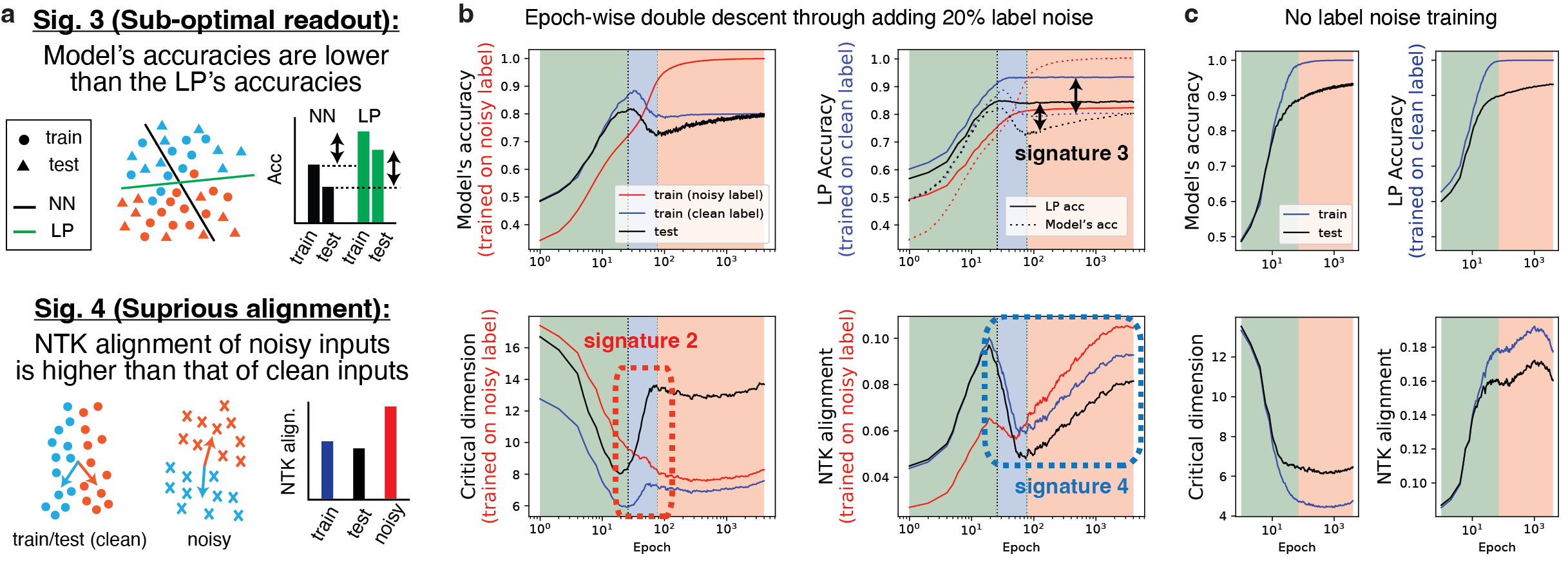}
    \caption{
    Diagnosing spurious learning in a double descent example.
    See~\autoref{sec:double descent} for details.
    }
    \label{fig:result dd}
    \ifnum\arxiv=0
    \vspace{-4mm}
    \fi
\end{figure}

We can flag such cases by identifying spurious learning dynamics through the representation-readout decomposition. Specifically, we identify four \textbf{diagnostic signatures} as follows.
\vspace{-1mm}
\begin{enumerate}[leftmargin=*]
    \item (\textbf{Readout overfit}) The model's generalization gap is much larger than that of the LP, indicating that the readout is over-adapting to training data (\autoref{fig:result spurious grokking}a, top).
    \item (\textbf{Representation degradation}) Decreasing LP accuracy alongside increasing $N_\text{crit}$ indicates that representation quality is actively degrading (\autoref{fig:result spurious grokking}a, bottom).
    \item (\textbf{Sub-optimal readout}) LP accuracy substantially exceeds model accuracy, indicating that the readout is under-performing relative to what the encoder affords (\autoref{fig:result dd}a, top).
    \item (\textbf{Spurious alignment}) The NTK alignment of a spurious signal substantially exceeds that of the true task signal, indicating that readout gradient updates are increasingly driven by spurious rather than task-relevant structure (\autoref{fig:result dd}a, bottom; see also~\autoref{app:ntk}).
\end{enumerate}



\subsection{Spurious grokking on MNIST via high initial weight norm and small training size}\label{sec:results spurious grokking}
We examine a previously reported grokking example on MNIST~\citep{liu2022omnigrok} (\autoref{fig:result spurious grokking}). Notably, the final test accuracy in this example is around $80\%$, while a support vector machine (SVM) can easily achieve near $98\%$ on the same task, suggesting that the training recipe severely limits learning. We use the diagnostic signatures to investigate why.
We find that the first two diagnostic signatures appear in this example (\autoref{fig:result spurious grokking}b). During the pre-grokking epochs (blue region), the model's generalization gap substantially exceeds that of the LP (\autoref{fig:result spurious grokking}b, double-headed arrows), indicating readout overfitting. Moreover, throughout the pre-grokking period (green and blue regions), LP accuracy decreases and $N_\text{crit}$ increases (\autoref{fig:result spurious grokking}b, red boxes), indicating that training is actively degrading representation quality rather than improving it.

\ifnum\arxiv=1
\paragraph{Diagnostic signatures disappear under standard training.} 
\else
\textbf{Diagnostic signatures disappear under standard training.} 
\fi
The grokking recipe of~\citep{liu2022omnigrok} requires multiplying initial weights by a factor greater than four and training on only $1$k of the $60$k available samples. Removing these non-standard components eliminates both the grokking pattern and the diagnostic signatures (\autoref{fig:result spurious grokking}c), confirming that the apparent delayed generalization arises from readout miscalibration on top of degrading representations rather than genuine representation improvement. 

\subsection{Epoch-wise double descent through label noise}\label{sec:double descent}
\ifnum\arxiv=0
\vspace{-1mm}
\fi
Epoch-wise double descent is typically induced by introducing label noise~\citep{nakkiran2021deep} (\autoref{fig:result dd}). We find that diagnostic signatures 2, 3, and 4 (\autoref{fig:result dd}a) appear in this example.
LP accuracy on clean labels exceeds the model's accuracy for both train and test (\autoref{fig:result dd}b, double-headed arrows), indicating a suboptimal readout. During the test accuracy descent (blue region), LP accuracy plateaus, $N_\text{crit}$ increases, and NTK alignment decreases (\autoref{fig:result dd}b, red boxes), indicating deteriorating representation quality and misaligned readout gradients. In the recovery phase (orange region), the NTK alignment of the noisy-label train set surpasses that of the clean-label train and test sets (\autoref{fig:result dd}b, blue boxes), indicating that readout updates become dominated by noisy labels.

\ifnum\arxiv=1
\paragraph{Diagnostic signatures disappear when label noise is removed.} 
\else
\textbf{Diagnostic signatures disappear when label noise is removed.} 
\fi
Removing label noise eliminates both double descent and the diagnostic signatures (\autoref{fig:result dd}c), confirming that both arise from the non-standard training recipe. We conclude that epoch-wise double descent here reflects a two-stage process: representation deterioration in the blue phase, followed by spurious signal domination in the orange phase. These results suggest that the diagnostic signatures generalize beyond grokking to non-monotone generalization more broadly, offering a unified lens for diagnosing generalization failures. See~\autoref{app:dd} for details.

\vspace{-2mm}
\section{Discussion}\label{sec:discussion}
\vspace{-2mm}
We proposed a representation-readout decomposition framework (\autoref{sec:methods decomposition}) for studying learning dynamics in deep neural networks, and applied it to understand grokking and epoch-wise double descent. The framework reveals two speeds$^{\ref{footnote:speed}}$ of learning and yields new insights into the underlying principles (\autoref{sec:results beyond lazy-to-rich},~\autoref{sec:result readout direction}) and geometric mechanisms (\autoref{sec:geometry}) of these phenomena, and provides diagnostic signatures that distinguish spurious from genuine generalization (\autoref{sec:spurious learning}).

The representation-readout decomposition also serves as a top-down$^{\ref{footnote:top-down}}$  complement to existing bottom-up approaches in mechanistic interpretability. 
Circuit-level analyses excel at identifying specific computational structures, but offer limited guidance across diverse tasks and architectures
The decomposition addresses this gap by characterizing how representation quality and readout calibration evolve jointly throughout training, providing principled criteria for where to look and what to expect from circuit-level investigations. 
We close with several directions for future work.

\vspace{-1mm}
\begin{itemize}[leftmargin=*]
\item \textbf{Intermediate-layer analysis.} In this work, the decomposition operates exclusively at the final linear layer. Extending the analysis to intermediate layers is a natural and important direction.
\item \textbf{Task-specific analysis.} Incorporating task-specific structure into the analysis could reveal more geometric signatures of underlying algorithms (see~\autoref{app:task-specific} for preliminary findings).
\item \textbf{Tools for practitioners.} Our diagnostic measures are often consistent across train and test splits (\autoref{app:metric consistency}), making them reliable training monitors. Developing lightweight estimates and guidance for hyperparameter selection would broaden real-world applicability.
\item \textbf{Continual and transfer learning.} 
Investigating encoder stability and readout adaptation in catastrophic forgetting and transfer learning through the lens of our diagnostic signatures.
\item \textbf{Connections to alignment and safety.} The diagnostic signatures for spurious generalization could provide a principled tool for detecting misalignment in safety-critical models before deployment.
\end{itemize}

\subsubsection*{Acknowledgments}
We gratefully acknowledge support from the Center for Computational Neuroscience at the Flatiron Institute and the Simons Foundation. S.C. acknowledges support from a Sloan Research Fellowship, a Klingenstein-Simons Award, and the Aramont Fellowship Fund. The computational work was carried out on the Flatiron Institute’s high-performance computing cluster. Yao-Yuan Yang worked in an advisory capacity.

\bibliography{refs}
\bibliographystyle{apalike}

\newpage
\appendix

\section*{Appendix}
This appendix provides supplementary material organized as follows. 
\begin{itemize}[leftmargin=*]
\item \autoref{app:measures} gives full implementation details and extended theoretical background for the four diagnostic measures used in our analysis: critical dimension, GLUE geometric measures, linear probing, and last-layer NTK-label alignment. 
\item \autoref{app:tasks and architectures} describes the architectures, tasks, and training hyperparameters for all grokking and double descent experiments. 
\item \autoref{app:further discussion} situates our framework within the broader literature, with extended discussions of the lazy-to-rich hypothesis, task-specific progress measures, other task-agnostic measures, weight-based theories of grokking, and previous applications of GLUE theory. 
\item \autoref{app:all measures} provides additional empirical results for all task-model pairs in both grokking and non-grokking settings, as well as the double descent experiments. 
\item \autoref{app:task-specific} presents preliminary task-specific analyses using pairwise critical dimension heatmaps and transfer linear probes.
\item \autoref{app:metric consistency} quantifies the train-test consistency of our diagnostic measures across tasks, architectures, and training stages, showing that representation-based measures track learning dynamics more reliably than standard loss and accuracy curves.
\end{itemize}

\section{Diagnostic Measures for Representation-Readout Decomposition Analysis}\label{app:measures}
In~\autoref{sec:methods} and~\autoref{fig:measures}, we briefly mention the high-level definition and intuition of the four types of measures used in our analysis. In this appendix section, we are going to provide concrete implementation details (\autoref{app:measures implementation}), a more in-depth introduction to each method (\autoref{app:critical dimension and GLUE}, \autoref{app:glue geometric measures}, \autoref{app:linear probe}, and~\autoref{app:ntk}), and discussions on what other possible measures can be used in future research.

\subsection{Implementations of all measures}\label{app:measures implementation}
We begin by providing implementation details for all the measures used in this paper for reproduction. For more contexts, motivations, and theories, please go to the corresponding subsections later.

First, all our measures only depend on the last-layer label-conditioned representations. Recall the notations from~\autoref{sec:methods measures}: for a task with $C$ output labels, group inputs by their label $y(\bx) \in \{0, \ldots, C-1\}$ as $\cX_c := \{\bx \in \cX : y(\bx) = c\}$, and group their embeddings by the encoder $\phi_\theta$ as $\cM_c := \{\phi_\theta(\bx) : \bx \in \cX_c\} \subset \Real^N$. Following the convention in recent literature, $\cM_c$ is referred to as the \textit{representation manifold} conditioned on label $c$. See~\autoref{app:tasks and architectures} for details on the encoder mapping and labels for each task and architecture used in our analysis.

\subsubsection{Critical dimension}\label{app:implementation critical dimension}
Consider a collection of dichotomies over $C$ output labels, i.e., a subset $\cY$ in $\{1,-1,0\}^C$, the critical dimension is defined as follows. When $y_c=0$, it means that the $c$-th label does not participate in the dichotomy.
Mathematically, assume\footnote{Empirically, this is achieved by downsampling the points in each $\cM_c$ to a fixed budget per class and, where indicated, applying a simple kernel.} that $\{\cM_c\}$ is linearly separable with respect to every $\by \in \cY$. For $1 \leq N_\proj \leq N$, let $\Pi_{N_\proj}$ be a random projection from $\Real^N$ to $\Real^{N_\proj}$, and let $p(N_\proj)$ denote the probability that $\{\Pi_{N_\proj}(\cM_c)\}$ remains linearly separable, taken over both the projection draw and a uniform draw of $y$ from $\cY$. By construction, $p(N) = 1$ and $p(1) = 0$.
Now, define $N_\crit := \sum_{N_\proj = 1}^{N} \big(1 - p(N_\proj)\big)$. When $p(\cdot)$ exhibits a symmetric phase transition, $N_\crit$ approximately coincides with the median-threshold dimension $\arg\max_{N_\proj} \{N_\proj : p(N_\proj) \geq 0.5\}$.

Numerically, $N_\crit$ can be estimated by binary search over the value of $p(N_\proj)$. A better (faster and more accurate) way is using an analytical formula derived in the GLUE theory~\citep{chou2025glue}:
\begin{equation}\label{eq:glue formula}
    N_\crit = \Exp_{\substack{\bt\sim\cN(0,I_N)\\ \by\sim\cY}}\left[\min_{\substack{\bv\in\Real^N:\\y_c\cdot\langle\bv,\bz_c\rangle\geq0\\\forall\bz_c\in\cM_c,\forall c}}\|\bv-\bt\|_2^2\right]
\end{equation}
where $\cN(0,I_N)$ denotes the $N$-dimensional isotropic Gaussian distribution, $\by\sim\cY$ denotes uniformly sampling $\by$ from $\cY$ at random, $\langle\cdot,\cdot\rangle$ denotes inner product between two vectors, and $\|\cdot\|_2$ denotes the $\ell_2$ norm of a vector. Note that the minimization inside the expectation is a quadratic program with linear constraints and hence can be efficiently solved by a standard convex optimization solver.

In all our analyses, we sampled 200 random $\bt$ to and picked $\cY$ to be the set of all pairwise dichotomies, and randomly subsampled a subset of 30 points from each label-conditioned manifold to ensure linear separability. See~\autoref{app:all measures} for results on other choices of dichotomy structure.

\subsubsection{GLUE geometric measures}\label{app:implementation GLUE geometry}
It turns out that through the strong duality property in convex optimization, the formula for the critical dimension in~\autoref{eq:glue formula} can be rewritten as follows.
\begin{equation}\label{eq:glue formula dual}
    N_\crit = \Exp_{\substack{\bt\sim\cN(0,I_N)\\ y\sim\cY}}\left[\max_{\substack{\bs_c\in\conv(\cM_c)\\\lambda_c\geq0}}\left(\frac{\langle \bt,\sum_cy_c\lambda_c\bs_c\rangle}{\|\sum_cy_c\lambda_c\bs_c\|_2}\right)^2\right]
\end{equation}
where $\conv(\cM_c)$ denotes the convex hull of $\cM_c$, and $\textsf{span}(\{s_c\})$ denotes the linear subspace spanned by $\{s_c\}$. Note that the maximization inside the expectation is simply the square of the $\ell_2$ norm of the projection of $t$ onto the convex cone spanned by the representation manifolds weighted by their respective labels $y$.

For each $(\by,\bt)$ pair sampled by the expectation operator in~\autoref{eq:glue formula dual}, we can denote $\bs_c(\by,\bt)\in\conv(\cM_c)$ for each $c$ as the optimizer for the inner maximization term. Namely, the distribution of $(\by,\bt)$induces a distribution over each $\conv(\cM_c)$, this is also known as the \textit{anchor point distribution} in GLUE theory~\citep{chou2025glue}. Next, for each label $c$, we can then define the anchor center to be $\bs_{c,0}:=\Exp_{\by,\bt}[\bs_c(\by,\bt)]$. And for each $(\by,\bt)$ pair, we can define the axis part of the anchor point of the $c$-th manifold as $\bs_{c,1}(\by,\bt)=\bs_c(\by,\bt)-\bs_{c,0}$. Finally, denote $\bS_0\in\Real^{C\times N}$ as the matrix with the $c$-th row being $\bs_{c,0}$ and $\bS_1(\by,\bt)\in\Real^{C\times N}$ as the matrix with the $c$-th row being $\bs_{c,1}(\by,\bt)$.

\begin{itemize}
    \item Effective manifold dimension $D$ is defined as
    \begin{equation}\label{eq:effective dimension}
        D:=\Exp_{\by,\bt}\left[\frac{1}{P_\by}(\bS_1(\by,\bt)\bt)^\top(\bS_1(\by,\bt)\bS_1(\by,\bt)^\top)^{\dagger}(\bS_1(\by,\bt)\bt)\right]
    \end{equation}
    where $P_\by$ denotes the number of non-zero entries in $\by$ (i.e., number of labels that are in comparison by $\by$), and $\dagger$ denotes matrix pseudoinverse.
    \item Effective manifold radius $R$ is defined as
    \begin{equation}\label{eq:effective radius}
        R:=\sqrt{\frac{\Exp_{\by,\bt}[c(\by,\bt)]}{\Exp_{\by,\bt}[b(\by,\bt)-c(\by,\bt)]}} 
    \end{equation}
    where $b(\by,\bt)=(\bS_1(\by,\bt)\bt)^\top(\bS_1(\by,\bt)\bS_1(\by,\bt)^\top)^\dagger(\bS_1(\by,\bt)\bt)$ and $c(\by,\bt)=(\bS_1(\by,\bt)\bt)^\top(\bS_0\bS_0^\top+\bS_1(\by,\bt)\bS_1(\by,\bt)^\top)^\dagger(\bS_1(\by,\bt)\bt)$.
    \item Anchor center alignment $\rho_c$ is defined as
    \begin{equation}\label{eq:anchor center}
        \rho_c := \frac{1}{C(C-1)}\sum_{c\neq c'}\frac{|\langle\bs_{c,0},\bs_{c',0}\rangle|}{\|\bs_{c,0}\|_2\cdot\|\bs_{c',0}\|_2} \, .
    \end{equation}
    \item Anchor axis alignment $\rho_a$ is defined as
    \begin{equation}\label{eq:anchor axis}
        \rho_a := \frac{1}{C(C-1)}\sum_{c\neq c'}\Exp_{\by,\bt}\left[\frac{|\langle\bs_{c,1}(\by,\bt),\bs_{c',1}(\by,\bt)\rangle|}{\|\bs_{c,1}(\by,\bt)\|_2\cdot\|\bs_{c',1}(\by,\bt)\|_2} \right] \, .
    \end{equation}
\end{itemize}

In all our analyses, we sampled 200 random $\bt$ to estimate $N_\crit$ and the GLUE geometric measures. 
Also, we considered random pairwise manifold comparisons, and each manifold contains 30 points randomly sampled from the point cloud. For each configuration, we ran 50 repeats and report the average results.
We provide detailed derivations and intuitions for these GLUE geometric measures in~\autoref{app:glue geometric measures}.

\subsubsection{Linear probe}\label{app:implementation LP}
Given $C$ label-conditioned manifolds $\{\cM_c\}$ in $\Real^N$, a linear probe is a matrix $W \in \Real^{C \times N}$ that performs $C$-class classification through a linear map $\bx \mapsto W\phi_\theta(\bx)$, followed by a softmax, where $W$ is fit by minimizing a regularized cross-entropy loss on the training set while keeping the encoder $\phi_\theta$ frozen.

We fit $W$ using AdamW with learning rate $10^{-3}$, $(\beta_1, \beta_2) = (0.9, 0.98)$, and weight decay $1.0$ for 200 full-batch epochs. The strong weight decay ensures that high probe accuracy reflects genuinely structured representations rather than memorization. $W$ is initialized as $W_{ij} \sim \mathcal{N}(0, 1/N)$ with no bias term. The train-test split matches that of the model being probed, and we report both train and test probe accuracy. See more details and interpretations in~\autoref{app:linear probe}

\subsubsection{Last-layer NTK-label alignment}\label{app:implementation NTK}
The last-layer NTK-label alignment quantifies how well the network's last-layer kernel aligns with the label structure. For an encoder $\phi_\theta : \cX \to \Real^N$ with a linear readout and a dataset $\{(\bx_i, y_i)\}_{i=1}^n$, the last-layer NTK takes the block-diagonal form $K_{\mathrm{NTK},(i,c),(i',c')} = \delta_{cc'} \, \phi_\theta(\bx_i)^\top \phi_\theta(\bx_{i'})$, i.e., the feature Gram matrix $K = \Phi\Phi^\top$ replicated once per class. 
Here $\delta_{cc'}$ denotes the Delta function, i.e., $\delta_{cc'}=1$ if $c=c'$ and $\delta_{cc'}=0$ if $c\neq c'$.
The alignment is the centered kernel alignment (CKA) between $K_{\mathrm{NTK}}$ and the label kernel $Y_{(i,c),(i',c')} = \delta_{cc'}\mathbf{1}[y_i = c]$, defined via the Hilbert-Schmidt Independence Criterion $\mathrm{HSIC}(K, K') = \frac{1}{(n-1)^2}\mathrm{tr}(K H K' H)$, where $H = I - \frac{1}{n}\mathbf{1}\mathbf{1}^\top$ is the centering matrix:
\begin{equation}\label{eq:ntk label alignment}
    \mathrm{Align}(K_{\mathrm{NTK}}, Y) = \frac{\mathrm{HSIC}(K_{\mathrm{NTK}}, Y)}{\sqrt{\mathrm{HSIC}(K_{\mathrm{NTK}}, K_{\mathrm{NTK}})\,\mathrm{HSIC}(Y, Y)}} \, .
\end{equation}
Since $K_{\mathrm{NTK}}$ factors through the feature Gram matrix, all three HSIC terms reduce to closed-form traces involving $K$ and the centered one-hot label matrix $\tilde{Y} \in \Real^{n \times C}$, requiring no backward passes or Hutchinson estimation. We compute alignment on both train and test sets at each checkpoint; see~\autoref{app:ntk} for the full derivation, interpretations, and intuitions.

\subsection{Critical dimension}\label{app:critical dimension and GLUE}
Critical dimension is the key concept in GLUE theory that quantifies the average-case separability—also known as the degree of manifold untangling—of neural representations. It was tightly related to the notion of manifold capacity~\citep{chung2018classification,wakhloo2023linear,chou2025glue} from theoretical neuroscience and statistical physics (see~\autoref{app:GLUE disc}). Recently, Chou \& Le et al.~\citep{chou2025featurelearninglazyrichdichotomy} showed that critical dimension excels at quantifying the degree of feature learning compared to other previously used kernel-based and weight-based measures.

In this subsection, we first explain the connection between critical dimension and the closely related notion of manifold capacity in~\autoref{app:critical dimension manifold capacity}. Next, we provide examples to illustrate why critical dimension captures the average-case separability of neural representations in~\autoref{app:critical dimension average case}. Finally, we explain the connection to feature learning in~\autoref{app:critical dimension feature learning}.

\subsubsection{Connection between critical dimension and manifold capacity}\label{app:critical dimension manifold capacity}
It all began with the concept of perceptron capacity, which asks how many distinct points can be stored in an $N$-dimensional space such that every dichotomy over these points admits a linear classifier. Cover~\citep{cover1965geometrical} famously showed that the answer is $2N$ as long as the points are not degenerate (i.e., in general position). Namely, the \textit{capacity} of a neural network for linear readout (i.e., perceptron) is $2$.

While the notion of perceptron capacity spurred a rich line of work on figuring out if it's possible to increase capacity of a network, or the capacity of other types of readout, etc., a recent line of work on \textit{manifold capacity} questions the beginning assumptions on modeling the patterns of representation by a single point in the neural activity space. The idea is that the collection of neural representations for the same category (e.g., cat or dog) may not collapse into a single point but rather occupy a manifold (which mathematically is simply a convex set, as here the readout is linear) in the neural activity space. So they asked about the capacity of a network for storing manifold-like patterns.

Following the tradition of the statistical physics approach in perceptron capacity, the initial attempt to define manifold capacity is through \textit{mean-field models}~\citep{chung2018classification,wakhloo2023linear,mignacco2025nonlinear}, which are mathematical models with infinitely many neurons $N_\textsf{mf}$ and infinitely many manifolds $P_\textsf{mf}$. In such a limit (i.e., $P_\textsf{mf},N_\textsf{mf}\to\infty$ with the load $P_\textsf{mf}/N_\textsf{mf}$ holds as a constant, a.k.a., thermodynamic limit), there is a phase transition on the linear separability of the mean-field model with respect to the load $P_\textsf{mf}/N_\textsf{mf}$. Concretely, there exists $\alpha>0$ such that the mean-field model is linearly separable for all dichotomies if and only if $P_\textsf{mf}/N_\textsf{mf}\leq\alpha$.

While mean-field models are analytically solvable, the downside is that one must make assumptions about how the data manifolds are distributed. For example, no correlations between the manifolds~\citep{chung2018classification} or Gaussian correlations~\citep{wakhloo2023linear}. These assumptions allow some cruel approximations in earlier applications of the manifold capacity theory~\citep{cohen2020separability,stephenson2021geometry,paraouty2023sensory}. However, the mean-field assumptions often do not hold in real-world data (e.g., Figure 2 in~\citep{chou2025glue}). This motivated a data-driven approach to manifold capacity, which was cultivated in the GLUE (Geometry Linked to Untangling Efficiency) theory~\citep{chou2025glue}. The key idea in GLUE is that, instead of building a mathematical model that generates infinitely many manifolds, they directly study a given number of manifolds from data and the dichotomies of interest, and then determine the number of subpopulations of neurons required to support linear separability with respect to these readouts. And this is exactly the notion of critical dimension as formally defined in~\autoref{app:implementation critical dimension}. Finally, in GLUE theory, the data-driven version of manifold capacity is hence defined as $P_\textsf{data}/N_\crit$ where $P_\textsf{data}$ is the number of manifolds in the data.

We encourage the interested reader to read the supplementary materials of~\citep{chou2025glue} for more details on the history of manifold capacity.

\subsubsection{Critical dimension quantifies average-case separability}\label{app:critical dimension average case}
To build intuition for critical dimension, we use a simple two-dimensional toy example that isolates the difference between \emph{best-case} and \emph{average-case} linear separability.

\paragraph{Best-case versus average-case separability.}
Consider two Gaussian classes in $\mathbb{R}^2$: class $+$ with mean $(\mu, 0)$ and class $-$ with mean $(-\mu, 0)$, both sharing covariance $\Sigma = \mathrm{diag}(\sigma_1^2,\, \sigma_\perp^2)$.  The best-case linear readout---i.e., the optimal linear discriminant (LDA) or the support vector machine (SVM)---recovers the Fisher direction $\Sigma^{-1}(\mu_+ - \mu_-) \propto e_1$ regardless of $\sigma_\perp$.  The resulting accuracy is $\Phi(\mu/\sigma_1)$, which is \emph{entirely independent of $\sigma_\perp$}.  Similarly, the SVM margin is determined solely by the separation $2\mu$ and $\sigma_1$.

\begin{figure}[h!]
\centering
\includegraphics[width=0.8\linewidth]{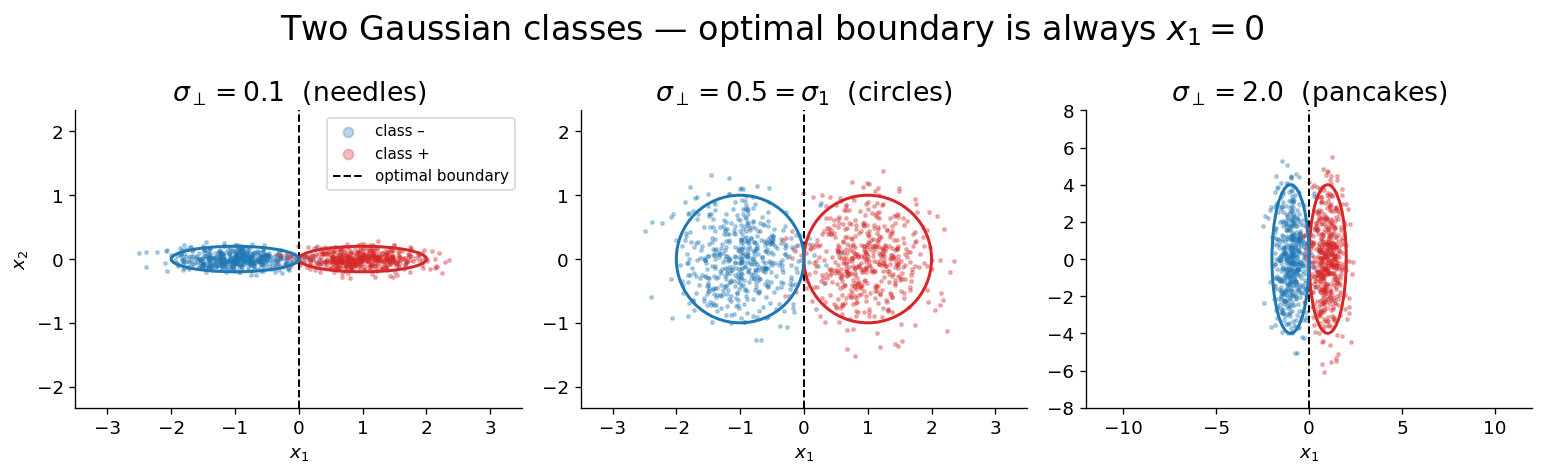}
\caption{Three regimes of the two-ellipsoid toy.  Both classes share the same covariance $\Sigma = \mathrm{diag}(\sigma_1^2, \sigma_\perp^2)$; the optimal separating boundary (dashed) is always $x_1 = 0$ regardless of $\sigma_\perp$.}
\label{fig:crit dim toy1}
\end{figure}

\paragraph{A toy with two parallel ellipsoids.}
We now vary $\sigma_\perp$ while holding $\mu$ and $\sigma_1$ fixed.  Three regimes are illustrated in \autoref{fig:crit dim toy1}: (i)~\emph{needles} ($\sigma_\perp \ll \sigma_1$), where both classes form elongated cigars along $x_1$; (ii)~\emph{circles} ($\sigma_\perp = \sigma_1$), where both classes are isotropic; and (iii)~\emph{pancakes} ($\sigma_\perp \gg \sigma_1$), where both classes spread widely along $x_2$.  In all three regimes, the optimal separating hyperplane is $x_1 = 0$ and the best-case accuracy is identical.

\begin{figure}[ht]
\centering
\includegraphics[width=0.8\linewidth]{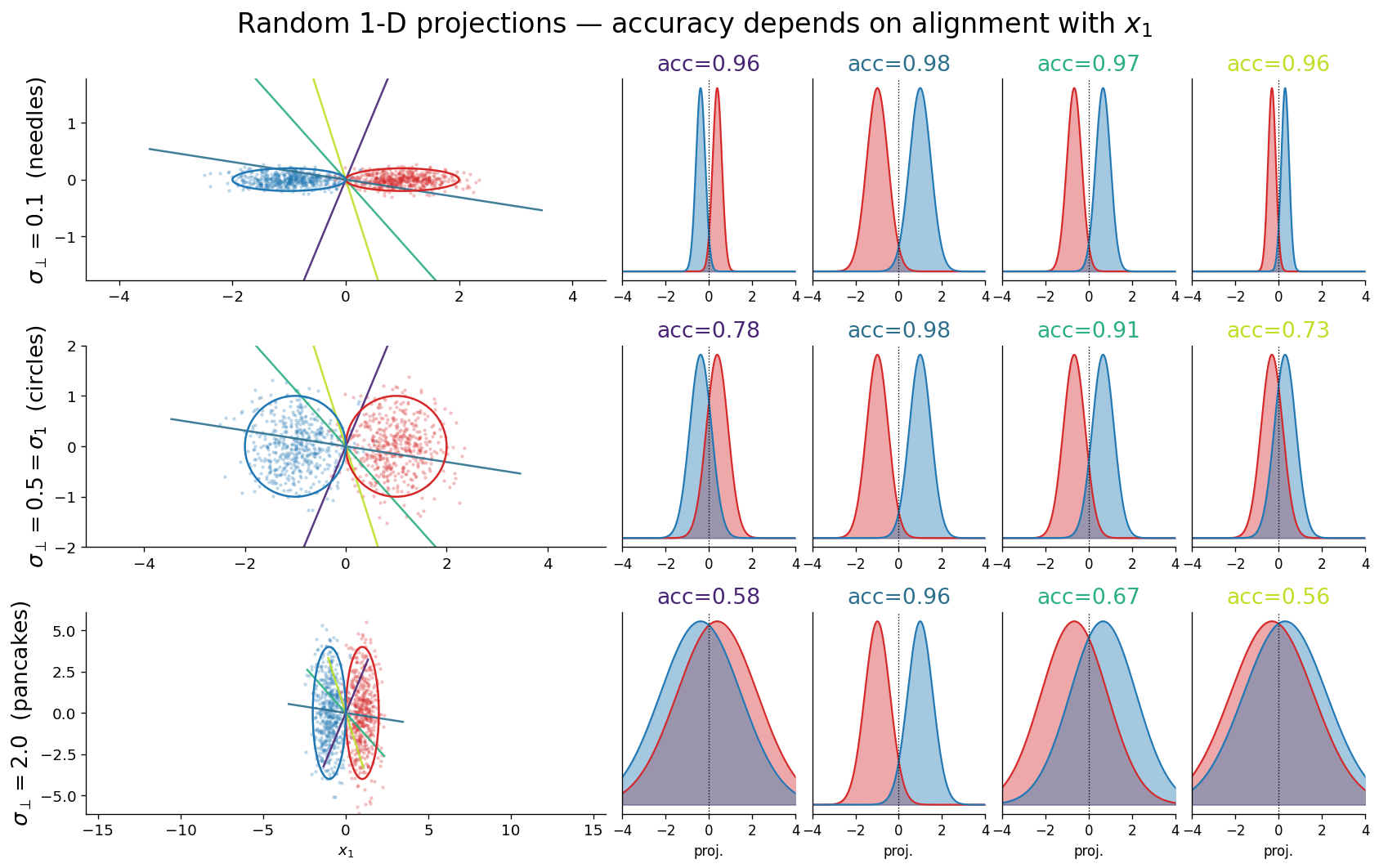}
\caption{Random 1-D projections of the two-ellipsoid toy (one column per projection direction, one row per regime).  Each panel shows the Bayes-optimal accuracy for that projection direction and the resulting one-dimensional class distributions.  As $\sigma_\perp$ grows, more projection directions yield near-chance separation.}
\label{fig:crit dim toy2}
\end{figure}

However, these three regimes differ dramatically in their \emph{average-case} separability.  Consider projecting the data onto a random one-dimensional direction $\theta \in [0, \pi)$; \autoref{fig:crit dim toy2} illustrates several such projections and the resulting one-dimensional class distributions.  For projections near the optimal direction $\theta \approx 0$ the two classes separate cleanly, but for directions near $\theta \approx \pi/2$ the projected distributions overlap strongly---and this overlap worsens as $\sigma_\perp$ grows.

\begin{figure}[ht]
\centering
\includegraphics[width=0.65\linewidth]{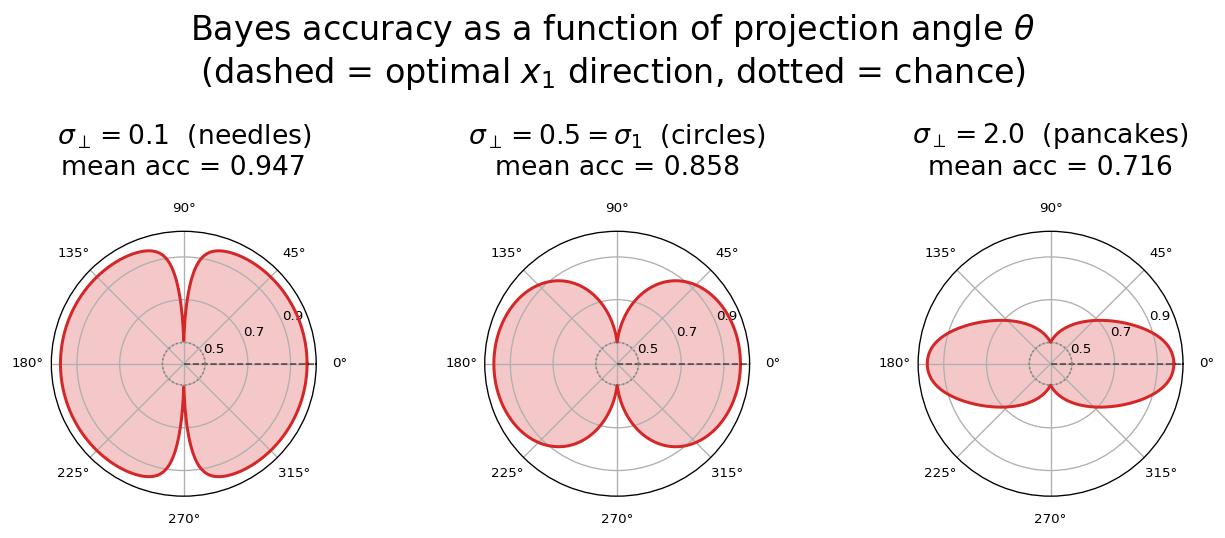}
\caption{Polar plot of the Bayes-optimal 1-D projection accuracy $\mathrm{Acc}(\theta)$ as a function of projection angle $\theta$ for each regime.  The filled region shrinks from a broad disc (needles) to a narrow lobe (pancakes), visualizing the shrinking cone of good projections.  Dotted circle marks chance accuracy ($0.5$).}
\label{fig:crit dim toy3}
\end{figure}

The Bayes-optimal accuracy for a projection at angle $\theta$ is
\begin{equation}\label{eq:acc at angle}
\mathrm{Acc}(\theta) = \Phi\!\left(\frac{\mu \lvert\cos\theta\rvert}{\sqrt{\sigma_1^2 \cos^2\theta + \sigma_\perp^2 \sin^2\theta}}\right),
\end{equation}
where $\Phi$ is the standard normal CDF.  \autoref{fig:crit dim toy3} shows $\mathrm{Acc}(\theta)$ as a polar plot over all projection angles.  In the needle regime, the filled region is nearly circular at high accuracy: almost any random projection recovers the class structure.  In the pancake regime, the region collapses to a narrow lobe near $\theta = 0$---a \emph{cone of good projections}---while the rest of the circle sits near chance.  The mean random-projection accuracy $\frac{1}{\pi}\int_0^{\pi} \mathrm{Acc}(\theta)\,d\theta$ therefore decreases monotonically with $\sigma_\perp$.

\begin{figure}[ht]
\centering
\includegraphics[width=0.8\linewidth]{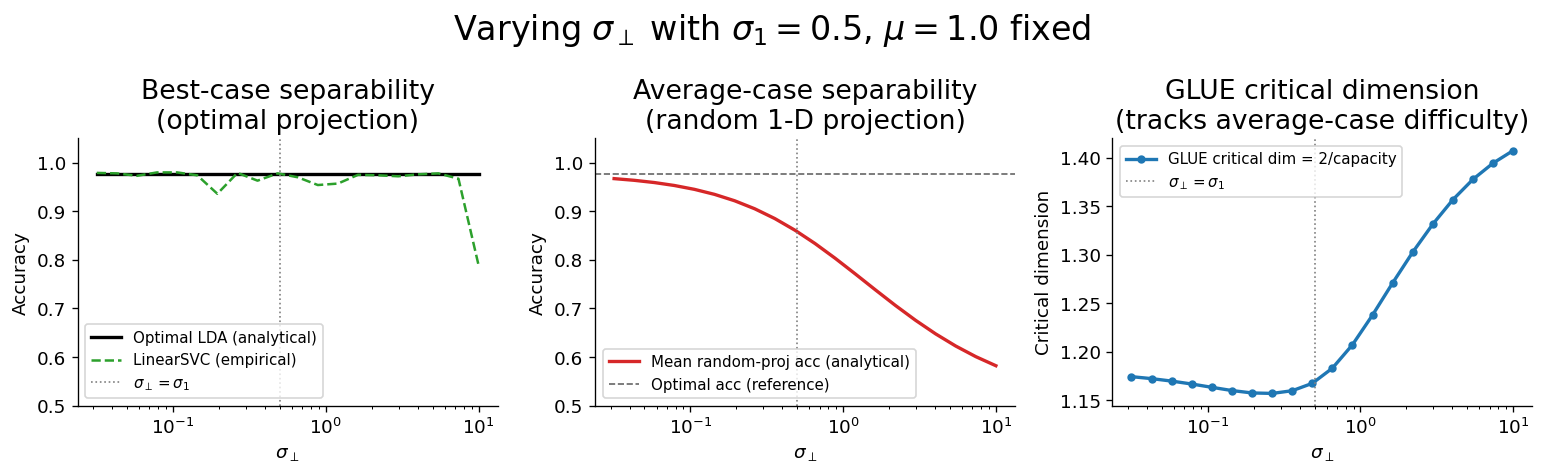}
\caption{Summary sweep over $\sigma_\perp$.  \emph{Left}: best-case separability (optimal LDA and LinearSVC) is flat.  \emph{Middle}: mean random-projection accuracy decreases monotonically.  \emph{Right}: Critical dimension increases monotonically, tracking the degradation in average-case separability.}
\label{fig:crit dim toy4}
\end{figure}

\paragraph{Critical dimension tracks average-case separability.}
\autoref{fig:crit dim toy4} summarizes all four quantities as $\sigma_\perp$ is swept over two orders of magnitude. The best-case measures---optimal LDA accuracy and LinearSVC accuracy---remain flat throughout, confirming that the optimal readout is blind to $\sigma_\perp$.  In contrast, the mean random-projection accuracy (middle panel) decreases monotonically, and the GLUE critical dimension (right panel) increases in lockstep. 

\paragraph{Intuition for neural representations.}
Why does average-case separability matter?  In practice, the downstream decoder that reads out from neural representations does not have access to the globally optimal direction---it operates on a finite training set and must generalize.  Moreover, a single population of neurons may need to support multiple distinct readouts simultaneously, only a few of which are aligned with any particular optimal direction.  When the critical dimension is low, weakly optimized or randomly initialized readout directions are already informative, and the representation is said to be \emph{untangled}.  When the critical dimension is high, only a narrow subset of directions supports accurate decoding, so downstream decoders must work harder to identify the correct subspace.  This is precisely the sense in which critical dimension quantifies the \emph{degree of manifold untangling} referenced in earlier work~\citep{dicarlo2007untangling,chung2018classification,chou2025glue}.

\subsubsection{Critical dimension quantifies the degree of feature learning}\label{app:critical dimension feature learning}
Chou and Le et al.~\citep{chou2025featurelearninglazyrichdichotomy} proposed to study feature learning through the geometry of task-relevant manifolds, where a task-relevant manifold is the point cloud of neural activity patterns corresponding to a given category or condition.  Their central observation is that feature learning can be viewed as a process of \emph{untangling} task-relevant manifolds---structuring the neural representational space to improve separation among manifolds.  Concretely, in a network that does not learn task-relevant features (e.g., the lazy or random-feature regime), the manifolds are poorly organized, making them harder to distinguish (e.g., smaller margin, smaller solution volume).  In contrast, when a network learns task-relevant features (e.g., rich learning, neural collapse), the manifolds become well-organized and easier to separate (e.g., larger margin, larger solution volume).

As established in \autoref{app:critical dimension average case}, manifold capacity $\alpha$ quantifies this degree of untangling via an \emph{average-case} notion of separability, and critical dimension $N_\mathrm{crit} = 2/\alpha$ is its reciprocal measure: lower critical dimension means the manifolds are more untangled.  This average-case perspective is precisely what makes critical dimension well-suited for quantifying feature learning.  As noted by \citep{chou2025featurelearninglazyrichdichotomy}, the margin in support vector machine (SVM) theory quantifies the degree of separability in the worst-case setting, whereas manifold capacity (and hence critical dimension) is average-case in the sense of the random projection---capturing not just whether an optimal readout exists, but how broadly separable the representations are.

\begin{figure}[ht]
\centering
\includegraphics[width=0.8\linewidth]{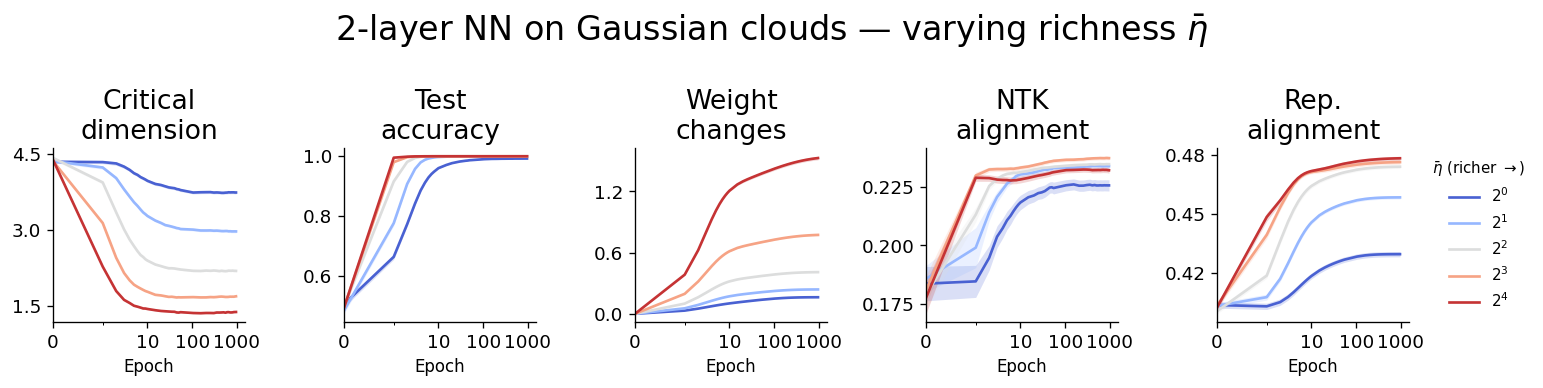}
\caption{Critical dimension tracks the degree of feature learning.  2-layer non-linear networks are trained on Gaussian clouds with the inverse scale factor $\bar\eta$ varied across five richness levels (blue $2^0$ laziest to red $2^4$ richest; shading is $\pm 1$ SEM over seeds).  Critical dimension $N_\mathrm{crit} = 2/\alpha$ (left panel) correctly orders all five conditions throughout training: richer $\bar\eta$ yields lower critical dimension.  Test accuracy (second panel) is insensitive to richness; weight changes (third), NTK--label alignment (fourth), and representation--label alignment (fifth) exhibit incorrect orderings in some regimes.  Experiment reproduced from \citep{chou2025featurelearninglazyrichdichotomy}.}
\label{fig:crit dim feature learning}
\end{figure}

\paragraph{Toy setup.}  To demonstrate this empirically, \citep{chou2025featurelearninglazyrichdichotomy} considers 2-layer non-linear networks of the form $f(\mathbf{x}) = \frac{1}{\sqrt{N}}\mathbf{a}^\top\sigma(W^\top\mathbf{x})$, where $W \in \mathbb{R}^{N \times d}$ is the hidden-layer weight matrix, $\mathbf{a} \in \mathbb{R}^N$ is a fixed random readout, and $\sigma$ is a ReLU activation.  The networks are trained on Gaussian cloud data in a teacher--student setting, where the input--label pairs are generated by a hidden signal $\beta^*$.  The inverse scale factor $\bar\eta \in \{2^0, 2^1, 2^2, 2^3, 2^4\}$ controls the effective learning rate, interpolating from a lazy (kernel-like) regime at $\bar\eta = 2^0$ to a rich (feature-learning) regime at $\bar\eta = 2^4$.  We reproduce their experiment and compare five measures across all richness levels: critical dimension, test accuracy, weight changes $\|W_t - W_0\|_F / \|W_0\|_F$, NTK--label alignment $\mathrm{CKA}(K_t^\mathrm{NTK}, \mathbf{yy}^\top)$, and representation--label alignment $\mathrm{CKA}(X_t X_t^\top, \mathbf{yy}^\top)$.

\paragraph{Results.}  \autoref{fig:crit dim feature learning} shows the result.  Critical dimension correctly and consistently orders all five richness levels throughout training: richer training (higher $\bar\eta$) produces representations with lower critical dimension, i.e., more untangled manifolds, as expected.  The other measures fail in different ways.  Test accuracy is nearly identical across all conditions and cannot distinguish richness levels at any point during training.  Weight changes and NTK--label alignment track some differences early in training but exhibit crossings and incorrect orderings in later epochs.  Representation--label alignment shows the wrong ordering entirely in some regimes, consistent with the finding in \citep{chou2025featurelearninglazyrichdichotomy} that the representation-label alignment would characterize the wrong ordering of wealthiness in initial features.  As the authors conclude, manifold capacity excels at extracting task-relevant structures in representations because it is data-driven and free from additional statistical assumptions on the data.

\paragraph{Theoretical grounding.}  \citep{chou2025featurelearninglazyrichdichotomy} also provide formal justification for this finding via Theorem~3.1, which states that in the proportional asymptotic limit ($P_\mathrm{train}, d, N \to \infty$ at the same rate), capacity monotonically tracks the learning rate: $\alpha(\eta, \psi_1, \psi_2) < \alpha(\eta', \psi_1, \psi_2)$ for every $0 < \eta < \eta'$.  Moreover, there exists an increasing invertible function $h_{\psi_1,\psi_2} : \mathbb{R}_+ \to [0,1]$ such that prediction accuracy satisfies $\mathrm{Acc}(\eta, \psi_1, \psi_2) = h_{\psi_1,\psi_2}(\alpha(\eta, \psi_1, \psi_2))$, establishing a monotone link between capacity and accuracy.  Translating to critical dimension: lower $N_\mathrm{crit}$ implies higher capacity, which implies higher prediction accuracy---making critical dimension a principled, theoretically grounded measure of the degree of feature learning.

\subsection{GLUE geometric measures}\label{app:glue geometric measures}
In~\autoref{app:critical dimension and GLUE} we introduced the connection between critical dimension and manifold capacity theory. One thing we skipped was about how the geometric organization of manifolds relates to the value of manifold capacity. Early work of Chung et al.~\citep{chung2018classification} used the mean-field formula for manifold capacity to identify that manifold dimension and manifold radius contribute to the capacity value. Specifically, the smaller the manifold dimension and radius, the higher the manifold capacity. One important thing is the anchor point distribution, which better captures the task-relevant structure of data, analogous to the support vector in SVM, but now it's average-case as opposed to the best-case in SVM.

The original mean-field theory framework only captured \textit{intra-class} geometry through dimension and radius. The GLUE framework~\citep{chou2025glue} extends this by introducing two additional measures that capture \textit{inter-class} geometric structure: anchor center alignment $\rho_c$ and anchor axis alignment $\rho_a$. Center alignment measures how similarly the anchor centers of different classes are oriented in representation space --- when $\rho_c$ is high, the mean anchor points of different classes point in similar directions, making them harder to separate. Axis alignment measures how parallel the local variation directions are across classes --- when $\rho_a$ is high, the per-sample fluctuations of different manifolds are correlated, which reduces the effective diversity of random projections and thus lowers capacity. Formal definitions of all four measures are given in~\autoref{app:implementation GLUE geometry}.

\begin{figure}[h!]
    \centering
    \includegraphics[width=0.8\linewidth]{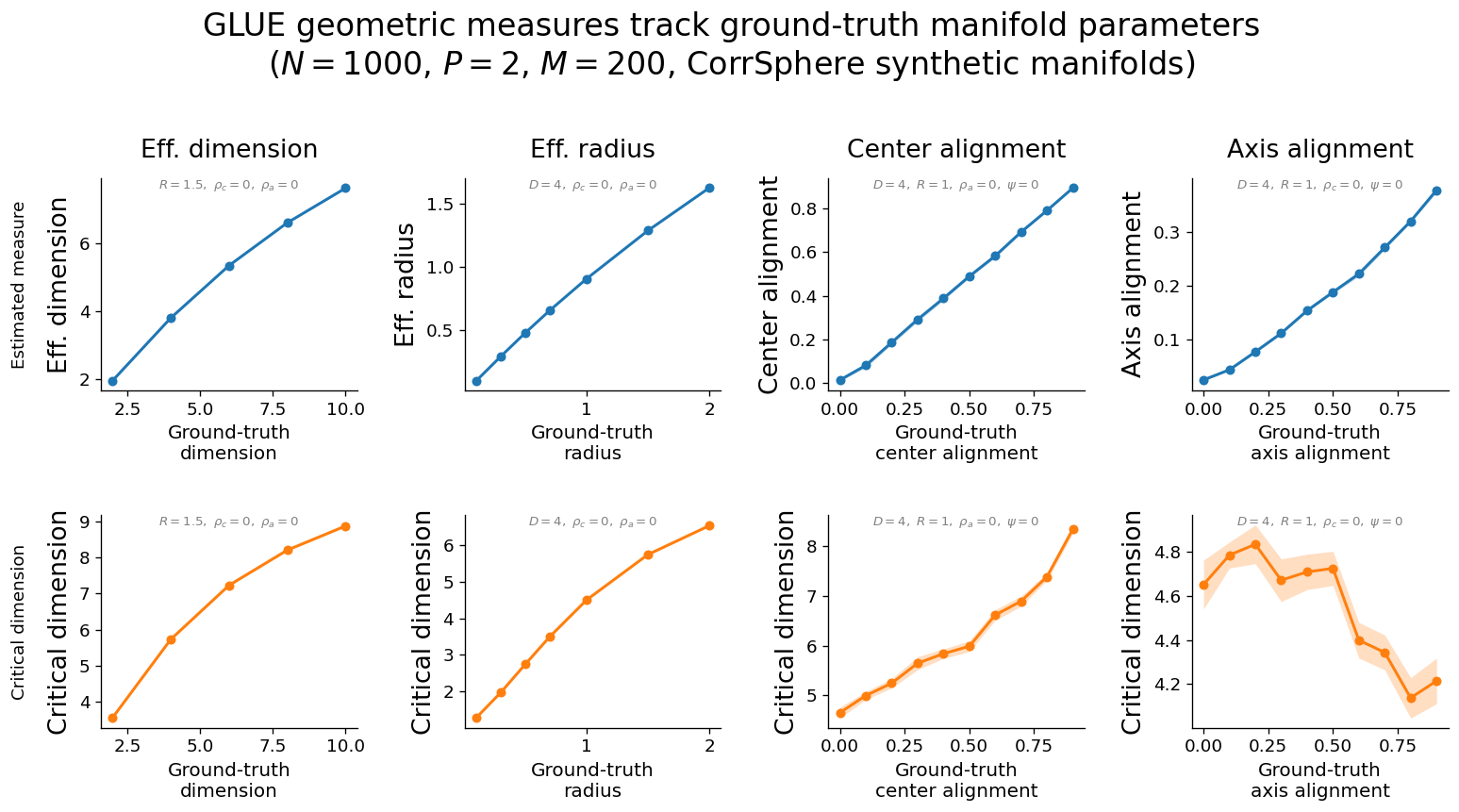}
    \caption{\textbf{Intuition for GLUE geometric measures.} Each measure is validated on synthetic Gaussian manifolds ($N=1000$, $P=2$, $M=200$) from~\citep{chou2025glue}. Each column sweeps one ground-truth parameter while holding the others fixed. \textbf{Top:} estimated GLUE measure vs.\ ground-truth parameter. \textbf{Bottom:} critical dimension $N_\crit = 2/\alpha$, where $\alpha$ is manifold capacity, vs.\ ground-truth parameter.}
    \label{fig:glue geometry}
\end{figure}

Together, these four measures decompose manifold geometry into two orthogonal aspects: intra-class geometry ($D$, $R$), which governs how spread out each manifold is on its own, and inter-class geometry ($\rho_c$, $\rho_a$), which governs how geometrically similar different class manifolds are to one another. In all cases, values that indicate more ``complex'' or entangled geometry --- larger dimension, larger radius, higher center alignment, higher axis alignment --- correspond to lower manifold capacity and thus higher critical dimension. This decomposition provides an interpretable diagnostic: when we observe a drop in critical dimension during training, we can attribute it to specific geometric changes in the learned representations.

To build intuition for these four measures, we reproduce the validation experiment from~\citep{chou2025glue} in~\autoref{fig:glue geometry}. The toy example generates synthetic Gaussian manifolds where each ground-truth parameter is swept independently, serving as a controlled setting where the geometric meaning of each measure can be verified directly. The top row confirms that each GLUE estimate faithfully tracks its corresponding ground-truth parameter; the bottom row shows that critical dimension inherits a monotone relationship with each, validating it as a composite readout of all four geometric effects.

\subsection{Linear probe}\label{app:linear probe}
Linear probing~\citep{alain2016understanding} is a diagnostic tool for evaluating the quality of learned representations. A linear classifier is trained on top of frozen network activations, and its accuracy measures whether class-relevant structure is \textit{linearly accessible} in the representation—decoupled from whether the full network exploits it via nonlinear layers. While prior work has used layer-wise LP accuracy to track how information propagates through the network~\citep{alain2016understanding}, we focus exclusively on the final-layer representation.

In analytically tractable settings, LP accuracy has a precise geometric meaning. In the SVM framework~\citep{cortes1995support}, LP accuracy tracks linear separability and the margin directly quantifies the degree of separation~\citep{bartlett2002rademacher}. In kernel and ridge regression~\citep{scholkopf2002learning}, generalization is governed by the linear separability of the kernel features, so LP accuracy on those features directly bounds the generalization gap. In the NTK / lazy training regime~\citep{jacot2018neural}, the network behaves as a kernel machine throughout training, and LP accuracy on the final layer is equivalent to the kernel classifier—LP accuracy and test accuracy coincide with no gap between them.

In deep networks trained in the rich / feature-learning regime, this equivalence breaks down. The LP classifier is applied directly to the frozen final-layer pre-logits, which are already the input to a linear output head, so LP involves no hidden layer. However, the nonlinear layers upstream can produce final-layer representations with poor linear structure even when train accuracy is high—the network routes around this via its nonlinear composition. LP accuracy exposes this gap: it measures representation quality independently of whether the network's downstream classifier uses the representation correctly.

\subsection{Last-layer NTK-label alignment}
\label{app:ntk}
In this subsection, we offer more details on the (last-layer) NTK-kernel alignment measure as introduced in~\autoref{sec:methods measures} and~\autoref{app:implementation NTK}. Specifically, we provide 2D toy examples in~\autoref{app:ntk_intuition} to provide more intuitions on thinking about the value of NTK alignment, and a connection between NTK alignment and readout weight learning dynamics in~\autoref{app:NTK readout learning}.

\subsubsection{Intuition for last-layer NTK alignment}
\label{app:ntk_intuition}
We refer the reader to \autoref{app:implementation NTK} for the formal definition of last-layer NTK alignment. This section develops the geometric intuition behind the metric through two contrasting binary classification toy examples.
By the end, the reader should have a clear picture of what the alignment score measures, why it is sensitive to class-discriminative structure in the representations, and why it rises as a network transitions from lazy to rich learning during grokking.

\begin{figure}[ht]
    \centering
    \includegraphics[width=\linewidth]{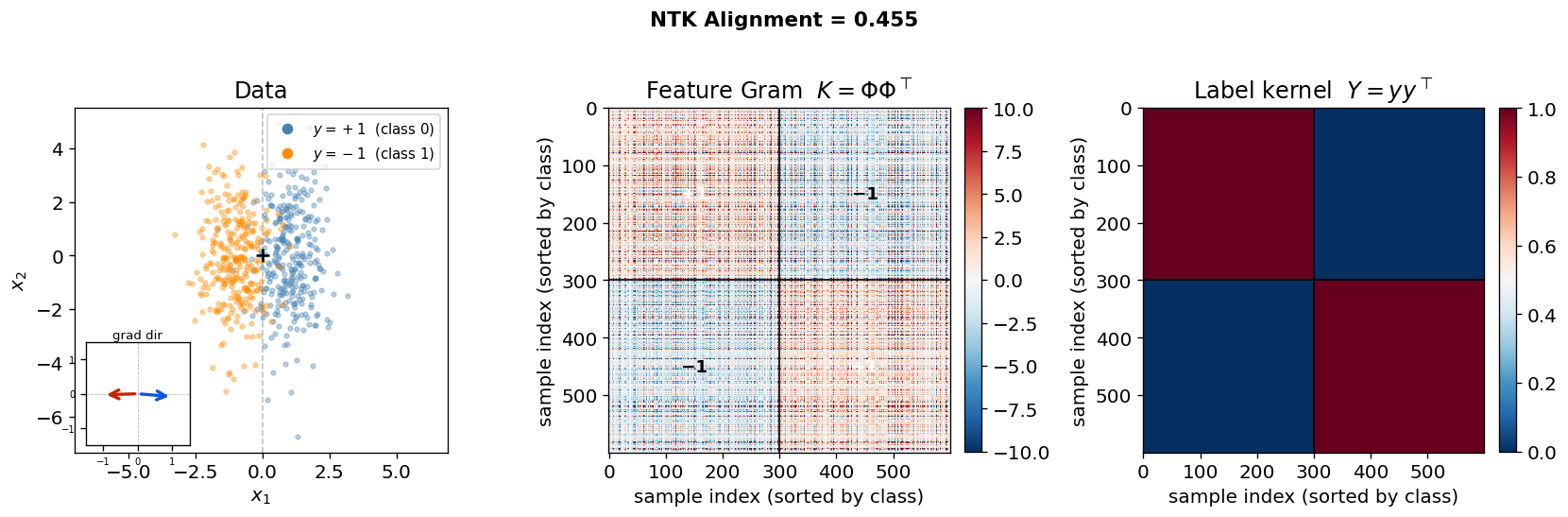}
    \caption{A toy 2D example of high NTK alignment.}
    \label{fig:ntk example 1}
\end{figure}

\begin{figure}[ht]
    \centering
    \includegraphics[width=\linewidth]{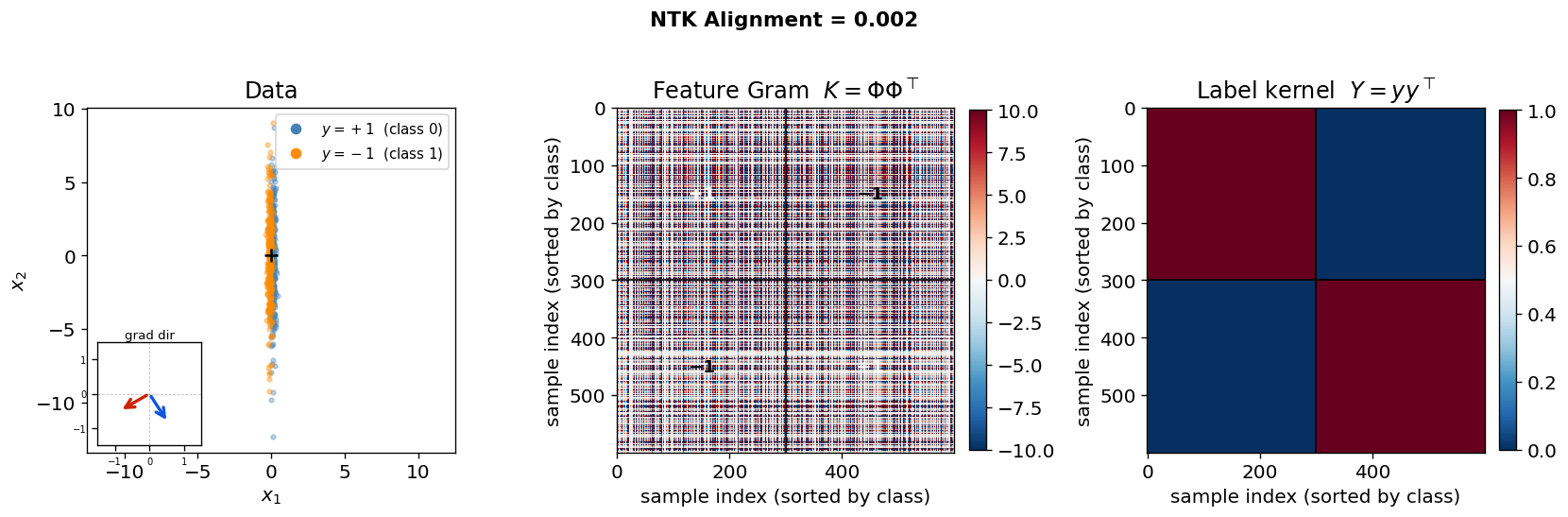}
    \caption{A toy 2D example of low NTK alignment.}
    \label{fig:ntk example 2}
\end{figure}

\paragraph{Binary classification.}
Consider a dataset of $n$ samples with features $\Phi \in \mathbb{R}^{N \times n}$ and binary labels $y_i \in \{+1, -1\}$.
The feature Gram matrix $K = \Phi^\top \Phi \in \mathbb{R}^{n \times n}$, shown in the \emph{middle panel} of~\autoref{fig:ntk example 1}~and~\autoref{fig:ntk example 2}, records all pairwise inner products $K_{ij} = \phi_i^\top \phi_j$ between representations.
The label kernel $Y \in \mathbb{R}^{n \times n}$ where $Y_{i,j}=y_iy_j$, shown in the \emph{right panel}, encodes class structure: $Y_{ij} = +1$ when samples $i$ and $j$ share a class and $Y_{ij} = -1$ otherwise. Alignment between $K$ and $Y$ is their normalized Frobenius inner product,
\begin{equation}
    \text{alignment}(K, Y)
    = \frac{\langle K, Y \rangle_F}{\|K\|_F \|Y\|_F}
    \propto \sum_{\substack{i,j \\ \text{same class}}} K_{ij}
           - \sum_{\substack{i,j \\ \text{diff.\ class}}} K_{ij}.
    \label{eq:ntk_alignment}
\end{equation}
where $\langle K, Y \rangle_F = \sum_{i,j} K_{ij} Y_{ij}$ is the Frobenius inner product and $\|K\|_F = \sqrt{\langle K, K \rangle_F}$ is the Frobenius norm. \autoref{eq:ntk_alignment} says that alignment is high precisely when within-class representations are more similar to each other than to representations from the other class (i.e., when the features cluster by label).

This has a direct connection to gradient-based learning. At initialization, with readout weights $W = 0$, the softmax outputs uniform class probabilities $p(c \mid \phi_i) = \tfrac{1}{2}$ for all samples. The gradient of the cross-entropy loss with respect to the readout weight vector for class $c$ is
\begin{equation}
    -\frac{\partial \ell}{\partial w_c}\bigg|_{W=0}
    \;\propto\; \bar{\phi}_c - \bar{\phi},
    \label{eq:readout_gradient}
\end{equation}
where $w_c$ is the readout weight vector of class $c$, and $\bar{\phi}_c$ is the mean representation of class $c$ and $\bar{\phi}$ is the overall mean. For binary classification, $\bar{\phi}_{+1} - \bar{\phi} = \tfrac{1}{2}(\bar{\phi}_{+1} - \bar{\phi}_{-1})$, so both readout vectors are pushed in opposite directions along the class-mean difference vector, i.e., $\bar{\phi}_{+1} - \bar{\phi}_{-1}$, shown as arrows in the inset of the \emph{left panel}.
If this difference is large relative to the within-class spread, the first gradient step is informative and $K$ aligns well with $Y$. If the class means nearly coincide, the gradient is small and $K$ carries no class structure, giving low alignment.
We refer the reader to \autoref{app:NTK readout learning} for a more detailed treatment of the connection between readout learning dynamics and NTK alignment.

\paragraph{Example 1: high NTK-label alignment.}
We sample $N = 300$ points per class from two Gaussians,
\begin{equation}
    \phi \sim
    \begin{cases}
        \mathcal{N}\!\left([1,\;0]^\top,\;\mathrm{diag}(0.5,\;3.0)\right) & \text{class } {+1}, \\[2pt]
        \mathcal{N}\!\left([-1,\;0]^\top,\;\mathrm{diag}(0.5,\;3.0)\right) & \text{class } {-1}.
    \end{cases}
\end{equation}

The class means are well-separated (by 2 units along $x_1$), while both classes share the same large variance along $x_2$.
Because the $x_2$ variance is identical for both classes, it does not contribute to $\bar{\phi}_{+1} - \bar{\phi}_{-1}$; only the discriminative $x_1$ direction matters.
As visible in the left panel of \autoref{fig:ntk example 1}, the inset arrows point cleanly along $\pm x_1$, confirming that the readout gradient is informative. The Gram matrix $K$ (middle panel) displays a clear block structure, with within-class inner products consistently positive and large while between-class ones are negative, and therefore aligns strongly with the label kernel $Y$ (right panel).

\paragraph{Example 2: low NTK-label alignment.}
We sample the same number of points from
\begin{equation}
    \phi \sim
    \begin{cases}
        \mathcal{N}\!\left([0.1,\;0]^\top,\;\mathrm{diag}(0.01,\;10.0)\right) & \text{class } {+1}, \\[2pt]
        \mathcal{N}\!\left([-0.1,\;0]^\top,\;\mathrm{diag}(0.01,\;10.0)\right) & \text{class } {-1}.
    \end{cases}
\end{equation}

The class means now nearly coincide (separated by only $0.2$ along $x_1$), while the variance along $x_2$ is very large ($= 10$) and shared by both classes. Every pairwise inner product $K_{ij} = \phi_i^\top \phi_j$ is dominated by the $x_2$ component, which is uninformative about class identity. Consequently, both same-class and different-class pairs produce similar inner products, and the Gram matrix (middle panel of \autoref{fig:ntk example 2}) shows no block structure. The inset in the left panel confirms this: because $\bar{\phi}_{+1} \approx \bar{\phi}_{-1}$, the readout gradient $\bar{\phi}_c - \bar{\phi}$ is negligibly small in magnitude. Moreover, although the expected gradient points along $x_1$ in the limit of many samples, each individual feature vector $\phi_i$ is dominated by its large $x_2$ component (variance $= 10.0$).
In any finite mini-batch, the $x_2$ fluctuations therefore swamp the tiny $x_1$ signal, so the effective gradient direction is not reliably aligned with the label-discriminative axis.
The resulting alignment with $Y$ (right panel) is low, even though the two classes are in principle linearly separable along $x_1$.

\subsubsection{Connection between NTK-label alignment and readout learning dynamics}\label{app:NTK readout learning}

A central challenge in understanding representation learning is bridging \emph{performance-level} quantities (e.g., training loss or classification accuracy) and \emph{representation-level} quantities (e.g., the geometry of the feature space). The former depends on the joint dynamics of the encoder and the readout, as well as on the choice of optimizer, making it difficult to attribute changes in performance to changes in representations
alone. The latter, on the other hand, can be computed as a static snapshot at any checkpoint, but does not obviously reflect what the optimizer is doing.

NTK-label alignment offers one principled bridge between these two levels.  As we show below, under a set of idealized assumptions, it is \emph{exactly} proportional to the initial rate of loss decay when training a linear readout on top of frozen features.  In practice, these assumptions are not fully satisfied, but alignment continues to serve as a tractable and interpretable proxy for how well the current representations support linear classification.

\paragraph{Setup and notation.}
Consider an encoder $\phi_\theta : \mathcal{X} \to \mathbb{R}^N$ and a dataset $\{(\mathbf{x}_i, y_i)\}_{i=1}^n$ with $C$ classes.  We collect the encoder outputs into the feature matrix $\Phi \in \mathbb{R}^{N \times n}$, whose $i$-th column is $\phi_i = \phi_\theta(\mathbf{x}_i) \in \mathbb{R}^N$. The linear readout is parameterized by $W \in \mathbb{R}^{C \times N}$, so that the matrix of logits is $W\Phi \in \mathbb{R}^{C \times n}$, with column $i$ giving the logit vector $W\phi_i \in \mathbb{R}^C$ for sample $i$.  We collect the one-hot label vectors into the label matrix $Y \in \mathbb{R}^{C \times n}$, whose $i$-th column is the one-hot encoding of $y_i$.  Finally, the \emph{feature Gram matrix} is $K = \Phi^\top\Phi \in \mathbb{R}^{n \times n}$, with $(i,i')$-entry $K_{ii'} = \phi_i^\top\phi_{i'}$.

We make the following assumptions throughout this section:
\begin{enumerate}[label=\textbf{A\arabic*.}]
    \item \textbf{Frozen encoder.} The encoder parameters $\theta$ are held fixed; only $W$ is trained.
    \item \textbf{MSE loss.} The training objective is
    \begin{equation}
        \ell(W) = \frac{1}{2n}\|W\Phi - Y\|_F^2,
    \end{equation}
    where $\|\cdot\|_F$ denotes the Frobenius norm.
    \item \textbf{Gradient flow.} Training follows continuous-time gradient descent, i.e., $\dot{W} = -\frac{\partial \ell}{\partial W}$.
    \item \textbf{Zero initialization.} $W(0) = 0$.
    \item \textbf{Mean-zero features.} $\Phi\mathbf{1} = 0$, i.e., the feature vectors are centered across the dataset.
    \item \textbf{Balanced classes.} $Y\mathbf{1} = 0$, i.e., each class contains the same number of samples.
\end{enumerate}
Assumptions~A1--A4 are needed to connect alignment to the loss decay rate. Assumptions~A5--A6 additionally ensure that the centered and uncentered versions of the alignment numerator coincide; we discuss this further below.

\paragraph{Gradient flow on the readout.}
Under A1--A3, the gradient of $\ell$ with respect to $W$ is
\begin{equation}
    \frac{\partial \ell}{\partial W} = \frac{1}{n}(W\Phi - Y)\Phi^\top,
\end{equation}
so the gradient flow reads
\begin{equation}
    \dot{W} = -\frac{1}{n}(W\Phi - Y)\Phi^\top.
\end{equation}
It is convenient to track the \emph{residual} $R = W\Phi - Y \in
\mathbb{R}^{C \times n}$, which measures the discrepancy between the current predictions and the labels.  Since the encoder is frozen ($\dot{\Phi} = 0$), differentiating $R$ with respect to time gives $\dot{R} = \dot{W}\Phi$, and substituting the gradient flow equation yields
\begin{equation}
    \dot{R} = -\frac{1}{n}\,R\,K.
\end{equation}
This is a matrix ODE: the residual is driven by right-multiplication by the Gram matrix $K$.  It decouples across classes (rows of $R$) but couples samples (columns of $R$) through $K$.

\paragraph{Alignment as initial loss decay rate.}
We now differentiate the loss $\ell(t) = \frac{1}{2n}\|R(t)\|_F^2$
along the gradient flow.  Using the residual ODE,
\begin{equation}
    \dot{\ell} = \frac{1}{n}\langle R,\, \dot{R} \rangle_F
    = -\frac{1}{n^2}\,\mathrm{tr}(R\,K\,R^\top),
\end{equation}
where $\langle A, B \rangle_F = \mathrm{tr}(A^\top B)$ denotes the Frobenius inner product.  Evaluating at $t=0$ and using A4 ($R(0) = W(0)\Phi - Y = -Y$), we obtain the \emph{initial loss decay rate}:
\begin{equation}
    \label{eq:loss_decay}
    -\dot{\ell}(0) = \frac{1}{n^2}\,\mathrm{tr}(Y\,K\,Y^\top).
\end{equation}
We now connect this to the NTK-label alignment defined in~\autoref{eq:ntk label alignment}.  Recall that the alignment numerator is $\mathrm{HSIC}(K_\mathrm{NTK}, Y)$ (see~\autoref{app:implementation NTK} for definitions), which after reduction through the block-diagonal structure of $K_\mathrm{NTK}$ and $Y$ is proportional to $\mathrm{tr}(HKH\cdot YY^\top H)$, where $H = I - \frac{1}{n}\mathbf{1}\mathbf{1}^\top$ is the centering matrix satisfying $H^2 = H$ and $H^\top = H$.  Under A5--A6, the feature vectors and label vectors are already mean-zero, so $HK = K$ and $HY = Y$, and therefore
\begin{equation}
    \mathrm{tr}(HKH \cdot YY^\top H)
    = \mathrm{tr}(K\,YY^\top)
    = \mathrm{tr}(Y\,K\,Y^\top),
\end{equation}
which matches~\autoref{eq:loss_decay} up to the constant $\frac{1}{n^2}$.
Hence, under A1--A6:

\begin{equation}
    \boxed{-\dot{\ell}(0) \;\propto\; \mathrm{HSIC}(K_\mathrm{NTK},\, Y).}
\end{equation}

The full CKA normalization in~\autoref{eq:ntk label alignment} divides by $\sqrt{\mathrm{HSIC}(K_\mathrm{NTK},
K_\mathrm{NTK})\,\mathrm{HSIC}(Y, Y)}$, rendering the alignment invariant to rescaling of $\Phi$ (e.g., changes in feature norm across training).  This normalization is important for comparability across checkpoints and models, but does not affect the fundamental interpretation.

\paragraph{Interpretation.}
Under A1--A6, the (last-layer) NTK-label alignment at a given checkpoint answers the following counterfactual question: If the encoder were frozen at this checkpoint and a fresh linear readout were trained from zero with MSE loss under gradient flow, how fast would the loss decay initially? Concretely:
\begin{itemize}
    \item \textbf{High alignment} means the label directions are concentrated in the high-eigenvalue subspace of $K$.  A linear readout trained on these frozen features would make rapid initial progress, indicating that the representations are already well-organized for linear classification.
    \item \textbf{Low alignment} means the label directions are diffuse or concentrated in the low-eigenvalue subspace of $K$. A linear readout would make slow initial progress, indicating that the encoder has not yet structured its representations in a label-relevant way.
\end{itemize}

\paragraph{Scope and limitations.}
The above connection is exact only under A1--A6, which are idealized. In practice:
\begin{itemize}
    \item \textbf{Moving encoder (A1).}  During end-to-end training, $\Phi$ evolves jointly with $W$.  The alignment at epoch $t$ is therefore a snapshot of an instantaneous quantity, not a description of the full dynamics.  Nevertheless, it characterizes how amenable the \emph{current}
    representations are to linear readout learning.
    \item \textbf{Loss choice (A2).}  For cross-entropy loss the residual ODE acquires a nonlinear dependence on the current logits, breaking the exact proportionality.  Alignment remains a meaningful proxy but is no longer exactly tied to the loss decay rate.
    \item \textbf{Initialization (A4).}  The identification $R(0) = -Y$ relies on $W(0) = 0$.  If the readout has already been partially trained (as is the case when measuring alignment mid-training), the connection is approximate.
    \item \textbf{Normalization (A5--A6 and CKA).}  The CKA normalization removes scale information.  Two checkpoints with equal alignment but different $\|K\|_F$ would yield different raw loss decay rates. Additionally, if features or classes are not perfectly centered or balanced, the centered and uncentered alignment numerators differ by terms involving
    the mean feature vector and class frequencies.
\end{itemize}
Despite these caveats, NTK-label alignment has been used as an empirical probe of representation quality throughout the learning dynamics literature~\citep{cortes2012algorithms, baratin2021implicit,canatar2021spectral}, and our derivation provides a precise theoretical grounding for its interpretation in the idealized linear-frozen-encoder regime.

\section{Architecture, Task, and Training Details}\label{app:tasks and architectures}                 
We provide full architecture and training details for all experiments in the main text. For the grokking experiments, we consider four tasks: modular addition (\autoref{app:modular_addition}), permutation composition (\autoref{app:perm_comp}), sparse parity (\autoref{app:sparse_parity}), and MNIST digit classification (\autoref{app:mnist}). Each task is evaluated on an MLP; modular addition, permutation composition, and sparse parity are additionally evaluated on a Transformer, giving seven task-model pairs in total. For the double descent experiment (\autoref{app:details dd}), we use a ResNet-18 trained on CIFAR-10 with varying levels of label noise. All hyperparameters were chosen to reproduce the grokking and double descent phenomena reported in prior work~\citep{power2022grokking,liu2022omnigrok,nakkiran2021deep}, and are held fixed across seeds unless otherwise noted. Computing resources are described in \autoref{app:computer}.

\subsection{Grokking examples}\label{app:details grokking}

Throughout this section, $\beta$ denotes an output scale parameter that multiplies the weights of the final linear layer at initialisation. Concretely, if $W_{\mathrm{out}}$ is the last layer's weight matrix, it is initialised as $\beta \cdot \tilde{W}_{\mathrm{out}}$ where $\tilde{W}_{\mathrm{out}}$ follows the standard initialisation scheme. Large $\beta$ pushes the model into the rich (feature-learning) regime and tends to produce grokking; small $\beta$ keeps the model in the lazy (kernel) regime and suppresses it. To keep gradient magnitudes comparable across values of $\beta$, the optimizer learning rate is rescaled to $\eta = \eta_{\mathrm{cfg}} / \beta$, where $\eta_{\mathrm{cfg}}$ is the value specified in the config.

\subsubsection{Modular addition}\label{app:modular_addition}

\paragraph{Task.}
The task consists of predicting $(a + b) \bmod p$, where pairs $(a, b)$ with $a > b$, $a, b \in \{0, \ldots, p-1\}$ are fed to the model. For the Transformer, an additional constant cue token is appended to the input sequence, following~\citep{power2022grokking,nanda2023progress}. For both architectures we set $p = 113$, giving a total dataset of $p(p-1)/2 = 6{,}328$ unique pairs.

\paragraph{Architectures.}
The \textbf{MLP} takes the pair $(a, b)$ as input and sums their embeddings before passing them through two linear layers:
\begin{equation*}
    \mathbf{x} = \mathbf{e}_a + \mathbf{e}_b, \quad
    \mathbf{h} = \mathrm{ReLU}(W_{\mathrm{in}}\,\mathbf{x}), \quad
    \mathbf{z} = \mathrm{ReLU}(W_{\mathrm{out}}\,\mathbf{h}), \quad
    \hat{y} = W_{\mathrm{unemb}}\,\mathbf{z},
\end{equation*}
where $\mathbf{e}_a, \mathbf{e}_b \in \mathbb{R}^{256}$ are learned embeddings ($d_{\mathrm{emb}} = 256$), $W_{\mathrm{in}} \in \mathbb{R}^{128 \times 256}$, $W_{\mathrm{out}} \in \mathbb{R}^{256 \times 128}$, and $W_{\mathrm{unemb}} \in \mathbb{R}^{p \times 256}$.

The \textbf{Transformer} is a one-block decoder-only model with $d_{\mathrm{model}} = 256$, $4$ attention heads ($d_{\mathrm{head}} = 64$), and an MLP layer with hidden dimension $d_{\mathrm{mlp}} = 1024$ and ReLU activation. A single LayerNorm is applied to the final representation before the unembedding layer; there is no LayerNorm inside the attention or MLP sub-layers. The input sequence is $[a,\, b,\, \mathrm{sep}]$ with learned positional embeddings ($d_{\mathrm{vocab}} = p + 1 = 114$).

\paragraph{Training.}
All models are trained with AdamW~\citep{loshchilov2018decoupled}, cross-entropy loss, $(\beta_1, \beta_2) = (0.9, 0.98)$, weight decay $1.0$, batch size $200$, and a cosine learning rate schedule. Table~\ref{tab:modadd_hparams} summarises the per-condition hyperparameters; all other settings are shared.

\begin{table}[h]
\centering
\caption{Hyperparameters for the modular addition task.}
\label{tab:modadd_hparams}
\begin{tabular}{lcccc}
\toprule
 & \multicolumn{2}{c}{\textbf{MLP}} & \multicolumn{2}{c}{\textbf{Transformer}} \\
\cmidrule(lr){2-3}\cmidrule(lr){4-5}
 & Grokking & Non-grokking & Grokking & Non-grokking \\
\midrule
$\beta$                                                  & $1$                & $0.005$            & $2$                & $0.3$                \\
lr (config, $\eta_{\mathrm{cfg}}$)                       & $5 \times 10^{-4}$ & $5 \times 10^{-5}$ & $1 \times 10^{-3}$ & $1 \times 10^{-3}$   \\
lr (effective, $\eta_{\mathrm{cfg}}/\beta$)              & $5 \times 10^{-4}$ & $1 \times 10^{-2}$ & $5 \times 10^{-4}$ & $3.3 \times 10^{-3}$ \\
Epochs                                                   & $2{,}000$          & $1{,}000$          & $500$              & $500$                \\
Train samples                                            & $3{,}164\ (50\%)$  & $3{,}797\ (60\%)$  & $3{,}797\ (60\%)$  & $3{,}797\ (60\%)$    \\
\bottomrule
\end{tabular}
\end{table}

\subsubsection{Permutation composition}\label{app:perm_comp}

\paragraph{Task.}
The task consists of predicting the composition $\sigma \circ \tau$ of two permutations drawn from the symmetric group $S_5$, which has $5! = 120$ elements. The input is the ordered pair $(\sigma, \tau)$, giving a total dataset of $120^2 = 14{,}400$ pairs. For the Transformer, a constant cue token is appended to the input sequence.

\paragraph{Architectures.}
The \textbf{MLP} embeds each permutation separately ($d_{\mathrm{emb}} = 256$), concatenates the two embeddings, and passes the result through two linear layers with tied output weights:
\begin{equation*}
    \mathbf{x} = [\mathbf{e}_\sigma;\, \mathbf{e}_\tau], \quad
    \mathbf{h} = \mathrm{ReLU}(W_{\mathrm{in}}\,\mathbf{x}), \quad
    \mathbf{z} = W_{\mathrm{out}}\,\mathbf{h}, \quad
    \hat{y} = W_{\mathrm{emb}}^\top \mathbf{z},
\end{equation*}
where $\mathbf{x} \in \mathbb{R}^{512}$, $W_{\mathrm{in}} \in \mathbb{R}^{512 \times 512}$, $W_{\mathrm{out}} \in \mathbb{R}^{256 \times 512}$, and the output logits are produced via tied embedding weights $W_{\mathrm{emb}}^\top \in \mathbb{R}^{120 \times 256}$.

The \textbf{Transformer} shares the same architecture as the modular addition Transformer: a one-block decoder-only model with $d_{\mathrm{model}} = 256$, $4$ attention heads ($d_{\mathrm{head}} = 64$), $d_{\mathrm{mlp}} = 1024$, ReLU activation, and a single final LayerNorm. The input sequence is $[\sigma,\, \tau,\, \mathrm{sep}]$ with learned positional embeddings ($d_{\mathrm{vocab}} = 121$).

\paragraph{Training.}
All models are trained with AdamW~\citep{loshchilov2018decoupled}, cross-entropy loss, and $(\beta_1, \beta_2) = (0.9, 0.98)$. The MLP uses a cosine learning rate schedule; the Transformer uses a flat (constant) learning rate. Table~\ref{tab:perm_hparams} summarises the per-condition hyperparameters.

\begin{table}[h]
\centering
\caption{Hyperparameters for the permutation composition task.}
\label{tab:perm_hparams}
\begin{tabular}{lcccc}
\toprule
 & \multicolumn{2}{c}{\textbf{MLP}} & \multicolumn{2}{c}{\textbf{Transformer}} \\
\cmidrule(lr){2-3}\cmidrule(lr){4-5}
 & Grokking & Non-grokking & Grokking & Non-grokking \\
\midrule
$\beta$                                                  & $1$                & $0.01$             & $1$                & $0.1$                \\
lr (config, $\eta_{\mathrm{cfg}}$)                       & $1 \times 10^{-3}$ & $1 \times 10^{-5}$ & $1 \times 10^{-3}$ & $1 \times 10^{-3}$   \\
lr (effective, $\eta_{\mathrm{cfg}}/\beta$)              & $1 \times 10^{-3}$ & $1 \times 10^{-3}$ & $1 \times 10^{-3}$ & $1 \times 10^{-2}$   \\
Weight decay                                             & $1.0$              & $1.0$              & $1.0$              & $0.01$               \\
Batch size                                               & $2{,}000$          & $2{,}000$          & $200$              & $200$                \\
Scheduler                                                & cosine             & cosine             & flat               & flat                 \\
Epochs                                                   & $10{,}000$         & $1{,}000$          & $400$              & $1{,}000$            \\
Train samples                                            & $6{,}000\ (42\%)$  & $10{,}800\ (75\%)$ & $5{,}760\ (40\%)$  & $10{,}800\ (75\%)$   \\
\bottomrule
\end{tabular}
\end{table}

\subsubsection{Sparse parity}\label{app:sparse_parity}

\paragraph{Task.}
The sparse parity task is a binary classification problem over binary vectors $\mathbf{x} \in \{{\pm}1\}^n$ with $n = 40$. The label is the parity of $k = 3$ fixed bits from a randomly chosen support set $S \subset \{1,\ldots,n\}$ with $|S| = k$: $y = \prod_{i \in S} x_i$. The support set is fixed across all examples and is unknown to the model.

\paragraph{Architectures.}
The \textbf{MLP} is a single hidden-layer network with ReLU activation:
\begin{equation*}
    \hat{y} = W_{\mathrm{out}}\,\mathrm{ReLU}(W_{\mathrm{in}}\,\mathbf{x}),
\end{equation*}
where $W_{\mathrm{in}} \in \mathbb{R}^{1000 \times 40}$ and $W_{\mathrm{out}} \in \mathbb{R}^{1 \times 1000}$ (no output bias). The model produces a scalar logit.

The \textbf{Transformer} follows the architecture of~\citep{barak2022hidden}. It is a single-block encoder with a CLS token at position $0$ followed by the $40$ bit positions ($n_{\mathrm{ctx}} = 41$, $d_{\mathrm{vocab}} = 3$). The model has $d_{\mathrm{model}} = 1024$, with $128$ attention heads each of per-head dimension $d_{\mathrm{attn}} = 8$ and an MLP layer with $d_{\mathrm{mlp}} = 1024$ and GELU activation. There is no LayerNorm. Attention is CLS-only: only the CLS token produces queries, attending over all positions; its updated representation is passed to the MLP and unembedding layers.

\paragraph{Training.}
The MLP is trained with SGD and hinge loss; the Transformer is trained with AdamW \citep{loshchilov2018decoupled} ($(\beta_1, \beta_2) = (0.9, 0.98)$) and cross-entropy loss. The Transformer uses a linear warmup of 10 steps followed by a constant learning rate. Since the MLP and Transformer use different optimizers and loss functions, the full per-condition configuration is listed in Table~\ref{tab:sparsity_hparams}.

\begin{table}[h]
\centering
\caption{Hyperparameters for the sparse parity task. Training sample counts have no natural percentage since examples are drawn i.i.d.\ from $\{{\pm}1\}^{40}$.}
\label{tab:sparsity_hparams}
\begin{tabular}{lcccc}
\toprule
 & \multicolumn{2}{c}{\textbf{MLP}} & \multicolumn{2}{c}{\textbf{Transformer}} \\
\cmidrule(lr){2-3}\cmidrule(lr){4-5}
 & Grokking & Non-grokking & Grokking & Non-grokking \\
\midrule
$\beta$                                                  & $1$    & $0.1$  & $1$                & $1$                  \\
Optimizer                                                & SGD    & SGD    & AdamW              & AdamW                \\
Loss                                                     & hinge  & hinge  & cross-entropy      & cross-entropy        \\
lr (config, $\eta_{\mathrm{cfg}}$)                       & $0.1$  & $0.1$  & $1 \times 10^{-3}$ & $5 \times 10^{-4}$   \\
lr (effective, $\eta_{\mathrm{cfg}}/\beta$)              & $0.1$  & $1.0$  & $1 \times 10^{-3}$ & $5 \times 10^{-4}$   \\
Weight decay                                             & $0.01$ & $0.01$ & $0.1$              & $0.01$               \\
Batch size                                               & $50$   & $50$   & $512$              & $512$                \\
Epochs                                                   & $2{,}000$ & $2{,}000$ & $3{,}000$       & $1{,}000$            \\
Train samples                                            & $1{,}200$ & $5{,}000$ & $1{,}000$       & $5{,}000$            \\
\bottomrule
\end{tabular}
\end{table}

\subsubsection{MNIST digit classification}\label{app:mnist}

\paragraph{Task.}
The task is 10-class classification of handwritten digit images from the MNIST dataset~\citep{lecun1998gradient}. Each input is a $28 \times 28$ grayscale image (784 features) and the label is one of 10 digit classes. The full MNIST training set contains 60,000 images; both the grokking and non-grokking conditions use a random subset, with the test set subsampled to the same size as the training set.

\paragraph{Architecture.}
A three-layer MLP with two hidden layers of width 200 and ReLU activations:
\begin{equation*}
    \hat{y} = W_3\,\mathrm{ReLU}(W_2\,\mathrm{ReLU}(W_1\,\mathbf{x})),
\end{equation*}
where $W_1 \in \mathbb{R}^{200 \times 784}$, $W_2 \in \mathbb{R}^{200 \times 200}$, $W_3 \in \mathbb{R}^{10 \times 200}$. At initialisation all weight matrices are scaled globally by a factor $\alpha$ (the \emph{initialization scale}):
\begin{equation*}
    W_i \leftarrow \alpha \cdot \tilde{W}_i, \quad \tilde{W}_i \sim \text{standard init.}
\end{equation*}
Unlike the other tasks, the lazy--rich axis here is controlled by $\alpha$ applied to \emph{all} layers, not by the output-layer $\beta$ parameter. Accordingly $\beta = 1$ for both conditions and the lr-rescaling $\eta = \eta_{\mathrm{cfg}}/\beta$ does not apply.

\paragraph{Training.}
All models are trained with AdamW~\citep{loshchilov2018decoupled}, MSE loss on one-hot targets, $(\beta_1, \beta_2) = (0.9, 0.999)$, learning rate $1 \times 10^{-3}$, weight decay $2 \times 10^{-4}$, batch size $200$, and a constant learning rate schedule (no scheduler). Table~\ref{tab:mnist_hparams} summarises the per-condition hyperparameters.

\begin{table}[h]
\centering
\caption{Hyperparameters for the MNIST digit classification task.}
\label{tab:mnist_hparams}
\begin{tabular}{lcc}
\toprule
 & Grokking & Non-grokking \\
\midrule
Initialization scale ($\alpha$) & $6.0$              & $1.0$                \\
Epochs                          & $200{,}000$        & $1{,}000$            \\
Train samples                   & $1{,}000\ (1.7\%)$ & $10{,}000\ (16.7\%)$ \\
\bottomrule
\end{tabular}
\end{table}

\subsection{Double descent example}\label{app:details dd}

\paragraph{Task.}
The task is 10-class image classification on CIFAR-10~\citep{krizhevsky2009learning}, which contains 50{,}000 training images and 10{,}000 test images of size $32 \times 32 \times 3$. We reproduce the epoch-wise double descent phenomenon of Nakkiran et al.~\citep{nakkiran2021deep}: test error first decreases, then increases, and finally decreases again as training continues. To induce this behavior, a fraction of training labels are corrupted: each noisy example is assigned a uniformly random incorrect class, with the corruption fixed at the start of training (the same wrong label is used across all augmented views). We consider three noise levels: $0\%$, $10\%$, and $20\%$.

\paragraph{Architecture.}
We use a ResNet18 adapted for CIFAR-10 spatial resolution. The standard conv1 layer ($7{\times}7$, stride 2) is replaced by a $3{\times}3$, stride-1 convolution, and the max-pool layer is replaced by an identity map, so the feature map resolution is preserved at $32{\times}32$ through the first residual group. The spatial flow is $32{\to}32{\to}16{\to}8{\to}4{\to}1$ (with average pooling), yielding a $512$-dimensional pre-classifier feature vector followed by a linear output head.

\paragraph{Training.}
Models are trained with Adam~\citep{kingma2014adam} ($\beta_1 = 0.9$, $\beta_2 = 0.999$, $\varepsilon = 10^{-8}$), cross-entropy loss, learning rate $10^{-4}$ (fixed, no schedule), no weight decay, and batch size $128$. Training runs for $4{,}000$ epochs. Data augmentation consists of random crops (padding 4) and random horizontal flips; no augmentation is applied at test time. Checkpoints are saved at $200$ log-uniformly spaced epochs.

\begin{table}[h]
\centering
\caption{Hyperparameters for the double descent task.}
\label{tab:dd_hparams}
\begin{tabular}{lc}
\toprule
 & \textbf{ResNet18} \\
\midrule
Optimizer                          & Adam \\
$(\beta_1, \beta_2)$               & $(0.9,\; 0.999)$ \\
Learning rate                      & $1 \times 10^{-4}$ (fixed) \\
Weight decay                       & $0$ \\
Batch size                         & $128$ \\
Epochs                             & $4{,}000$ \\
Noise levels                       & $0\%,\; 10\%,\; 20\%$ \\
Train samples                      & $50{,}000$ \\
\bottomrule
\end{tabular}
\end{table}

\subsection{Computer resources}\label{app:computer}

All model training was run on a single NVIDIA A100 GPU. For small models trained over a few hundred epochs (e.g., ModAdd Transformer at 500 epochs, PermComp Transformer at 400 epochs), each individual training run takes approximately 1–5 minutes. For longer runs (e.g., PermComp MLP grokking at 10,000 epochs, MNIST grokking at 200,000 epochs), training takes roughly 30–60 minutes per run. With 3 seeds × 2 conditions (grokking/non-grokking) across 6 task×architecture pairs, total training compute is on the order of 10–20 GPU-hours. Post-training analysis (metric computation, GLUE manifold analysis, linear probes) runs entirely on AMD EPYC CPU. GLUE computation takes approximately 1 minute per checkpoint per run; across about 200 log-uniformly spaced checkpoints per run, this amounts to roughly 3 CPU-hours per run, parallelized across up to 3,000 CPU cores. Total analysis compute is on the order of 100–200 CPU-hours.

\section{Further Discussion of Previous Work}\label{app:further discussion}
We provide further discussion of how our framework relates to four broad threads of prior work on grokking, as well as background on GLUE theory and its previous applications (\autoref{app:GLUE disc}). We address the lazy-rich hypothesis~(\autoref{app:lazy-rich disc}), which frames grokking as a transition between kernel regression and feature learning regimes. We discuss task-specific progress measures~(\autoref{app:task-specific disc}), which construct mechanistically interpretable grokking indicators, and contrast their approach with our task-agnostic framework. We compare our measures against other task-agnostic progress measures proposed in the literature~(\autoref{app:other measure disc}), highlighting key differences in interpretability, architectural generality, and empirical robustness. Finally, we review weight-based theories of grokking~(\autoref{app:weight-based disc}), discussing both the empirical counterexamples that challenge them and their theoretical limitations as a lens for understanding generalization.

\subsection{Relation to lazy-rich hypothesis}\label{app:lazy-rich disc}

The lazy-rich hypothesis by Kumar et al.~\citep{kumar2023grokking} framed grokking as a transition from kernel regression dynamics with a fixed Neural Tangent Kernel (NTK) to genuine feature learning, in which the kernel itself evolves to better align with the task. Two prior results of note to our work relate to this framing. Firstly, Zheng et al.~\citep{zheng2024delays} reported evidence that conflicts with the network behaving as kernel regression with $K_\mathrm{NTK}$ fixed at initialization before grokking, observing that the kernel distance between the NTK at different epochs, defined for kernel matrices $K_1,K_2$ as $1-\frac{\langle K_{1},K_2\rangle_F}{\|K_1\|_F\|K_2\|_F}$, varies meaningfully prior to grokking onset. Secondly, Kumar et al.~\citep{kumar2023grokking} reported a monotone increase in NTK-label alignment over training in their toy setting of polynomial regression.

Our findings engage with both of these results. Consistent with Zheng et al.~\citep{zheng2024delays}, we find that both representation learning and readout calibration are actively progressing before grokking onset, hence the learning dynamics do not resemble lazy learning in our setting, providing additional evidence against the fixed-kernel picture (i.e., lazy learning) of pre-grokking training. However, we find a more nuanced picture of the NTK-alignment dynamics reported by Kumar et al.~\citep{kumar2023grokking}: across our experiments, we observe both monotonically increasing NTK-alignment as well as non-monotonic behavior of NTK-alignment over training (e.g.,~\autoref{fig:all grok permcomp mlp},~\autoref{fig:all grok sparse transformer}).

\subsection{Relation to task-specific grokking analyses}\label{app:task-specific disc}

A complementary line of work has developed task-specific progress measures for grokking by exploiting the algebraic structure of particular tasks. Barak et al.~\citep{barak2022hidden} introduced the notion of progress measures in the grokking context—scalar quantities that vary more smoothly over training than loss or accuracy, and thus provide earlier and more continuous signal about a network's progress toward generalization. They instantiated this idea for the sparse parity task using a Fourier gap measure derived from the task's Boolean structure. Subsequently, Nanda et al.~\citep{nanda2023progress} and Chughtai et al.~\citep{chughtai2023toymodeluniversalityreverse} identified analogous progress measures for modular addition and group composition respectively, in each case leveraging the specific algebraic structure of the task to construct interpretable, mechanistically grounded measures.

Our metrics perform well in revealing the hidden progress of the network across all of these settings without requiring any task-specific analysis or structural assumptions. A point of nuance is that our measures are less directly task-interpretable than their metrics, which admit clean mechanistic readings tied to the learned algorithm. We view this as a natural cost of universality: by not assuming any task structure, our metrics sacrifice some interpretability in exchange for broad applicability and ease of implementation, making them better suited to practical settings where task-specific decompositions are unavailable or prohibitively expensive to derive.

\subsection{Relation to other task-agnostic progress measures}\label{app:other measure disc}
Several recent works have proposed task-agnostic progress measures for grokking that, like ours, do not rely on task-specific structural assumptions. Humayun et al.~\citep{humayun2024deep} introduced \emph{local complexity}, a measure of the density of linear regions in a network's input-space partition, and showed that it consistently exhibits a descent-ascent-descent pattern across grokking examples. 
While task-agnostic in principle, the measure is sensitive to hyperparameter choices that must be tuned per architecture and is restricted to networks with continuous piecewise linear activation functions~\cite[Appendix C]{humayun2024deep}. 
Merrill \& Tsilivis et al.~\citep{merrill2023tale} proposed weight sparsity as a progress measure, observing that the sparsification of weights contributing to the network output co-occurs with the onset of grokking. Clauw et al.~\citep{clauw2024information} applied O-Information---an entropy-based measure of higher-order statistical dependencies---to model activations, identifying training phases that correspond to the qualitative behavior of the train and test loss curves during grokking.

Our framework differs from these approaches primarily in interpretability. While the measures described above are effective at empirically detecting grokking, none situates its findings within a broader theoretical framework that explains why generalization is delayed. For instance, Merrill \& Tsilivis et al.~\citep{merrill2023tale} do not ground weight sparsity in underlying principles connecting it to the network's learning dynamics. By contrast, our measures are grounded in the geometry of learned representations and connect directly to feature learning theory~\citep{chou2025featurelearninglazyrichdichotomy}, providing an interpretable account of the representational changes that underlie the transition to generalization.

\subsection{Relation to weight-based theories of grokking}\label{app:weight-based disc}
Weight-based analyses are a natural starting point for understanding neural network behavior, since the weights fully parameterize the learned function~\citep{neyshabur2017exploringgeneralizationdeeplearning,nagarajan2019generalizationdeepnetworksrole,bartlett2017spectrallynormalizedmarginboundsneural}. Within the grokking literature, two influential weight-based accounts have been proposed. Liu et al.~\citep{liu2022towards} argued that grokking arises from the dynamics of weights under high initial weight norm: weight decay slowly pushes the network toward a regime of smaller weight norm, with delayed generalization explained by the fact that large weight norm enables overfitting while reducing the marginal penalty of training loss, slowing the transition to generalizing solutions. Varma \& Shah et al.~\citep{varma2023explaining} offered a complementary mechanistic account, framing grokking as a competition between two circuits with equivalent training loss: a memorization circuit and a slower-to-learn generalizing circuit that achieves lower test loss with smaller weight norm. Under their framework, weight decay creates pressure that eventually favors the more parameter-efficient generalizing circuit, producing the characteristic delay.

More recent work has identified counterexamples to weight-based accounts of grokking. Kumar et al.~\citep{kumar2023grokking} demonstrate grokking on modular addition trained without weight decay, observing weight norm \emph{increasing} during the transition to generalization, and Golechha et al.~\citep{golechha2024progress} reported similar weight norm dynamics for MNIST under a modified loss function. These results suggest that weight norm reduction is neither necessary nor universally diagnostic of grokking.

Beyond this empirical evidence against weight-norm-centric explanations, weight-based theories are by construction blind to the structure of the network's learned representations—the geometric organization of features that ultimately determines what the network has learned. Our framework directly probes this representational structure, providing a complementary lens that captures aspects of grokking dynamics that weight-based analyses cannot. 

A further distinction between our framework and existing weight-based accounts of grokking lies in their respective theoretical foundations. While seminal work on generalization~\citep{neyshabur2017exploringgeneralizationdeeplearning,nagarajan2019generalizationdeepnetworksrole,bartlett2017spectrallynormalizedmarginboundsneural} provides rigorous complexity bounds, these often involve nuanced spectral or margin-based properties that extend beyond the simple weight-norm heuristics prioritized in current grokking literature~\citep{liu2022towards,varma2023explaining}. Furthermore, because scalar weight changes do not necessarily correspond to functional changes—for instance, under simple reparameterizations—a more mathematically principled treatment of network dynamics is required. Our approach addresses this by utilizing representational measures with strong theoretical backing in Neural Tangent Kernel (NTK) theory~\citep{jacot2020neuraltangentkernelconvergence} and GLUE theory (see \autoref{app:GLUE disc}), offering a robust lens that captures aspects of the transition to generalization that weight-based analyses may overlook.

\subsection{Previous applications of GLUE framework}\label{app:GLUE disc}
The Geometry Linked to Untangling Efficiency (GLUE) theory~\citep{chou2025glue}, which serves as an extension of earlier work that developed Manifold Capacity Theory~\citep{chung2018classification,wakhloo2023linear,mignacco2025nonlinear,chou2025glue,slatton2026linear}, provides a suite of geometric measures for characterizing the structure of learned representations, and has been applied successfully to two problem settings closely related to ours. Chou \& Le et al.~\citep{chou2025featurelearninglazyrichdichotomy} showed that the critical dimension — which is inversely proportional to the capacity metric they use — tracks the network's feature learning, and that GLUE geometry measures more broadly are effective at classifying subtypes of feature learning dynamics. Chou et al.~\citep{choudiagnosing} extended this line of work to out-of-distribution generalization, demonstrating that a collection of GLUE measures computed on in-distribution features from the model robustly predicts model performance on OOD data, outperforming a broad set of alternative measures in their comparison. Beyond ML applications, manifold capacity theory has been applied to study representations in biological neural networks, including the macaque visual cortex~\citep{kuoch2024probing}, social learning~\citep{paraouty2023sensory}, and olfactory processing circuits~\citep{hu2024representational}, demonstrating its utility as a general framework for characterizing representational structure across both artificial and biological systems.

These results provide empirical motivation for our use of critical dimension and GLUE geometry measures to probe the structure of learned representations during grokking. In particular, the demonstrated ability of these measures to discriminate between feature learning regimes and to predict generalization in OOD settings suggests they are well-suited to capturing the representational changes that underlie the delayed generalization characteristic of grokking.

\section{Additional Results on Diagnostic Measures}\label{app:all measures}
We provide additional results for all six task-model pairs in both grokking (\autoref{app:all grok}) and non-grokking (\autoref{app:all nogrok}) settings across three seeds each. Definitions for the shaded regions used throughout the main text are given in~\autoref{app:event detection}.

\subsection{Grokking examples}\label{app:all grok}
\begin{figure}[h!]
    \centering
    \includegraphics[width=\linewidth]{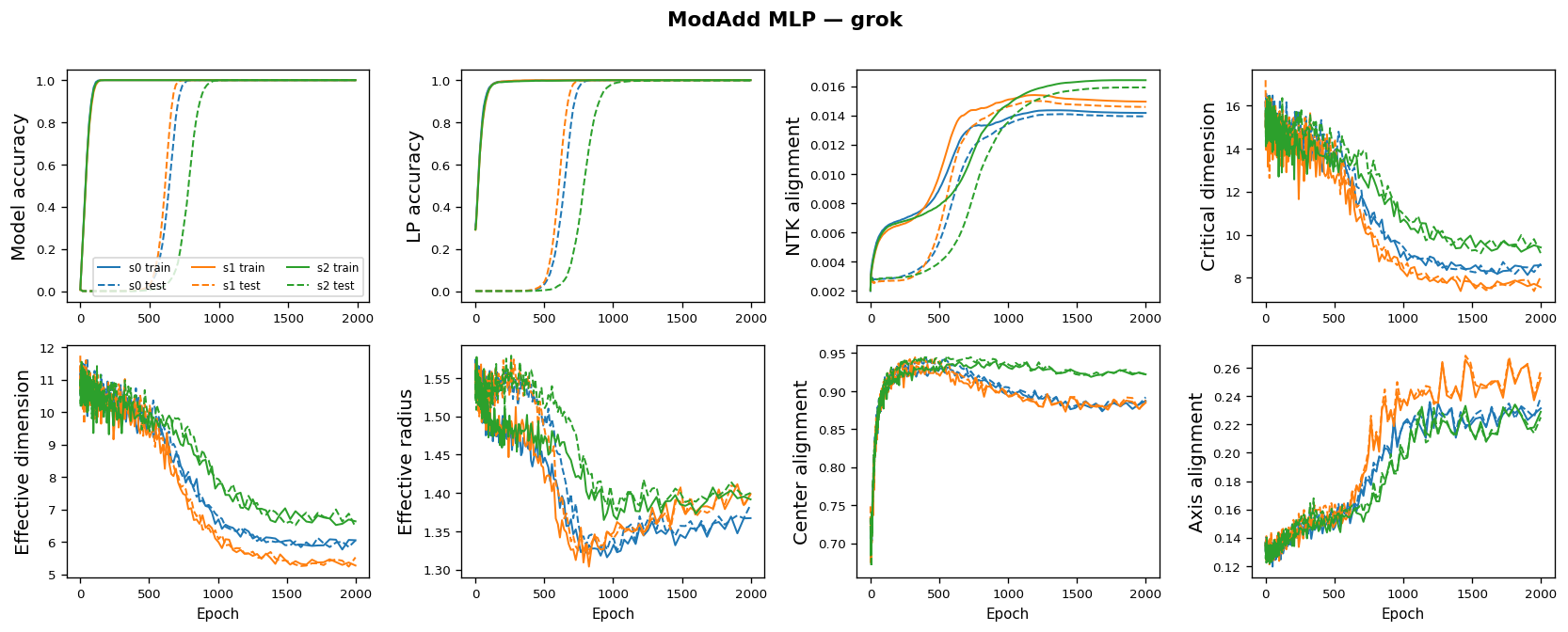}
    \caption{Grokking examples of ModAdd MLP.}
    \label{fig:all grok modadd mlp}
\end{figure}

\begin{figure}[h!]
    \centering
    \includegraphics[width=\linewidth]{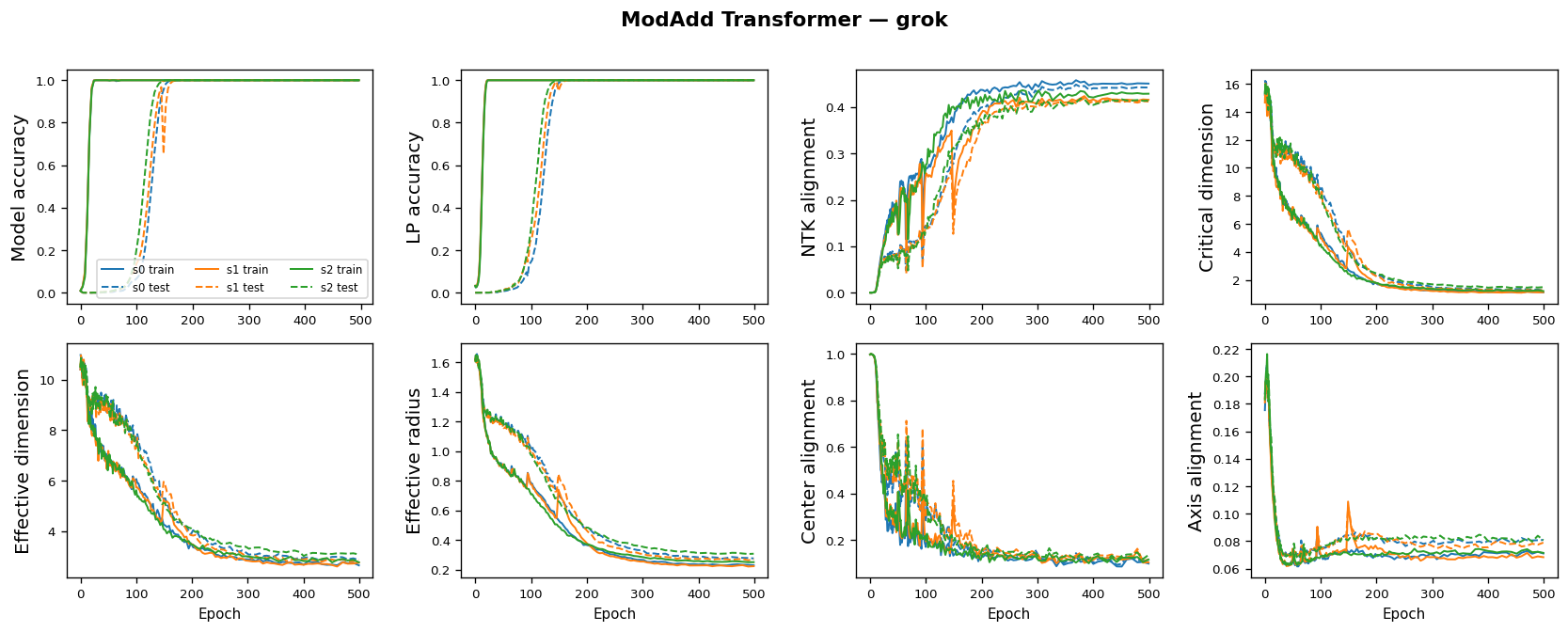}
    \caption{Grokking examples of ModAdd Transformer.}
    \label{fig:all grok modadd transformer}
\end{figure}

\begin{figure}[h!]
    \centering
    \includegraphics[width=\linewidth]{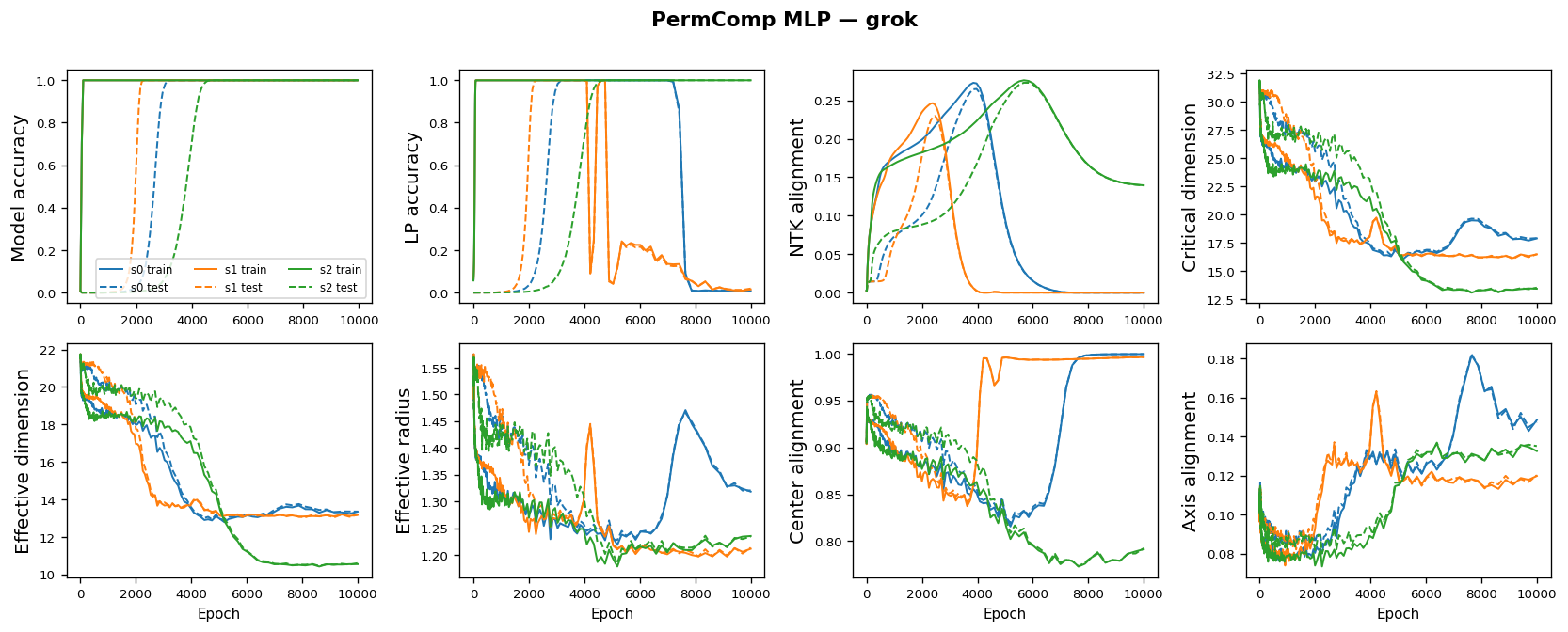}
    \caption{Grokking examples of PermComp MLP.}
    \label{fig:all grok permcomp mlp}
\end{figure}

\begin{figure}[h!]
    \centering
    \includegraphics[width=\linewidth]{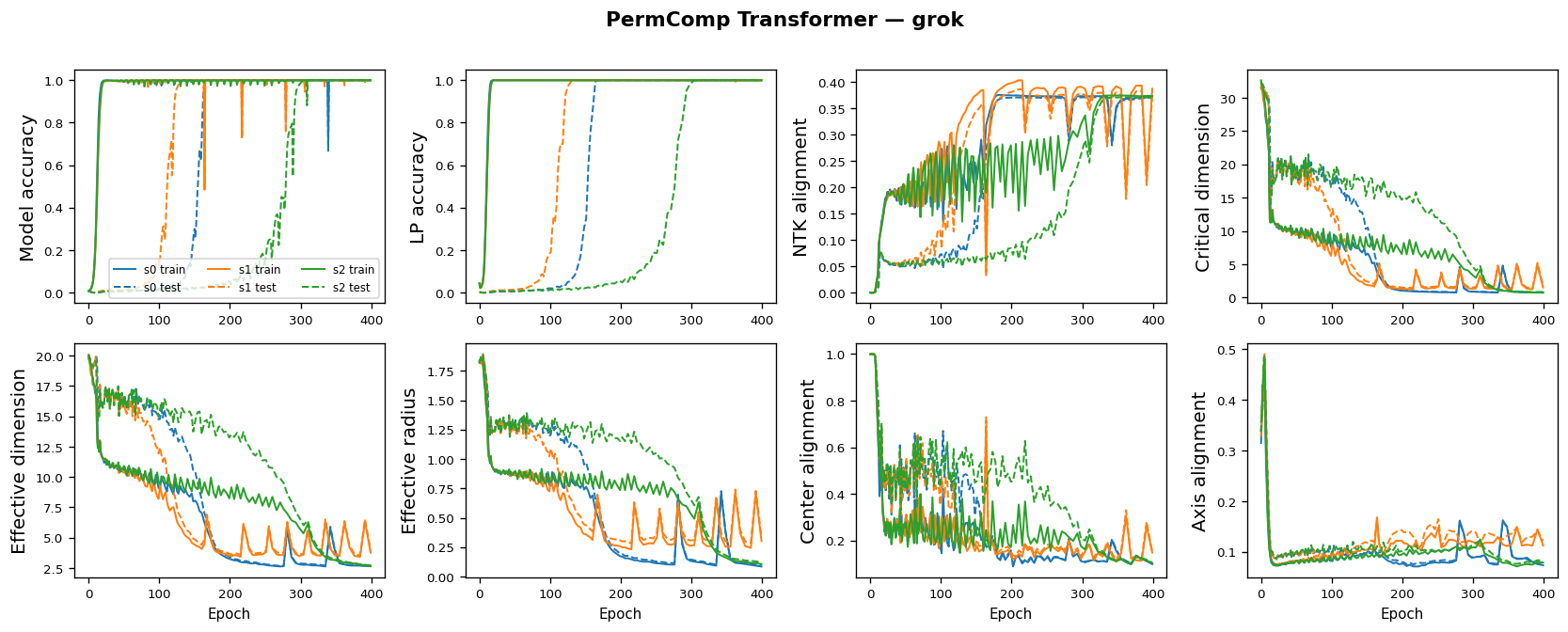}
    \caption{Grokking examples of PermComp Transformer.}
    \label{fig:all grok permcomp transformer}
\end{figure}

\begin{figure}[h!]
    \centering
    \includegraphics[width=\linewidth]{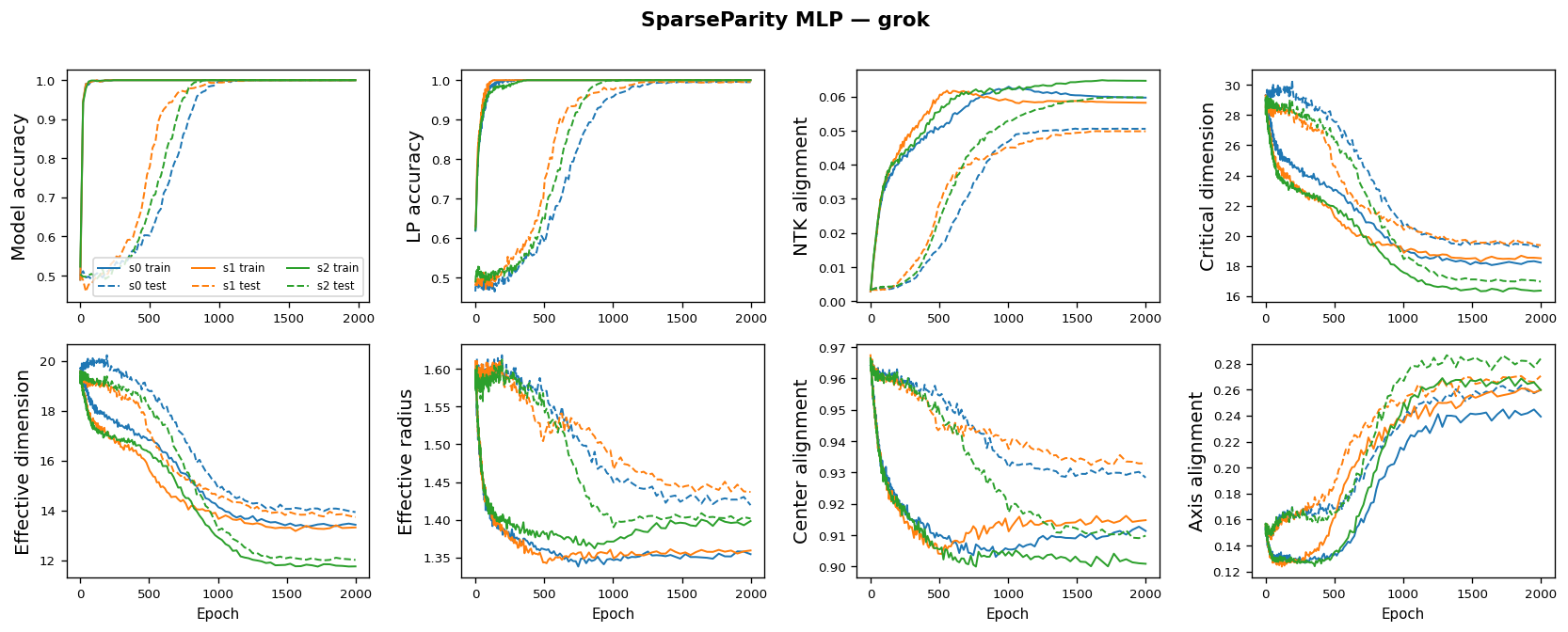}
    \caption{Grokking examples of SparseParity MLP.}
    \label{fig:all grok sparse mlp}
\end{figure}

\begin{figure}[h!]
    \centering
    \includegraphics[width=\linewidth]{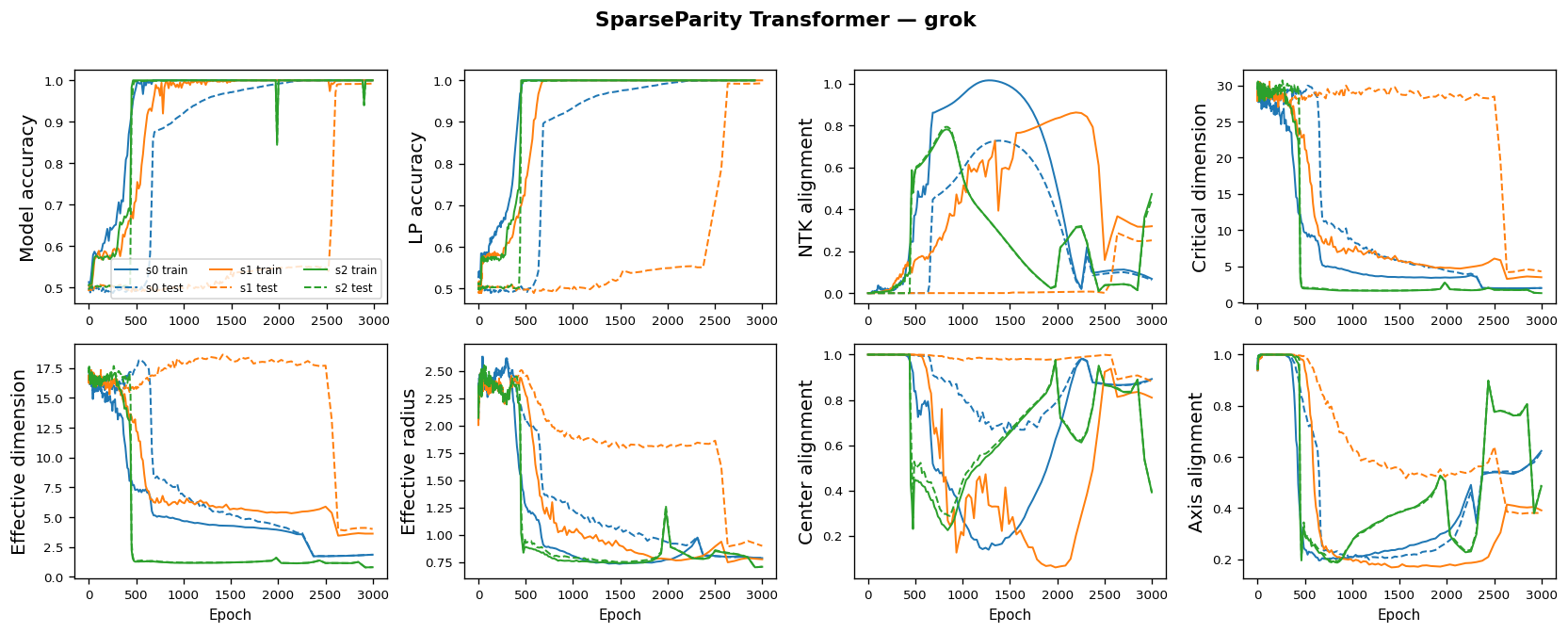}
    \caption{Grokking examples of SparseParity Transformer.}
    \label{fig:all grok sparse transformer}
\end{figure}

We present grokking examples for all six task-model pairs across three seeds each (\autoref{fig:all grok modadd mlp}--\autoref{fig:all grok sparse transformer}). For each run, we plot train and test accuracy alongside the critical dimension $N_\text{crit}$ and the four-phase shading defined in \autoref{app:event detection}. Across all settings, $N_\text{crit}$ decreases gradually throughout training, consistent with the continuous representation learning reported in \autoref{sec:results beyond lazy-to-rich}. One exception is the SparseParity Transformer, where grokking is observed in only one of three seeds (seed=1); the remaining seeds do not exhibit a clear grokking pattern and are included for completeness.

\subsection{Non-grokking examples}\label{app:all nogrok}
\begin{figure}[h!]
    \centering
    \includegraphics[width=\linewidth]{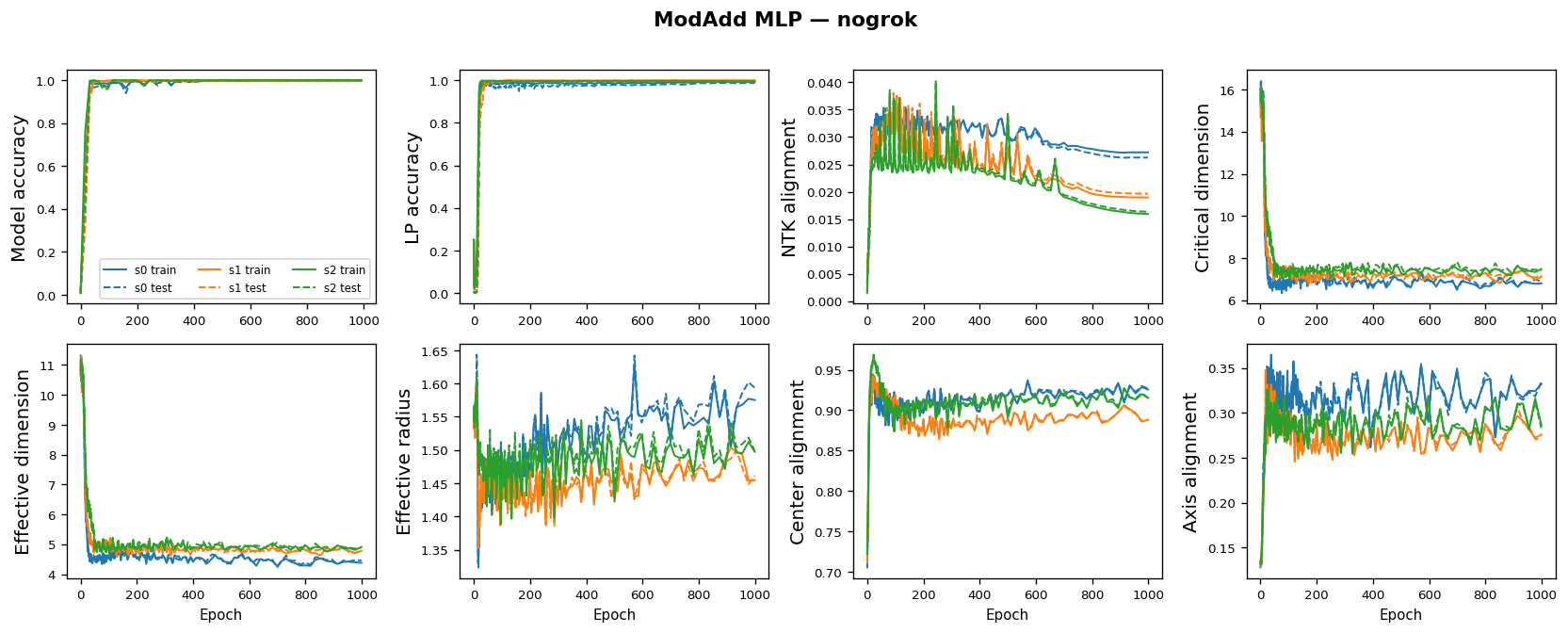}
    \caption{Non-grokking examples of ModAdd MLP.}
    \label{fig:all nogrok modadd mlp}
\end{figure}

\begin{figure}[h!]
    \centering
    \includegraphics[width=\linewidth]{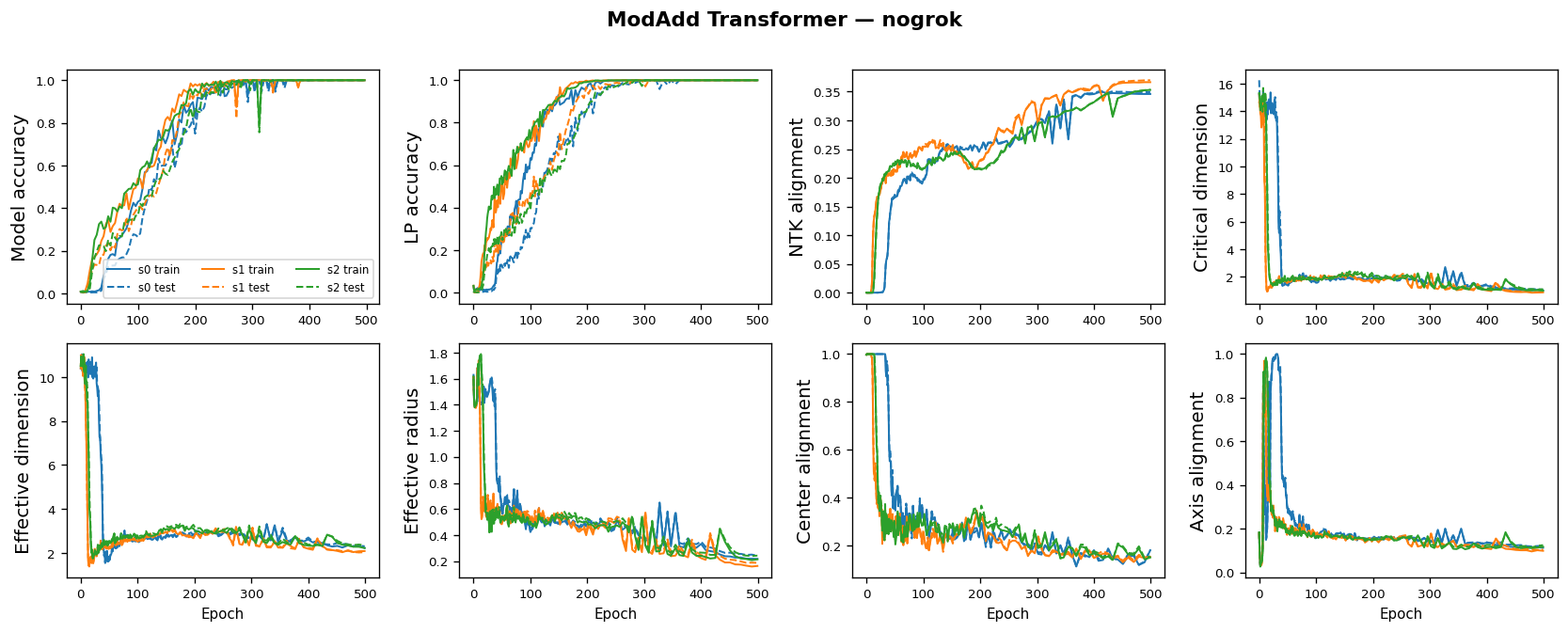}
    \caption{Non-grokking examples of ModAdd Transformer.}
    \label{fig:all nogrok modadd transformer}
\end{figure}

\begin{figure}[h!]
    \centering
    \includegraphics[width=\linewidth]{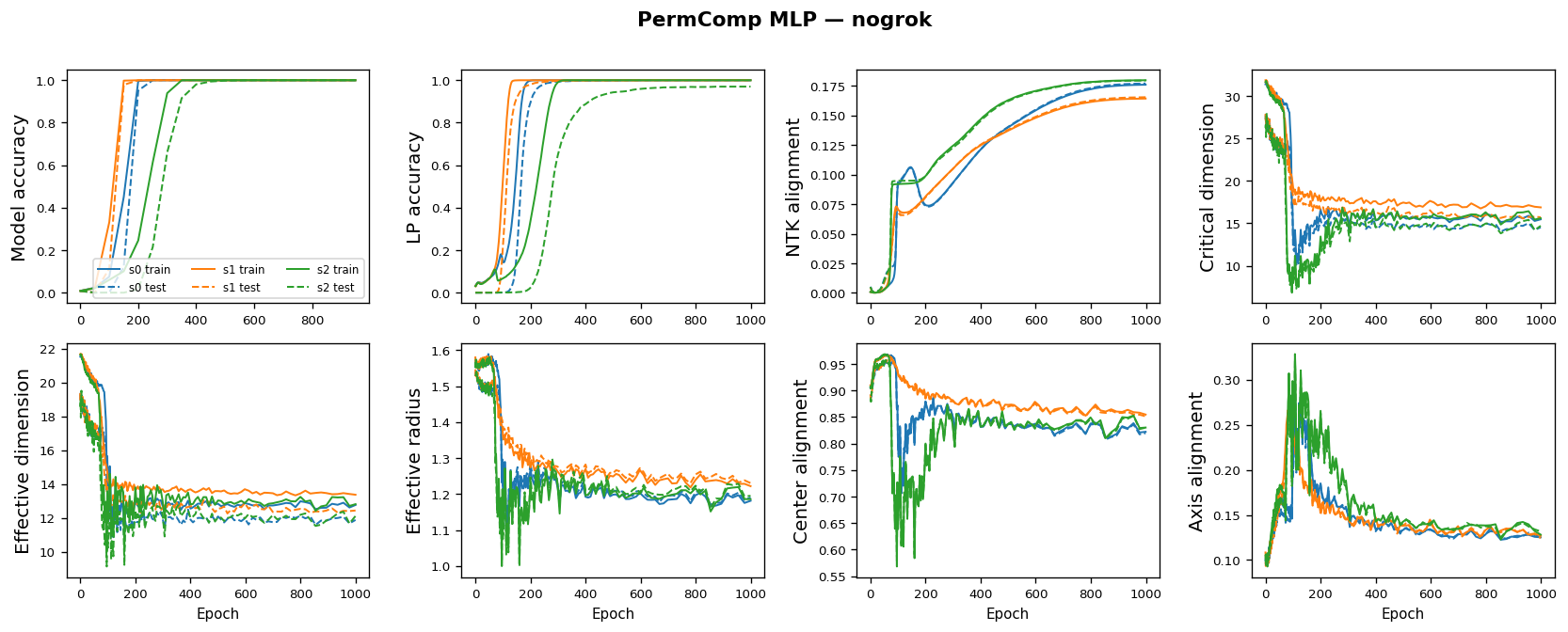}
    \caption{Non-grokking examples of PermComp MLP.}
    \label{fig:all nogrok permcomp mlp}
\end{figure}

\begin{figure}[h!]
    \centering
    \includegraphics[width=\linewidth]{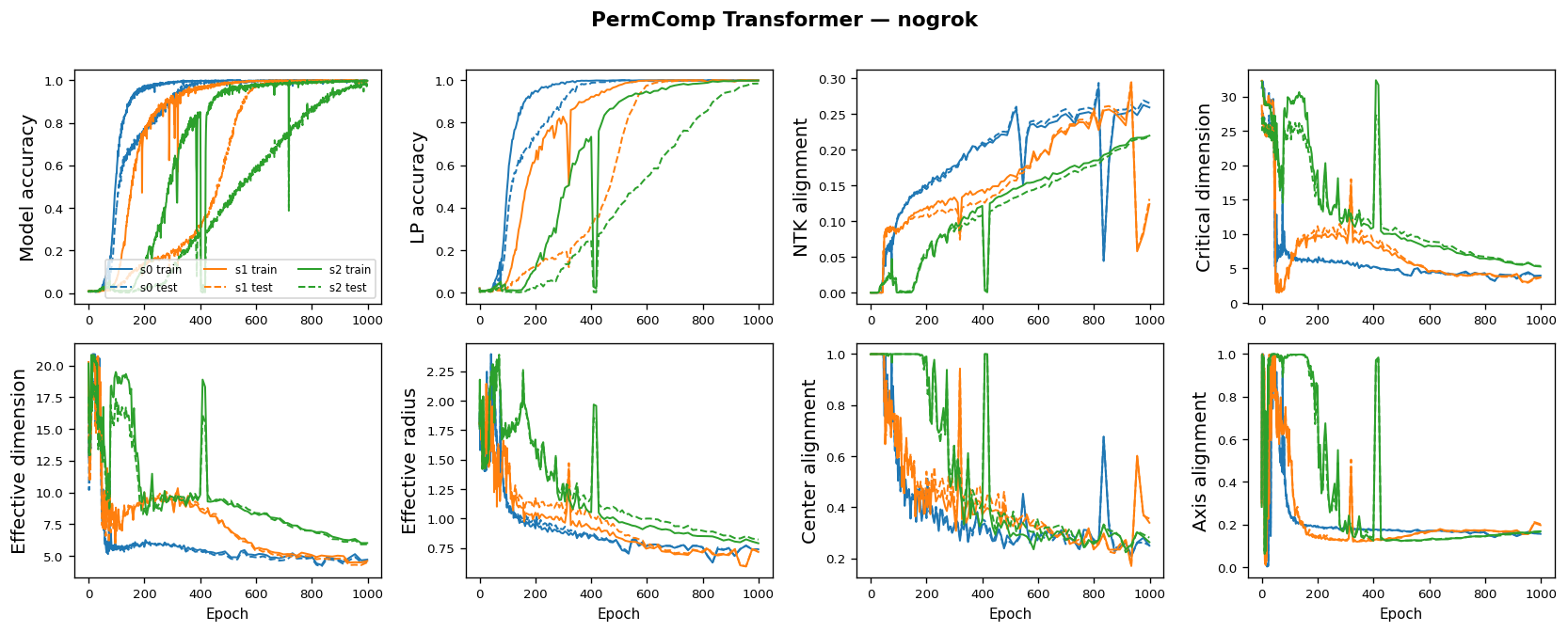}
    \caption{Non-grokking examples of PermComp Transformer.}
    \label{fig:all nogrok permcomp transformer}
\end{figure}

\begin{figure}[h!]
    \centering
    \includegraphics[width=\linewidth]{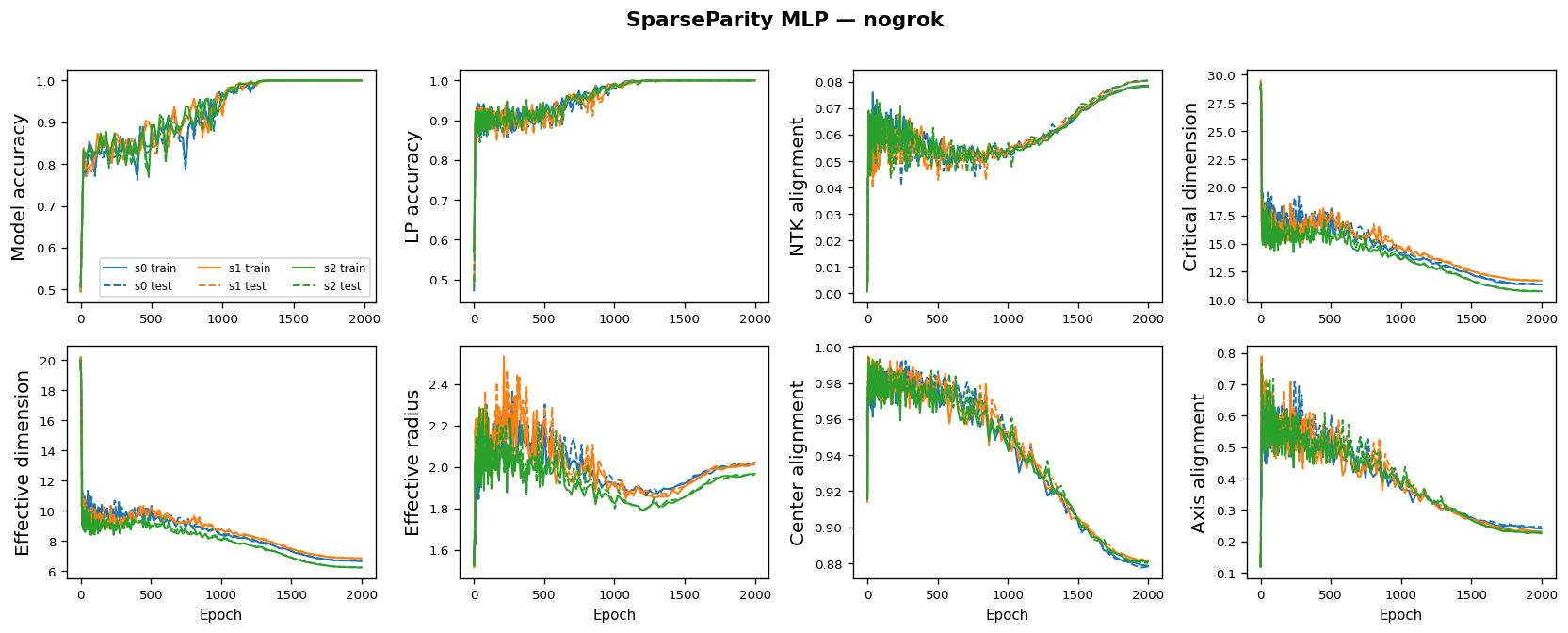}
    \caption{Non-grokking examples of SparseParity MLP.}
    \label{fig:all nogrok sparse mlp}
\end{figure}

\begin{figure}[h!]
    \centering
    \includegraphics[width=\linewidth]{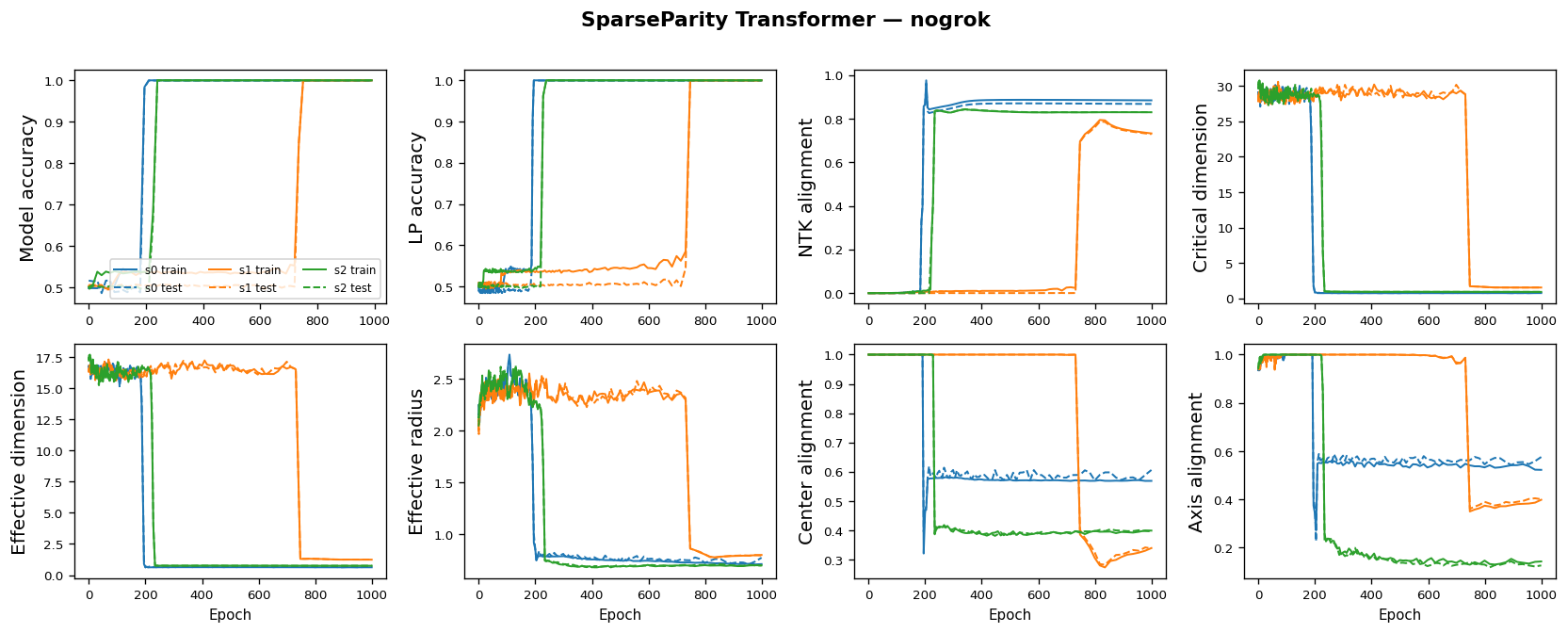}
    \caption{Non-grokking examples of SparseParity Transformer.}
    \label{fig:all nogrok sparse transformer}
\end{figure}

We present non-grokking counterparts for all six task-model pairs across three seeds each (\autoref{fig:all nogrok modadd mlp}--\autoref{fig:all nogrok sparse transformer}). These are obtained by modifying the training recipe to eliminate the grokking pattern (see \autoref{app:event detection} for details).

\subsection{Epoch-wise double descent through label noises}\label{app:dd}
\begin{figure}[h!]
    \centering
    \includegraphics[width=\linewidth]{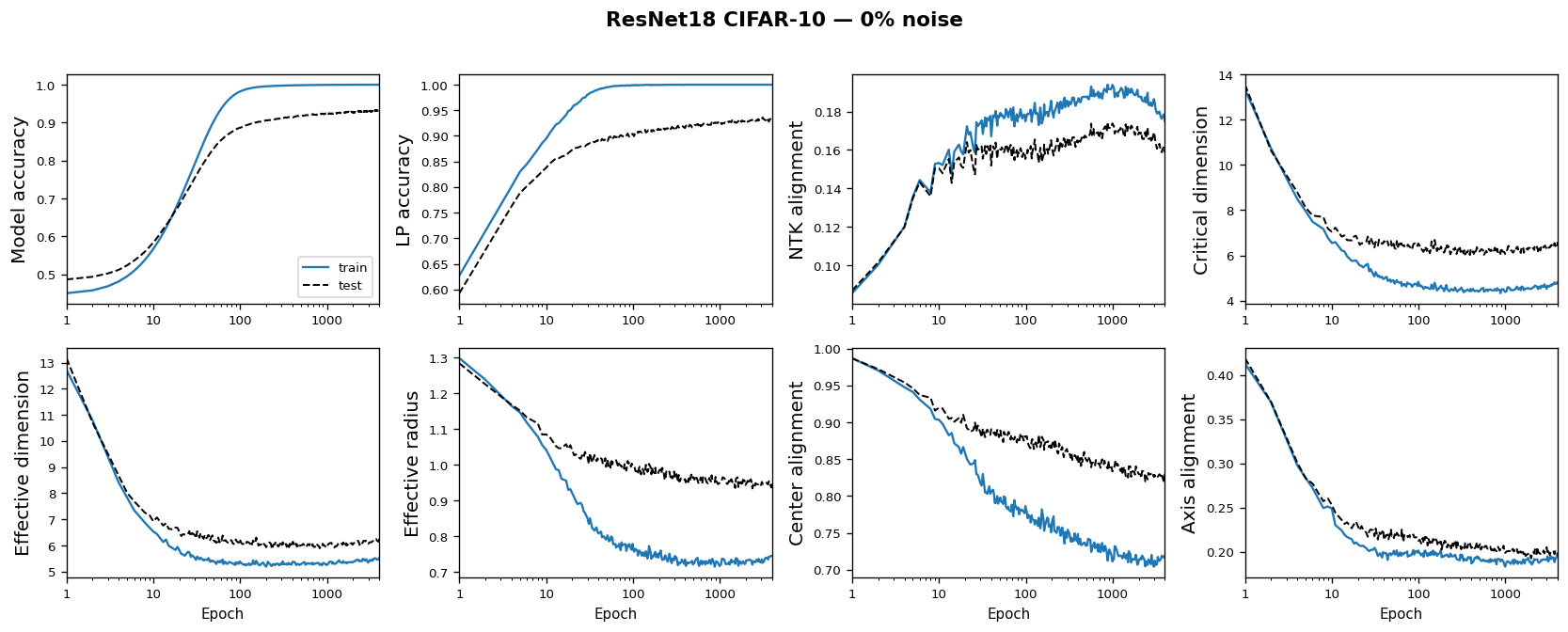}
    \caption{Epoch-wise double descent through label noises, 0\% label noise.}
    \label{fig:all dd 0 noise}
\end{figure}

\begin{figure}[h!]
    \centering
    \includegraphics[width=\linewidth]{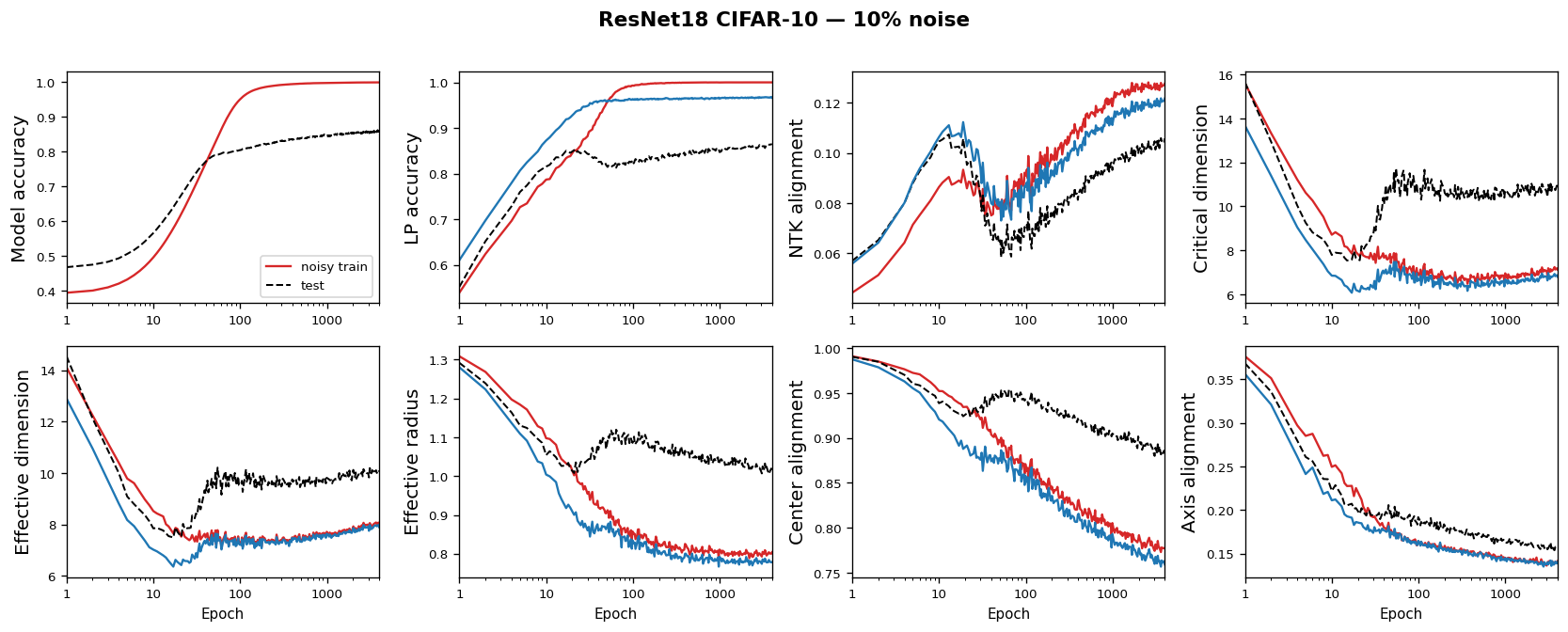}
    \caption{Epoch-wise double descent through label noises, 10\% label noise.}
    \label{fig:all dd 10 noise}
\end{figure}

\begin{figure}[h!]
    \centering
    \includegraphics[width=\linewidth]{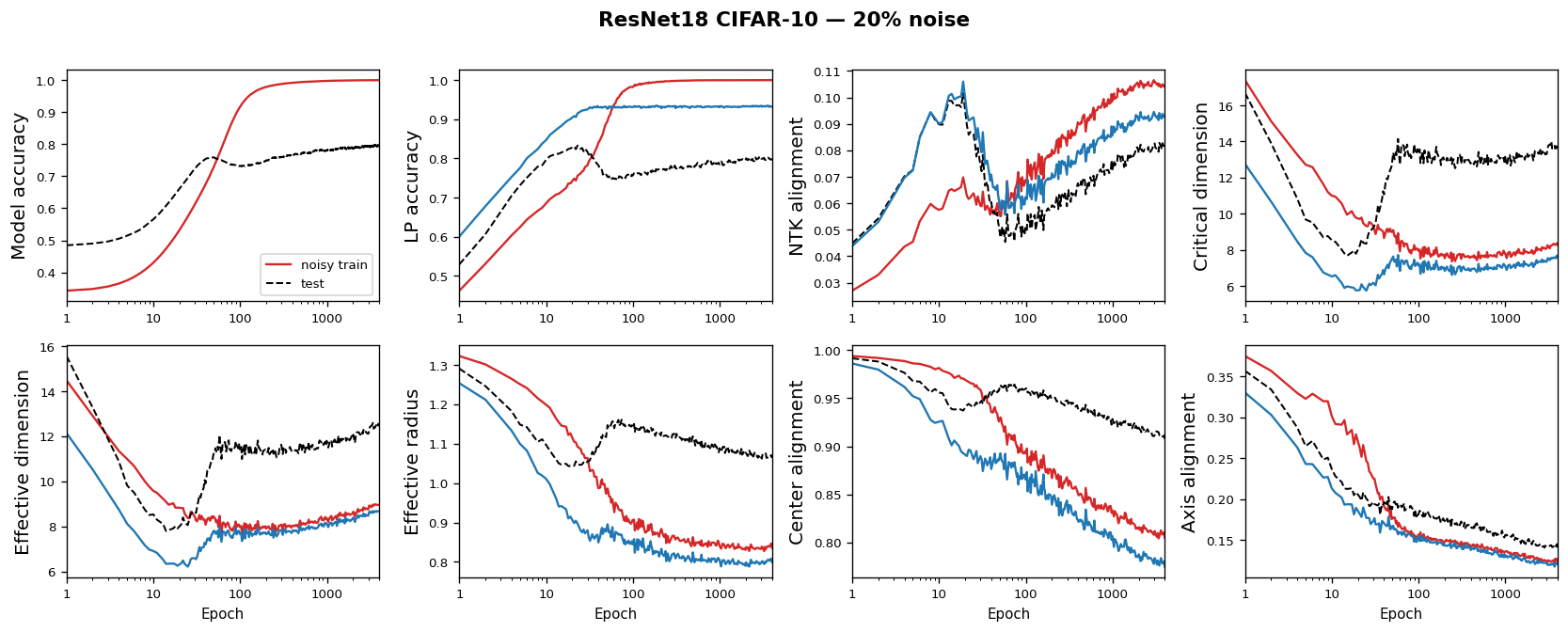}
    \caption{Epoch-wise double descent through label noises, 20\% label noise.}
    \label{fig:all dd 20 noise}
\end{figure}

We present additional results for the epoch-wise double descent experiment across three levels of label noise: $0\%$ (\autoref{fig:all dd 0 noise}), $10\%$ (\autoref{fig:all dd 10 noise}), and $20\%$ (\autoref{fig:all dd 20 noise}). At $0\%$ noise, the double descent pattern and diagnostic signatures are absent, serving as a clean baseline. As label noise increases, the double descent dip in test accuracy becomes more pronounced, and the diagnostic signatures—plateauing LP accuracy, rising $N_\text{crit}$, and increasing train-test NTK alignment gap—become increasingly visible, consistent with the two-stage mechanism described in~\autoref{sec:double descent}.

\clearpage
\subsection{Event detection in the plots in main text}\label{app:event detection}

\textbf{Grokking cases.}
We identify three critical epochs from the training curves to partition training into four phases. Train-100 is the first epoch at which training accuracy reaches 99\% and remains there for at least 3 consecutive checkpoints. 
Grok onset is the first epoch at which test accuracy exceeds 5\% of its eventual maximum (computed on a smoothed curve) and remains above that threshold for at least 3 consecutive checkpoints; for SparseParity, where test accuracy starts near chance, we instead use an absolute threshold of 60\%. 
Grok offset is defined analogously, using 95\% of the eventual maximum test accuracy. Smoothing is performed with a uniform moving average of window size 5 prior to threshold crossing detection. These three events divide training into four phases, shown as background shading in all epoch-axis plots: memorization (start → train-100, green), post-memorization (train-100 → onset, blue), grokking transition (onset → offset, orange), and post-grokking (offset → end, red).

\textbf{Non-grokking cases.}
For models that do not exhibit grokking, we define two critical epochs: train-100, the first epoch at which training accuracy reaches 99\% and remains there for at least 3 consecutive checkpoints, and test-100, defined analogously for test accuracy. These two events divide training into three phases, shown as background shading in non-grokking plots: pre-memorization (start → train-100, blue), generalization gap (train-100 → test-100, yellow), and post-generalization (test-100 → end, red). For MNIST, where test accuracy plateaus below 100\%, the test-100 threshold is never reached; we therefore use only the train-100 boundary, yielding two phases: memorization (start → train-100, green) and post-memorization (train-100 → end, blue).

\textbf{Double descent cases.}
For double descent experiments (ResNet18 trained on CIFAR-10 with label noise), we identify two critical epochs from the validation accuracy curve. Peak is the epoch of maximum validation accuracy, corresponding to the early-stopping optimum before double descent degrades generalization. Recovery is the first epoch after the subsequent valley at which the smoothed validation accuracy begins a sustained monotone increase, detected as the first point in the post-valley curve where accuracy increases for at least 5 consecutive checkpoints (smoothing window of 15). These two events divide training into three phases: pre-peak (start → peak, green), double descent valley (peak → recovery, blue), and post-recovery (recovery → end, red). Note that for the 0\% noise control, where no double descent occurs, we instead use only train-100 as a single dividing event between the memorization and post-memorization phases.

\textbf{No double descent cases.} 
For the clean training case (0\% label noise), no double descent occurs and the validation accuracy increases monotonically after memorization. We therefore use a simpler two-phase scheme: train-100, the first epoch at which training accuracy reaches 99\% on the clean labels, divides training into a memorization phase (start → train-100, green) and a generalization phase (train-100 → end, red). Since there is no valley or recovery event, the double descent phase boundaries do not apply.

\newpage
\section{Task-specific analysis}\label{app:task-specific}
The representation-readout decomposition described in the main text is task-agnostic.
In this section, we make a preliminary exploration of how to leverage more fine-grained task structures in the analysis, e.g., the algebraic structure relating different class pairs (for instance, the fact that the label for $(a,b)$ and the label for $(a+1,b)$ differ by exactly one in $\mathbb{Z}_p$).
We take a first step toward incorporating task structure into the analysis. Specifically, we compute the pairwise critical dimension $N_\mathrm{crit}(c,c')$ for every label pair $(c,c')$, obtaining a $C \times C$ matrix that captures the fine-grained geometry of the representation. We then ask whether these structured representations support transfer to a related but distinct task via a frozen linear probe.

\begin{figure}[h!]
  \centering
  \includegraphics[width=0.85\linewidth]{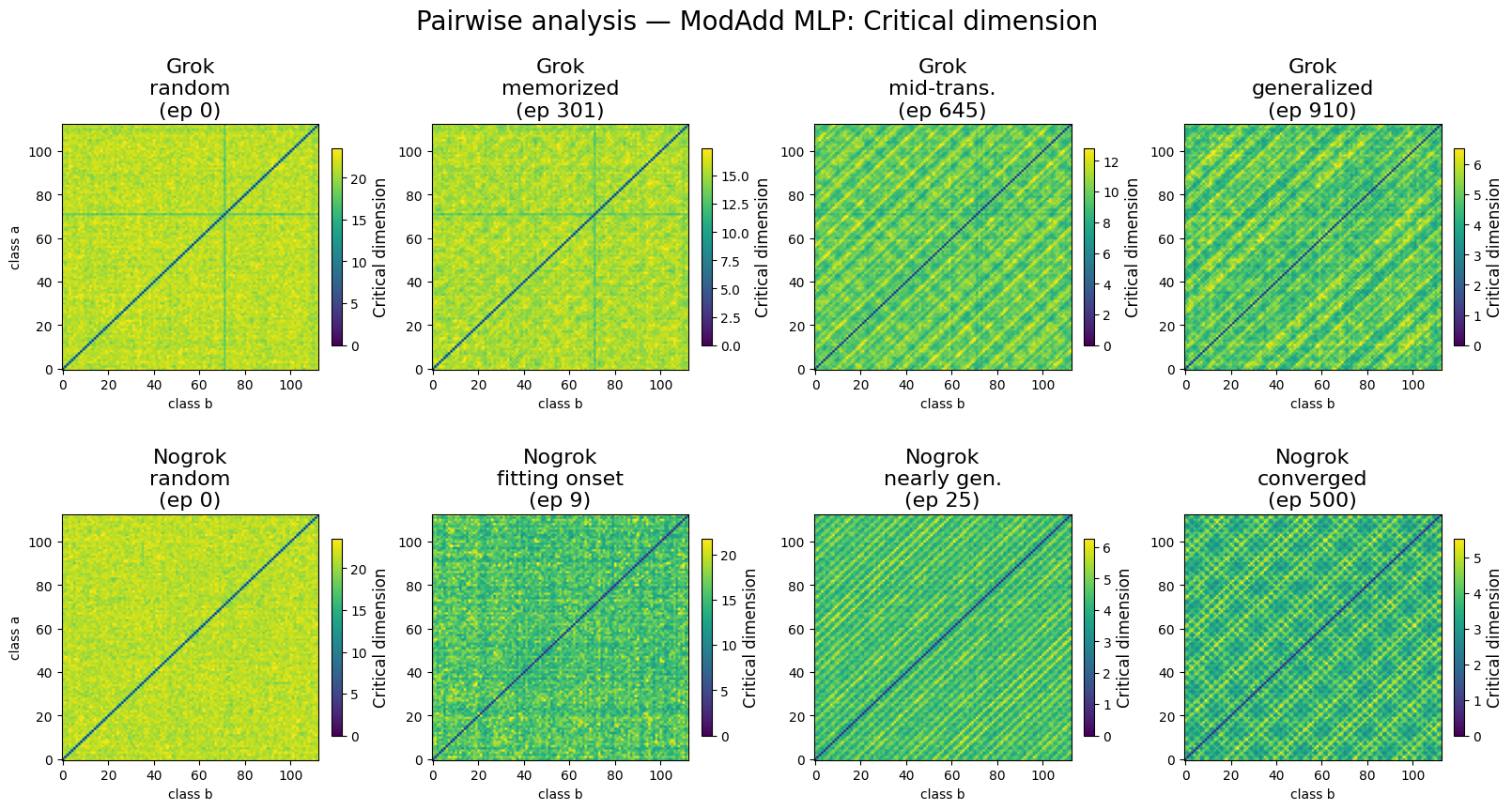}
  \includegraphics[width=0.85\linewidth]{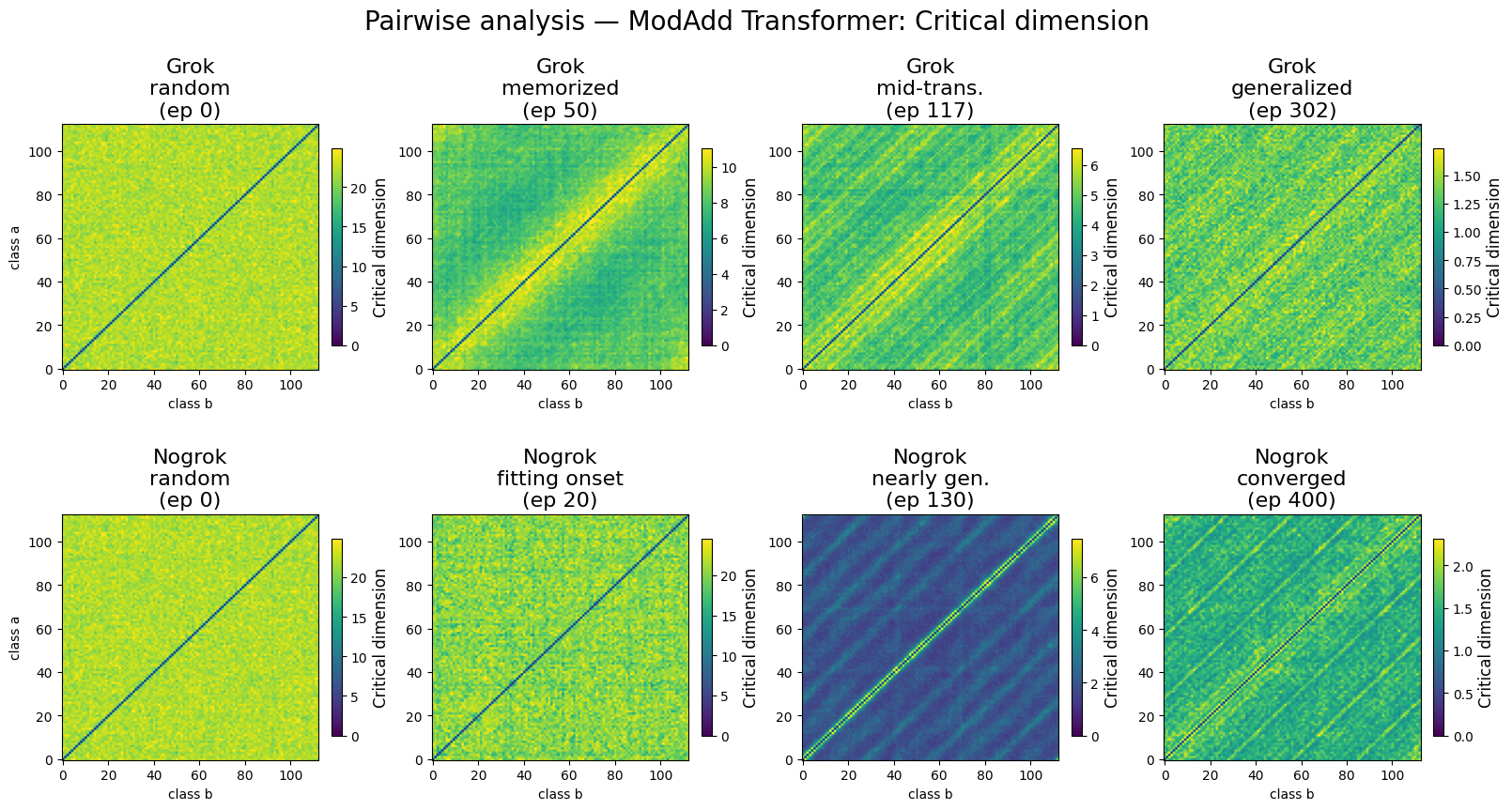}
  \caption{Pairwise critical dimension ($N_\mathrm{crit}$) matrices for the ModAdd task. Top: MLP. Bottom: Transformer. Each panel shows a $113 \times 113$ matrix (class $a$ vs.\ class $b$). Rows correspond to grokking (top) and non-grokking (bottom) conditions; columns correspond to four representative epochs. Colorbars are scaled to the 99th percentile of each panel independently.}
  \label{fig:heatmap_modadd}
\end{figure}

\begin{figure}[h!]
  \centering
  \includegraphics[width=0.85\linewidth]{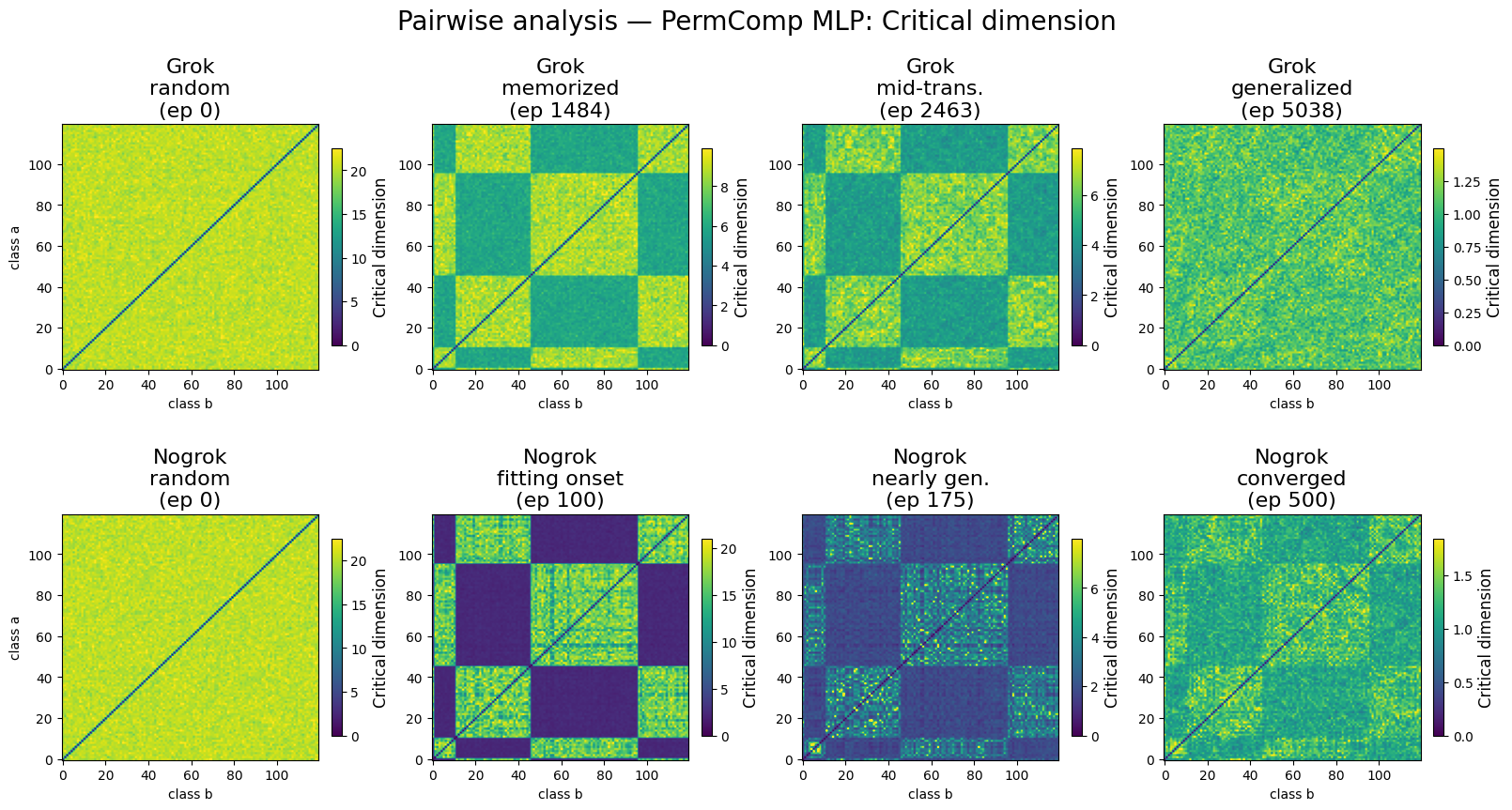}
  \includegraphics[width=0.85\linewidth]{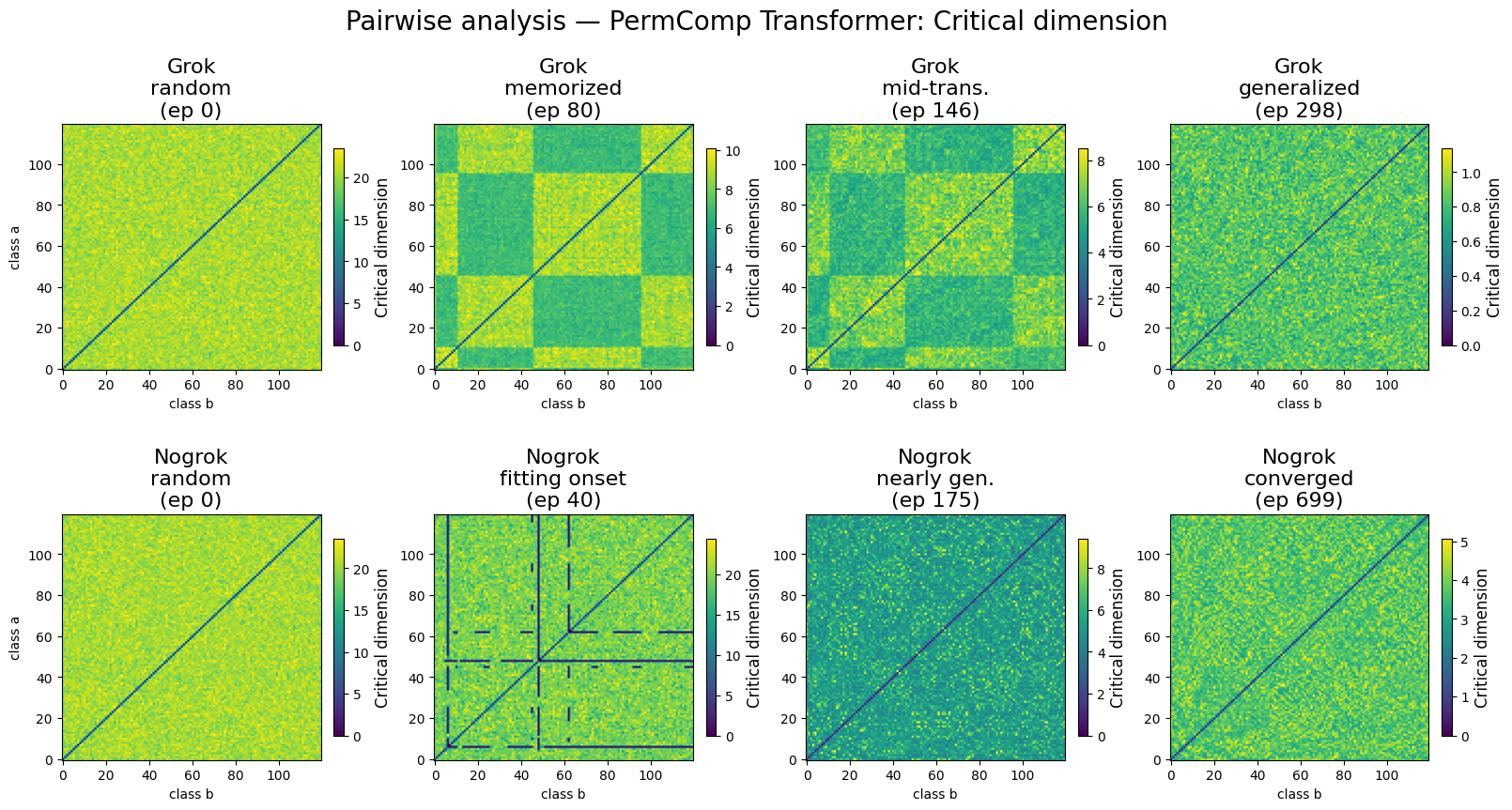}
  \caption{Pairwise critical dimension matrices for the PermComp task. Top: MLP. Bottom: Transformer. Each panel shows a $120 \times 120$ matrix with elements of $S_5$ reordered by conjugacy class (cycle type): identity, transpositions, double transpositions, 3-cycles, 4-cycles, (3+2)-cycles, 5-cycles.}
  \label{fig:heatmap_permcomp}
\end{figure}

\subsection{Pairwise critical dimension heatmaps}

\autoref{fig:heatmap_modadd} and~\autoref{fig:heatmap_permcomp} show the $p \times p$ critical dimension matrices at four representative epochs for the grokking and non-grokking conditions, for both the MLP and Transformer architectures. For \textbf{modular addition} ($p = 113$,~\autoref{fig:heatmap_modadd}), the matrices are unordered (rows/columns indexed by residue class $0, \ldots, 112$). For \textbf{permutation composition} ($p = 120$,~\autoref{fig:heatmap_permcomp}), classes are reordered by conjugacy class (cycle type) of the corresponding element of $S_5$: identity, transpositions, double transpositions, 3-cycles, 4-cycles, (3+2)-cycles, and 5-cycles.

At early epochs, both grokking and non-grokking models show roughly uniform matrices. As training progresses, the grokking models develop visually structured patterns in the pairwise critical dimension matrices. The non-grokking models also develop structure, but the patterns differ qualitatively from those of the grokking models. We do not offer an interpretation of these patterns here; we simply note their existence as a starting point for future investigation.

\subsection{Transfer to related tasks}
To probe whether the pairwise structure observed above has behavioral consequences, we freeze each trained encoder and fit a linear probe on a related transfer task. For modular addition we predict the \emph{difference} $(a - b) \bmod p$ from all $\binom{p}{2} = 6328$ ordered pairs $(a,b)$ with $a > b$, using an independent 60/40 train/test split (seed 42). For permutation composition, we predict the \emph{inverse composition} $\sigma_i \circ \sigma_j^{-1} \in S_5$ (120 classes) from all $14400$ ordered pairs, again with an independent 60/40 split. Both probes are trained with AdamW (lr $= 10^{-3}$, wd $= 1.0$, $\beta = (0.9, 0.98)$, 200 epochs, full batch).

\begin{figure}[h!]
  \centering
  \includegraphics[width=0.8\linewidth]{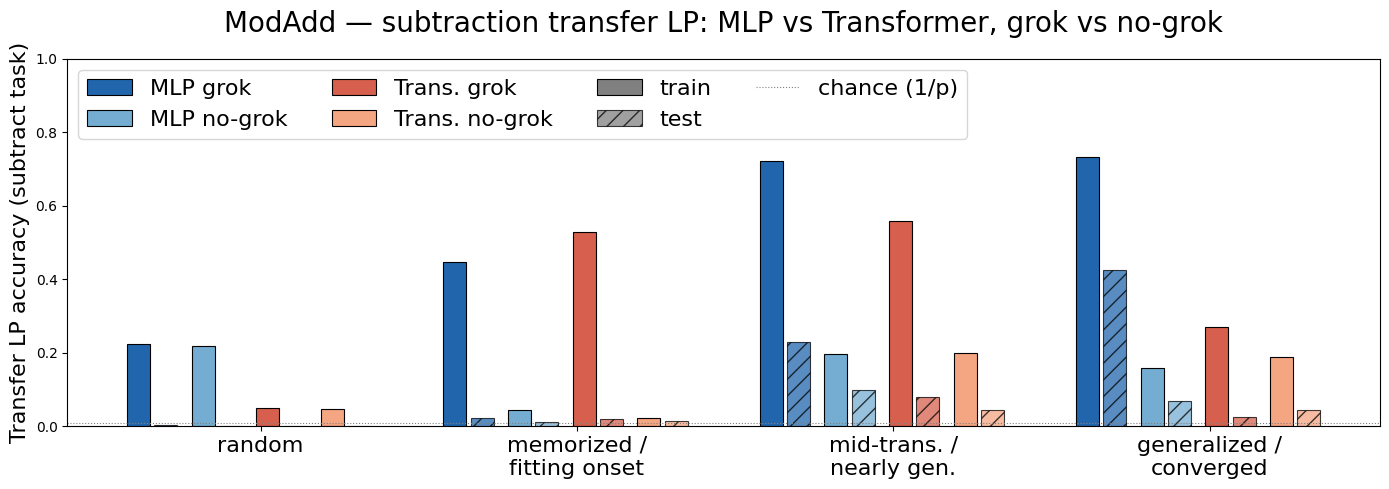}
  \includegraphics[width=0.8\linewidth]{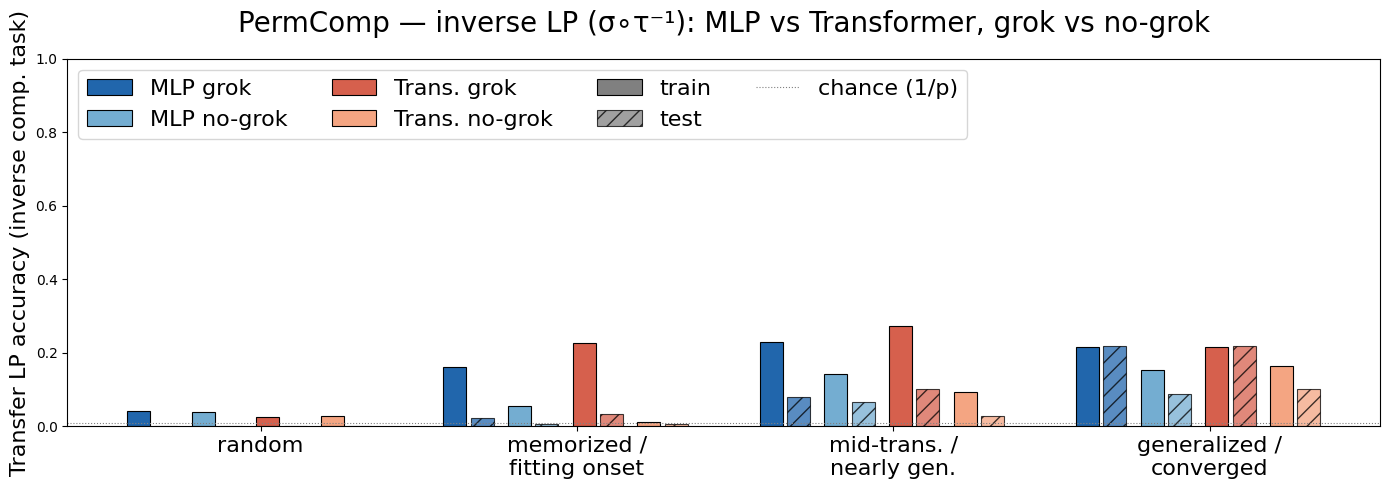}
  \caption{Transfer linear probe accuracy. Top: ModAdd subtraction task ($(a-b) \bmod p$, chance $= 1/113$). Bottom: PermComp inverse composition task ($\sigma \circ \tau^{-1}$, chance $= 1/120$). Solid bars: train accuracy; hatched bars: test accuracy. Each cluster of bars corresponds to one training stage.}
  \label{fig:transfer}
\end{figure}

\autoref{fig:transfer} shows the linear probe accuracy on the transfer tasks at the representative epochs. For both tasks and both architectures, the grokked models consistently achieve higher transfer test accuracy than the non-grokking models at late epochs. The two architectures show different profiles: for ModAdd, the MLP grokked model achieves substantially higher transfer accuracy than the Transformer; for PermComp, the differences are more modest. The mechanism behind these observations is left for future work.

\subsection{Discussion}
The pairwise critical dimension matrices reveal that grokked and non-grokked representations organize class-pair geometry differently. While the mechanisms behind these patterns are unclear, we observe functional consequences via transfer linear probes (\autoref{fig:transfer}): grokked models consistently outperform non-grokking ones at late training stages.
These findings are preliminary and descriptive; we view them as a starting point for a more principled, task-aware representation analysis.

\section{Metric Consistency Quantification}\label{app:metric consistency}
In our analysis framework, we use several representation-based measures (e.g., NTK alignment, critical dimension, GLUE geometric measures) to probe the learning dynamics.
These metrics indicate that learning progresses continuously throughout training, rather than exhibiting the sharp phase transition suggested in prior work using the lazy-to-rich account~\citep{kumar2023grokking}. They also provide a stable signal for tracking learning dynamics over time~\citep{chou2025featurelearninglazyrichdichotomy}. In particular, even in the presence of grokking—characterized by a delayed and rapid increase in test accuracy relative to training accuracy— the train and test curve of these representation metrics closely mirrors each other during training. This suggests that representation metrics offer a more consistent and informative measure of training progress than standard loss or accuracy curves.

\begin{figure}[h!]
    \centering
    \includegraphics[width=1.0\linewidth]{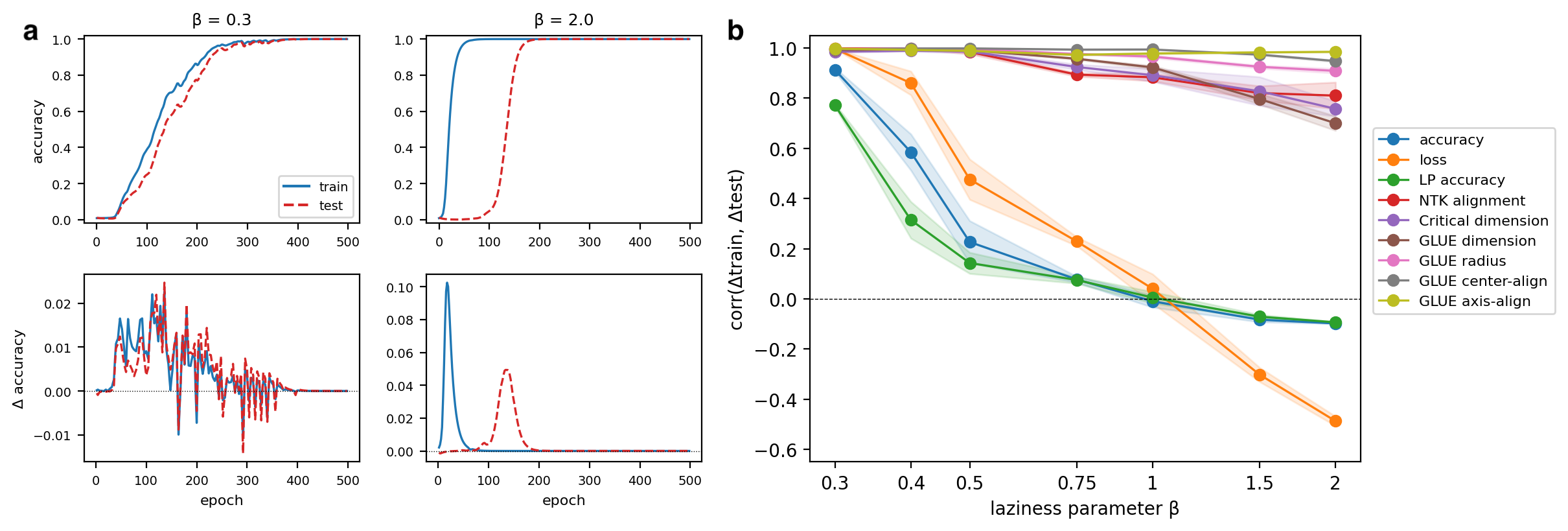}
    \caption{Consistency between train and test curves across different values of the laziness parameter \(\beta\). Consistency is defined as the correlation between the first differences of the train and test curves after spline smoothing. \textbf{a}, Time series of train and test curves for different values of \(\beta\), along with their corresponding first differences. \textbf{b}, For standard metrics (loss and accuracy), this correlation decreases and eventually becomes negative as \(\beta\) increases, indicating the onset of grokking. In contrast, NTK alignment, critical dimension, and GLUE geometric measures maintain consistently high correlation, providing a more reliable signal for tracking learning dynamics. Shaded regions denote the standard deviation across random seeds.}
    \label{app:beta}
\end{figure}

\subsection{Implementation and definition of the consistency measure}

We quantify consistency between train and test curves by measuring their co-movement via correlation ~\citep{shumway2025time}. To reduce noise, we first smooth each trajectory using a cubic smoothing spline, which provides a stable, nonparametric estimate of the underlying trend by balancing data fidelity and curvature ~\citep{wahba1990estimating, hastie2017smoothing}. Formally, the smoothing spline is defined as the solution to the regularized least-squares problem:
\[
\sum_{i=1}^{n} \left( y_i - f(x_i) \right)^2 
+ \lambda \int \left( f''(u) \right)^2 \, du,
\]
where \( f \) is a spline function and \( \lambda \) is a regularization parameter selected via generalized cross-validation. We then compute first differences of the smoothed trajectories, \( \Delta X_t = X_t - X_{t-1} \), and measure consistency as the Pearson correlation between train and test dynamics:
\[
\mathrm{Corr}\left( \Delta X^{\text{train}}_t, \; \Delta X^{\text{test}}_t \right),
\]
where \( X_t \) denotes the smoothed time series of the metric of interest (e.g., loss, accuracy, LP accuracy, or GLUE score) over training. 

\subsection{Consistency of different metrics across learning rates}

In \autoref{app:beta}, we report the consistency measure across different learning rates for the Modular Addition task using a Transformer architecture. Following Kumar et al.~\citep{kumar2023grokking}, the laziness parameter $\beta$ (denoted $\alpha$ in their work) controls the extent of grokking, with larger values inducing stronger effects. Specifically, $\beta$ is implemented by rescaling the output logits, i.e., multiplying the network outputs by a factor $\beta$ and dividing the learning rate by a factor of $\beta$, thereby inducing lazy training for large $\beta$. 

As shown in \autoref{app:beta}, for standard metrics such as loss and accuracy, the correlation between train and test dynamics becomes negative when $\beta>1$, indicating the onset of grokking. In contrast, for the representation metrics considered in our analysis framework (NTK alignment, critical dimension, GLUE geometric measures), the correlation remains consistently high and close to 1 across all learning rates. These results suggest that the proposed consistency measure captures the grokking phenomenon, while GLUE representation metrics provide a stable and reliable signal for tracking learning dynamics over time.

\begin{figure}[ht]
    \centering
    \includegraphics[width=1\linewidth]{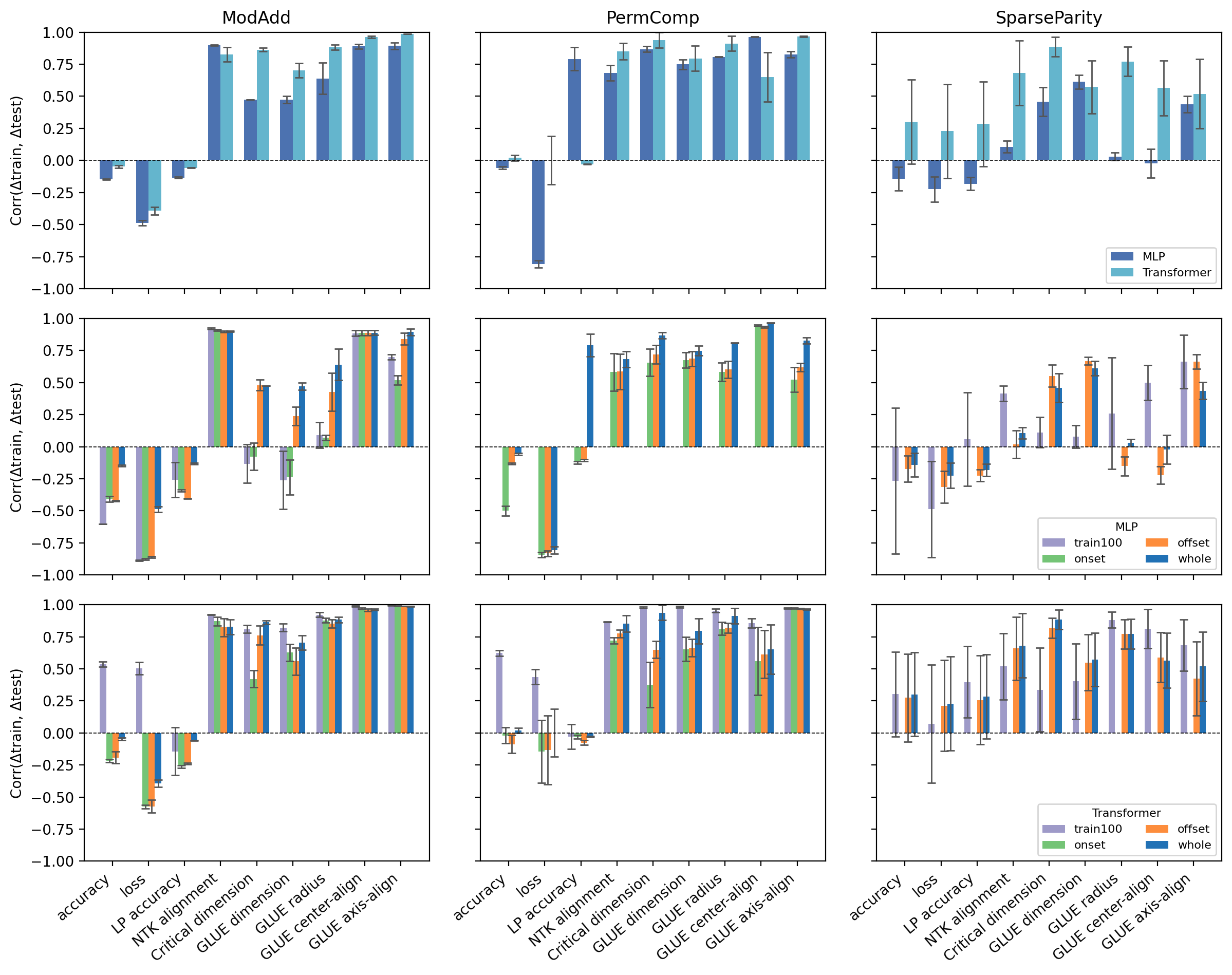}
    \caption{Consistency between the test and train curve across tasks, architectures, and training stages. The top row reports overall consistency across different tasks and architectures. Standard metrics (loss and accuracy) exhibit predominantly negative correlations, while representation metrics (NTK alignment, critical dimension, and GLUE geometric measures) maintain high consistency, close to 1. The middle and bottom rows further analyze consistency computed over different training segments (train100, onset, offset, and full trajectory). The onset and offset phases of grokking significantly lowers the consistency. Error bars indicate the standard error of the mean across seeds.}
    \label{app:consistensy}
\end{figure}

\subsection{Consistency across tasks, architectures, and training stages}

In \autoref{app:consistensy}, we report the consistency measure across different tasks and architectures. For standard metrics such as loss and accuracy, the correlation between train and test dynamics is often negative or close to zero, indicating the presence of grokking and poor alignment of the train and test curve in learning dynamics. In contrast, the representation metrics described above exhibit consistently positive high correlations across most tasks and architectures. The last two rows of \autoref{app:consistensy} further analyze consistency across different stages of training. We observe that the onset of grokking coincides with a drawdown of the correlation between train and test dynamics.

\end{document}